\documentclass[11pt]{article}

\PassOptionsToPackage{table,dvipsnames}{xcolor}
\usepackage[preprint]{acl}

\usepackage{times}
\usepackage{latexsym}
\usepackage[T1]{fontenc}
\usepackage[utf8]{inputenc}
\usepackage{microtype}
\usepackage{inconsolata}

\usepackage{amsmath}
\usepackage{amssymb}
\usepackage{mathtools}
\usepackage{amsthm}
\usepackage{url}
\usepackage{booktabs}
\usepackage{siunitx}
\usepackage{colortbl}
\usepackage{multirow}
\usepackage{enumitem}
\usepackage{pifont}
\usepackage{graphicx}
\usepackage{subcaption}
\usepackage{wrapfig}
\usepackage[most]{tcolorbox}
\tcbuselibrary{listings}

\newcommand{\cmark}{\ding{51}}
\newcommand{\xmark}{\ding{55}}

\definecolor{mygreen}{rgb}{0.0, 0.5, 0.0}
\definecolor{myred}{rgb}{0.7, 0.0, 0.0}
\definecolor{text_ding_blue}{HTML}{215E9A}

\usepackage[capitalize,noabbrev]{cleveref}

\theoremstyle{plain}

\theoremstyle{definition}

\theoremstyle{remark}

\title{KnowMT-Bench: Benchmarking Knowledge-Intensive\\Long-Form Question Answering in Multi-Turn Dialogues}

\author{
\bfseries Junhao Chen$^{1,3*}$, Yu Huang$^{1,4*}$, Siyuan Li$^{1*}$, Rui Yao$^{1}$, Hanqian Li$^{1}$ \\
\bfseries Hanyu Zhang$^{1}$, Jungang Li$^{1}$, Jian Chen$^{1}$, Bowen Wang$^{3}$, Xuming Hu$^{1,2\dagger}$ \\
$^{1}$Hong Kong University of Science and Technology (Guangzhou) \\
$^{2}$Hong Kong University of Science and Technology, $^{3}$The University of Osaka \\
$^{4}$City University of Hong Kong \\
\texttt{\{jchen024,yhuang489,sli974\}@connect.hkust-gz.edu.cn, xuminghu@hkust-gz.edu.cn} \\
$^{*}$Equal contribution. $^{\dagger}$Corresponding author.
}

\begin{document}
\maketitle

\begin{abstract}
Multi-Turn Long-Form Question Answering (MT-LFQA) is a key application paradigm of Large Language Models (LLMs) in knowledge-intensive domains. However, existing benchmarks are limited to single-turn dialogue, while multi-turn dialogue benchmarks typically assess other orthogonal capabilities rather than knowledge-intensive factuality. To bridge this critical gap, we introduce \textbf{KnowMT-Bench}, a new benchmark designed to systematically evaluate MT-LFQA for LLMs across knowledge-intensive fields, including medicine, finance, and law. To faithfully assess the model's real-world performance, KnowMT-Bench employs a dynamic evaluation setting where models generate their own multi-turn dialogue histories given logically progressive question sequences. Our experiments on a diverse suite of LLMs show a clear degradation in both factual capability and information delivery efficiency in multi-turn dialogue. We further probe the underlying causes and investigate mitigation strategies spanning domain-specific finetuning, active context denoising, and RAG. The results show domain-dependent gains from finetuning, while active context denoising and RAG more consistently mitigate these issues, with RAG notably reversing this performance degradation. These findings underscore the importance of our benchmark for evaluating LLMs' conversational factual capabilities in real-world applications. Code and data are publicly available.\footnote{
\href{https://github.com/Sven-LI-sankyuu/KnowMT-Bench}
{\texttt{https://github.com/Sven-LI-sankyuu/KnowMT-Bench}}
}
\end{abstract}

\section{Introduction}

Large Language Models (LLMs) have become an increasingly important interface for accessing and interacting with knowledge, especially in knowledge-intensive domains such as medicine, finance, and law~\citep{wu2023bloomberggpt, huang2023lawyer, singhal2025toward, wang2024direct}. A common real-world usage paradigm is consultation-style interaction, where users progressively clarify their needs through multi-turn dialogues and expect a detailed long-form answer that synthesizes multiple key points~\citep{kurtz1996calgary, cfp2020practicestandards}. Building on these observations, we formalize such a challenge as Multi-Turn Long-Form Question Answering (MT-LFQA): an open-domain QA task that requires the model to synthesize multiple facts into a paragraph-level answer for the final-turn question, given the context of multi-turn dialogue history.

\begin{table}[t]
\renewcommand{\arraystretch}{1} %
\setlength{\tabcolsep}{3pt}  %
\centering
\small
\caption{Comparison of KnowMT-Bench with existing QA benchmarks. \# Avg. Tokens refer to the token counts of the ground truth answer, computed by the GPT-4o tokenizer. \textbf{*}: FintextQA is partially open-sourced; we report the 75-word value from their paper. MT denotes Multi-Turn, and OD refers to Open-Domain.}
\begin{tabular}{lcccc}
\toprule
Benchmark  & \# Avg.\ Tokens & MT & OD  \\
\midrule
CoQA~\citep{reddy2019coqa} &  2.52  & \cmark & \xmark  \\
QASA~\citep{lee2023qasa}  & 50 & \xmark & \xmark  \\
K-QA~\citep{manes2024k}  & 119.89 & \xmark & \cmark  \\
MedLFQA~\citep{jeong2024olaph}      & 132.86 & \xmark & \cmark \\
FintextQA~\citep{chen2024fintextqa}   & 75\textbf{*} & \xmark & \cmark  \\
\midrule
\textbf{KnowMT-Bench (ours)} &  95.15 & \cmark & \cmark \\
\bottomrule
\end{tabular}%
\label{tab:benchmarks}
\vspace{-10pt}
\end{table}

\begin{figure*}[t]
    \centering
    \includegraphics[width=0.98\textwidth]{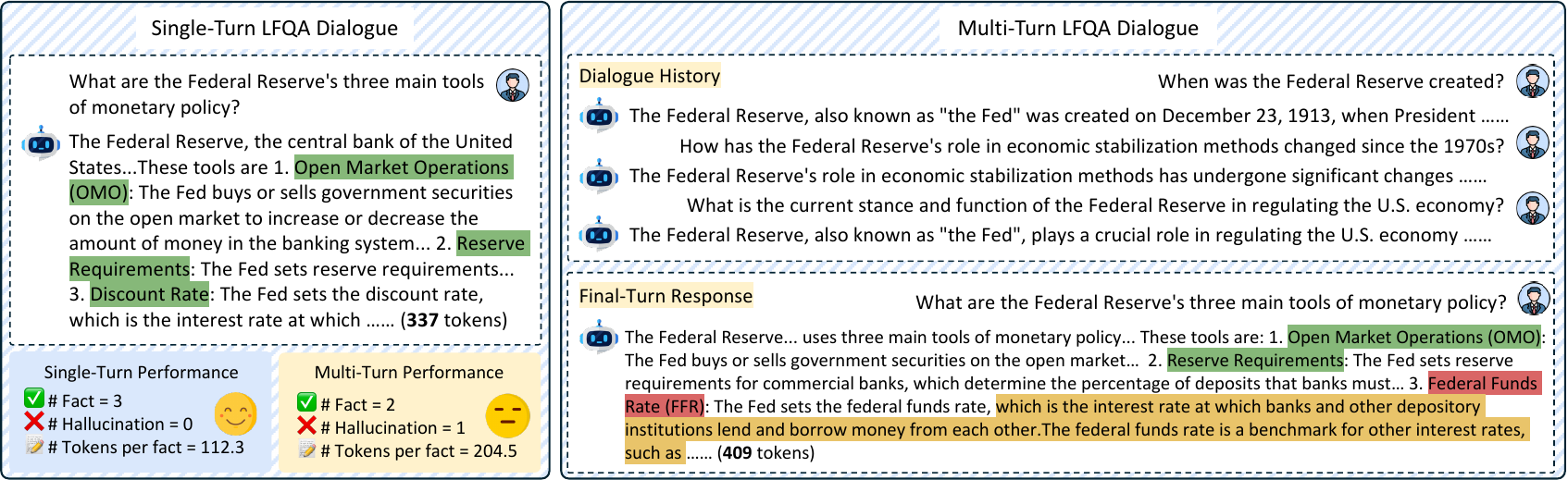}
    \caption{Single- vs. Multi-Turn LFQA on Llama-3.3-70B-Instruct; \colorbox[RGB]{134,184,121}{correct}, \colorbox[RGB]{233,196,106}{irrelevant}, and \colorbox[RGB]{217,102,102}{faulty} statements.}
    \label{fig:1_introduction}
    \vspace{-10pt}
\end{figure*}

As the common domains of MT-LFQA are often knowledge-intensive and high-stakes, the answers provided must be factually comprehensive and accurate, while exhibiting minimal factual hallucination. While numerous single-turn Long-Form Question Answering (LFQA) benchmarks have emerged, such as K-QA in medicine~\citep{manes2024k}, FinTextQA in finance~\citep{chen2024fintextqa}, and cLegal-QA in law~\citep{wang2025clegal}, the challenges are substantially amplified in a multi-turn context. In MT-LFQA, the dialogue history can introduce redundant information, which acts as noise to compromise the model's ability to generate a long-form answer adhering to these standards~\citep{laban2025llms}. As demonstrated in Figure~\ref{fig:1_introduction}, the model that produces factually sound answers in the single-turn setting generates a significant factual error within the multi-turn context. Concurrently, the volume of non-factual content increases, obscuring key information, which degrades the overall utility of the answer~\citep{zhou2024processing, hackenburg2025levers}. Therefore, the single-turn setup of existing LFQA benchmarks cannot faithfully assess a model's performance in the more challenging MT-LFQA scenario.

Existing multi-turn benchmarks are also misaligned with the specific challenges of MT-LFQA. First, conventional conversational QA benchmarks such as QuAC~\citep{choi2018quac}, CoQA~\citep{reddy2019coqa} are designed for short, often extractive, answers, making them unable to effectively assess the integration of multiple facts into a paragraph-level answer (see Table~\ref{tab:benchmarks} for a detailed comparison). Second, contemporary dialogue benchmarks for LLMs also prove unsuitable, as they either adopt evaluation paradigms like LLM-as-a-judge~\citep{zheng2023judging, bai2024mt}, which are inadequate for rigorous factuality assessment by relying on the judge's own fallible parametric knowledge~\citep{fu2023large, chen2024humans}, or they prioritize orthogonal capabilities like instruction-following~\citep{he2024multi}, fairness~\citep{fan2024fairmt}, thereby diluting the focus on factual completeness and correctness in LFQA. This clear gap necessitates a purpose-built benchmark to systematically measure factual capability and information delivery efficiency within MT-LFQA.

To bridge this critical gap, we introduce \textbf{KnowMT-Bench}, a benchmark designed to conduct a systematic study of MT-LFQA. We ground our research in medicine, finance, and law, which are the common knowledge-intensive domains. Our benchmark is thus founded on 801 evidence-grounded LFQA instances from these domains. To simulate a realistic human-LLMs interaction, the benchmark requires models to generate their own dialogue history following human-authored question sequences that progress logically. The final-turn answers are assessed using a comprehensive framework that leverages an automated fine-grained, Natural Language Inference (NLI)-based pipeline inspired by previous works~\citep{manes2024k, jeong2024olaph}, to analyze factual capability and information delivery efficiency. The reliability of the automated pipeline is supported by validating each step with human experts.

Our experiments on LLMs reveal that multi-turn contexts pose a severe challenge: model factual capability shows a pronounced degradation when shifting from single-turn to multi-turn LFQA, accompanied by a significant increase in verbosity that reduces information delivery efficiency (Section~\ref{sec:main_results}). 
Our analysis further reveals that the decline in information delivery efficiency is strongly associated with dialogue length, whereas factual capability degradation is mainly driven by contextual noise from the model's self-generated history, particularly relevant misinformation (Section~\ref{sec:pd}). In addition, we explore mitigation strategies spanning domain-specific finetuning, active context denoising, and RAG. Domain-specific finetuning yields domain-dependent benefits, while active context denoising and RAG more consistently alleviate these issues, with RAG notably capable of reversing this performance degradation (Section~\ref{sec:mitigation_strategies}).
These findings highlight the limitations of single-turn evaluations and underscore the necessity of our benchmark for assessing and improving the conversational robustness of LLMs under real-world knowledge-intensive applications.

In summary, our main contributions are as follows: (1) We introduce \textbf{KnowMT-Bench}, the new benchmark designed for the systematic evaluation of LLMs in MT-LFQA. (2) We design a comprehensive \textbf{evaluation framework} for MT-LFQA, employing a human-validated, automated pipeline to assess both \textbf{factual capability} and \textbf{information delivery efficiency}. (3) Our experimental results reveal a pronounced degradation in both model factual capability and information delivery efficiency within the multi-turn contexts. Crucially, we find that efficiency degradation is closely associated with dialogue length, while factual capability degradation is mainly attributable to contextual noise from self-generated history, and we evaluate mitigation strategies covering domain-specific finetuning, active context denoising, and RAG.

\section{Task Definition}
\label{sec:task_definition}

As an early systematic study of MT-LFQA, we begin by formalising this task, introducing notation to facilitate our analysis, and delimiting the scope of evaluation. We first formalise the single-turn LFQA task and then extend it to the multi-turn setting, which is central to this work.

\textbf{LFQA} is an open-domain QA task where a model is required to synthesise multiple facts into a paragraph-level answer. We formalise this setting as follows: the input is a set of knowledge-intensive questions $\mathcal{Q}=\{q_1,q_2,\dots,q_N\}$. For each question $q_i \in \mathcal{Q}$, the ground-truth consists of a set of must-have facts $\mathcal{F}_i = \{ f_{i1}, f_{i2}, \dots, f_{iM_i} \}$, with the collection of all fact sets denoted by $\mathbb{F} = \{\mathcal{F}_1,\mathcal{F}_{2},\dots,\mathcal{F}_N\}$. These facts are composed into free-form ground-truth answers $\mathcal{G}=\{g_1,g_2,\dots,g_N\}$, where each $g_i$ provides a complete and non-redundant representation of $\mathcal{F}_i$.
To evaluate this task, a QA model $\mathcal{M}$ is applied to the question set $\mathcal{Q}$ to generate answers $\mathcal{A}=\{a_1,a_2,\dots,a_N\}$, where each $a_i=\mathcal{M}(q_i)$. The generated answers $\mathcal{A}$ are then compared against the ground-truth answers $\mathcal{G}$, or equivalently against the supporting facts $\mathbb{F}$.

\textbf{MT-LFQA} is defined as the task that performs LFQA where the model is conditioned on the preceding conversational histories, and its definition naturally extends the single-turn setting formalised above. To formalise this task, we consider a set of $K$ dialogues $\mathcal{D} = \{d_1, \dots, d_K\}$. Each dialogue $d_k \in \mathcal{D}$ consists of a sequence of $N_k$ turns, where $N_k$ is the total number of turns in dialogue $d_k$. The conversational context for the final turn is the preceding history, denoted as $H_{k} = (q^{(k)}_1, a^{(k)}_1, \dots, q^{(k)}_{N_k-1}, a^{(k)}_{N_k-1})$. The final-turn question $q^{(k)}_{N_k}$ is a single-turn LFQA question from the set $\mathcal{Q}$. For a given $q^{(k)}_{N_k}$, we denote its corresponding ground-truth fact set and reference answer as $\mathcal{F}_j$ and $g_j$ respectively, where $q^{(k)}_{N_k} = q_j$ for some index $j \in \{1, \dots, N\}$.

In task MT-LFQA, a model $\mathcal{M}$ is required to generate a factual and complete long-form answer for the final-turn question $q^{(k)}_{N_k}$, conditioned on the context $H_k$. The model's output is $a^{(k)}_{N_k} = \mathcal{M}(H_{k}, q^{(k)}_{N_k})$ , where $H_{k}$ can be provided under two settings. In a \textit{static context} setting, the history is pre-defined, composed of question-answer pairs authored by humans or generated by a model. In a \textit{dynamic context} setting, which simulates an interactive session, the history is constructed on-the-fly by having the model $\mathcal{M}$ generate each answer in response to a pre-defined sequence of questions. Finally, the generated answer $a^{(k)}_{N_k}$ is then evaluated against its ground-truth ($\mathcal{F}_j$ and/or $g_j$), following the same assessment protocol as in LFQA.

\begin{figure*}[t]
    \centering
    \includegraphics[width=0.98\linewidth]{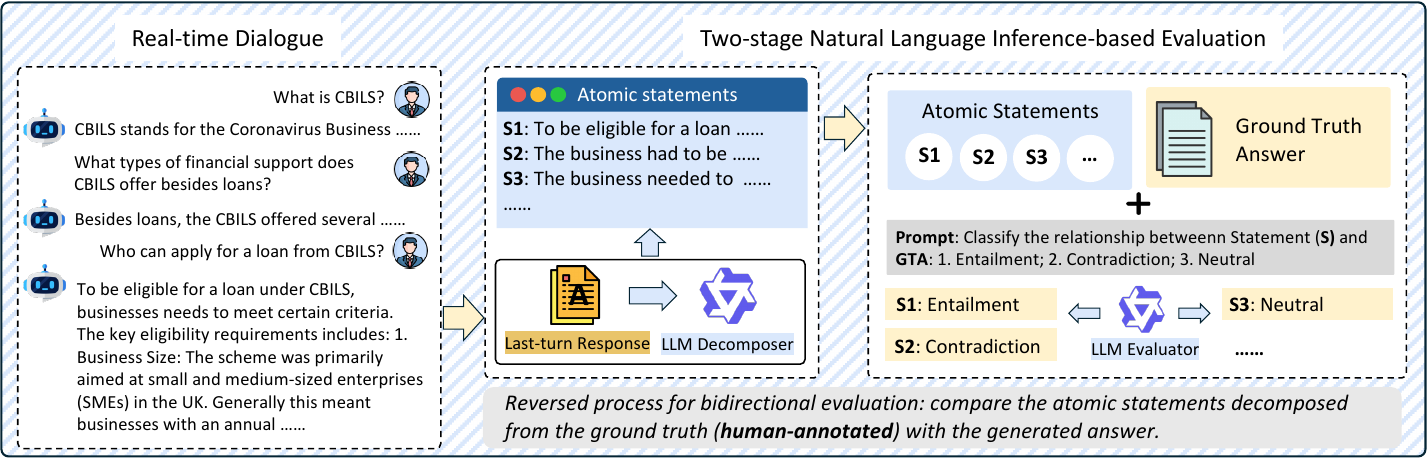}
    \caption{Overview of the two-stage natural language inference-based evaluation. Here, \textbf{GTA} refers to \textbf{G}round \textbf{T}ruth \textbf{A}nswer, and the detailed prompts for evaluation are provided in Appendix~\ref{app:prompts}.}%
    \label{fig:3_evaluation}
    \vspace{-10pt}
\end{figure*}

\section{KnowMT-Bench}

Based on the task definition, we introduce a new benchmark, \textbf{KnowMT-Bench} in this section.%

\subsection{Single-Turn LFQA Data Collection}
\label{sec:data_sourcing}

We first curate a high-quality set of single-turn LFQA instances. Each instance is a QA pair $(q_i, g_i)$, where $q_i \in \mathcal{Q}$ is a knowledge-intensive question and $g_i \in \mathcal{G}$ is its ground-truth answer. To ensure reliability, each $g_i$ is supported by an authoritative evidence set $\mathcal{E}_i=\{e_{i1},\dots,e_{iK_i}\}$ extracted from trusted sources such as official websites or expert-curated documents. 

Our benchmark focuses on three knowledge-intensive domains: medicine, finance, and law, and draws data from three sources: (i) prior LFQA benchmarks, (ii) authoritative financial-legal FAQs, and (iii) finance-related trending topics. In the medical domain, we include all 201 labeled QA pairs from the K-QA benchmark~\citep{manes2024k}, after removing redundant content to align with our task definition. For the legal domain, we collect 116 QA pairs from the SEC FAQ repository\footnote{\url{https://www.sec.gov/answers/faqs.htm}} and 184 QA pairs from the policy-focused subset of FinTextQA~\citep{chen2024fintextqa}, yielding 300 legal QA pairs after filtering out trivial single-point answers and repairing missing jurisdictional context. For the financial domain, we sample finance-related trending topics from Google Trends\footnote{\url{https://trends.google.com/trends/}} to broaden topic coverage. We then categorize them and construct 300 QA pairs through manual annotation with authoritative references, including their official website or encyclopedia verified by human experts.

In total, this process yields \textbf{801 high-quality single-turn LFQA instances} spanning three domains: finance (300), law (300), and medicine (201). Since many instances are related to multiple domains, we assign each instance to its \textbf{primary domain} for reporting. To facilitate fine-grained evaluation, we decomposed all ground truth answers into atomic facts, where $\sum_{i} |\mathcal{F}_i| = 4467$. Detailed annotation procedures and additional statistics are provided in Appendix~\ref{app:annotation_detail_single_turn}.

\vspace{-3pt}
\subsection{Question Sequence Generation}
\label{sec:multi_turn_generation}

To mirror real conversational patterns, we analyze the ShareGPT-Chinese-English-90k dataset~\citep{ShareGPT-Chinese-English-90k} and find the following distribution for knowledge-intensive dialogues up to five turns: 2-turn (38.5\%), 3-turn (38.2\%), 4-turn (14.0\%), and 5-turn (5.4\%). Since dialogues longer than 5 turns occur at a negligible rate ($\leq 5\%$), we merge them into 5 turns and set the maximum dialogue length to $N_{\max}=5$. In our benchmark, dialogue lengths are drawn from this empirical distribution of 2--5 turns (37.45\%, 37.45\%, 14.98\%, and 10.11\%, respectively).

For each single-turn question $q_j \in \mathcal{Q}$, we construct a multi-turn question sequence $\mathbf{q}^{(d)} = (q^{(d)}_1, \dots, q^{(d)}_{N_d})$ with $N_d \in \{2,\dots,5\}$ and $q^{(d)}_{N_d}=q_j$. The preceding turns $q^{(d)}_{1:N_d-1}$ are drafted with diverse lead-in patterns and then inspected and rewritten by human annotators to maintain a realistic conversational flow and eliminate redundant follow-up questions about the same issue. In practice, the model first generates the initial $2$--$3$-turn expansions, and human annotators then review and rewrite them, as well as extend a subset into $4$--$5$-turn sequences while validating shorter dialogues. We provide more details in Appendix~\ref{app:annotation_detail_expansion_templates}.

During evaluation, we evaluate models using a dynamic setting where the model self-generates its own dialogue history. Specifically, for a given question sequence $\mathbf{q}^{(d)}$, the history for the final turn, $H^{(d)}_{N_d}$, is constructed by recursively generating each intermediate answer:
$
\label{eq:history_generation}
a^{(d)}_t = \mathcal{M}\bigl(
q^{(d)}_1, a^{(d)}_1, \dots,
q^{(d)}_{t-1}, a^{(d)}_{t-1}, q^{(d)}_t
\bigr), \text{ for } t = 1, \dots, N_d - 1 .
$
Finally, the model generates the targeted answers $a^{(d)}_{N_d} = \mathcal{M}(H^{(d)}_{N_d}, q^{(d)}_{N_d})$ for MT-LFQA. Appendix~\ref{app:history-substitution} further validates the necessity of the dynamic setting.

\subsection{Evaluation Framework}
\label{sec:evaluation_framework}
To systematically evaluate model performance in MT-LFQA, we introduce a comprehensive evaluation framework, as shown in Figure~\ref{fig:3_evaluation}, a two-stage automated pipeline to assess factual alignment, and a three-dimensional metric suite to quantify model performance. The evaluation focuses on the final-turn response ($a_j$) in a dialogue, measuring it against the ground-truth answer ($g_j$) and the ground-truth must-have facts set $\mathcal{F}_j$.

\subsubsection{Two-Stage Evaluation Pipeline}
\label{sec:pipeline}

The core of our framework is a two-stage, NLI-based evaluation pipeline, which is designed to assess factual consistency at a fine-grained level. In the first stage, we employ \textbf{Qwen2.5-32B-Instruct}~\citep{team2024qwen2} as a \textbf{decomposer} to break down long-form answers into minimal, self-contained factual units. For the ground-truth answers ($g_j$), this process yields a set of atomic facts ($\mathcal{F}_j$), which subsequently undergoes manual verification and refinement to establish the gold standard. During testing, the model-generated answer ($a_j$) is dynamically decomposed into a corresponding set of atomic statements ($\mathcal{S}_j$). In the second stage, we use \textbf{Qwen2.5-14B-Instruct}~\citep{team2024qwen2} as the \textbf{NLI-based evaluator} to assess the factual consistency between ground-truth and generated answers. To avoid the quadratic computational complexity, $O(|\mathcal{S}_j| \cdot |\mathcal{F}_j|)$, of an exhaustive comparison between atomic statement sets, we adopt an efficient symmetric approach. Completeness is measured by judging each gold-standard fact $f \in \mathcal{F}_j$ against the full model-generated answer $a_j$. Conversely, correctness is measured by judging each generated statement $s \in \mathcal{S}_j$ against the full ground-truth answer $g_j$. The evaluator classifies each pair as \textbf{Entailment}, \textbf{Contradiction}, or \textbf{Neutral}. Accordingly, we define $\mathcal{S}_j^{+/-}=\{s\in\mathcal{S}_j:\mathrm{NLI}(s,g_j)=\textsc{Entailment/Contradiction}\}$ and
$\mathcal{F}_j^{+/-}=\{f\in\mathcal{F}_j:\mathrm{NLI}(f,a_j)=\textsc{Entailment/Contradiction}\}$.

We further validate the reliability of our evaluation pipeline through human annotation on 100 generated answers sampled across representative LLMs and dialogue settings. The automated decomposition achieves high fidelity to human annotations (18.1\% Symmetric Mean Absolute Percentage Error (SMAPE); 5.9\% omission rate), while the NLI-based judgment stage attains an F1-score of 83.6\% against majority-voted human labels on 1,687 evaluation pairs. Further details are available in Appendix~\ref{app:annotation_detail_evaluation} and Appendix~\ref{app:prompts}, respectively.

\subsubsection{Three-Dimensional Metric Framework}
\label{sec:metrics}
The NLI judgments are aggregated into a suite of metrics organized under three distinct dimensions.
\vspace{-5pt}
\paragraph{Factuality}
This dimension quantifies the correctness and completeness of the must-have facts provided. It is based on \textbf{Factual Precision} ($\mathbf{P}_f$), the fraction of generated statements that are entailed, and \textbf{Factual Recall} ($\mathbf{R}_f$), the fraction of ground-truth facts covered. These are combined into the \textbf{Factual F1} ($\mathbf{S}_f$). For compactness, we use $\langle x_j\rangle_{\mathcal D}$ to denote $\frac{1}{|\mathcal D|}\sum_{j\in\mathcal D}x_j$.
\vspace{-5pt}
\begin{equation*}
\begin{aligned}
\mathbf{R}_f&=\left\langle \tfrac{|\mathcal{F}_j^+|}{|\mathcal{F}_j|} \right\rangle_{\mathcal D},\enspace
\mathbf{P}_f=\left\langle \tfrac{|\mathcal{S}_j^+|}{|\mathcal{S}_j|} \right\rangle_{\mathcal D},\\
\mathbf{S}_f&=
\left\langle
\tfrac{2p_f^{(j)}r_f^{(j)}}{p_f^{(j)}+r_f^{(j)}}
\right\rangle_{\mathcal D}.
\end{aligned}
\end{equation*}

\vspace{-5pt}
\paragraph{Reliability (Factual Hallucination)} This dimension measures the extent of factual hallucination. Analogous to factuality, it is quantified using the \textbf{False Claim Rate} ($\mathbf{P_{fc}}$), the fraction of generated statements that are contradicted, and the \textbf{Misrepresentation Rate} ($\mathbf{R_m}$), the fraction of ground-truth facts contradicted by the ($a_j$). These are unified into the \textbf{Hallucination F1} ($\mathbf{S_h}$).
\vspace{-5pt}
\begin{equation*}
\begin{aligned}
\mathbf{R}_m&=\left\langle \tfrac{|\mathcal{F}_j^-|}{|\mathcal{F}_j|} \right\rangle_{\mathcal D},\enspace
\mathbf{P}_{fc}=\left\langle \tfrac{|\mathcal{S}_j^-|}{|\mathcal{S}_j|} \right\rangle_{\mathcal D},\\
\mathbf{S}_h&=
\left\langle
\tfrac{2p_{fc}^{(j)}r_m^{(j)}}{p_{fc}^{(j)}+r_m^{(j)}}
\right\rangle_{\mathcal D}.
\end{aligned}
\end{equation*}

\vspace{-10pt}
\paragraph{Information Delivery Efficiency} This dimension assesses the utility of the model's response by measuring the token cost of conveying information. This provides an intuitive measure of efficiency, as users directly interact with the token count. We report \textbf{$\mathbf{D_f}$}, the average tokens per correctly entailed fact, \textbf{$\mathbf{D_h}$}, the average tokens per contradicted fact.
\vspace{-2pt}
\begin{equation*}
\mathbf{D}_f=\langle \tfrac{T(a_j)}{|\mathcal{F}_j^+|} \rangle_{\mathcal D},\quad
\mathbf{D}_h=\langle \tfrac{T(a_j)}{|\mathcal{F}_j^-|} \rangle_{\mathcal D},
\end{equation*}

\vspace{-2pt}

where $T(a_j)$ is the token length of $a_j$. If the denominator of any term equals zero, i.e. $|\mathcal{F}_j^+|=0$, $|\mathcal{F}_j^-|=0$, that term is estimated by $\max\limits_{k \in \mathcal{D}} \tfrac{T(a_k)}{|\cdot|}$ of the corresponding metric. 

\begin{figure*}[tbp]
    \centering
    \begin{subfigure}[t]{0.49\linewidth}
        \centering
        \includegraphics[width=\linewidth]{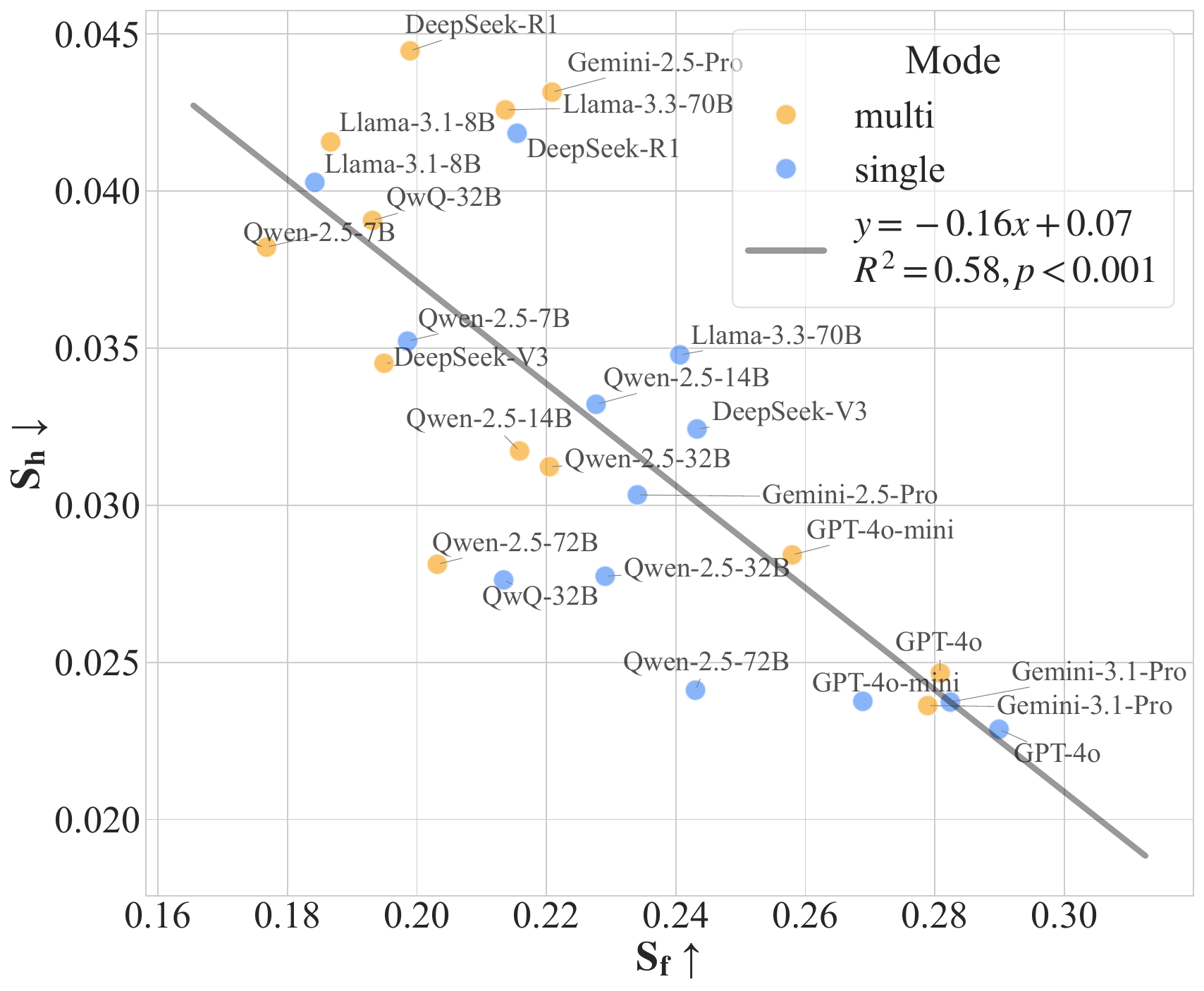}
        \vspace{-15pt}
        \caption{Factual capability}
        \label{fig:4_f1_vs_hall}
    \end{subfigure}
    \begin{subfigure}[t]{0.49\linewidth}
        \centering
        \includegraphics[width=\linewidth]{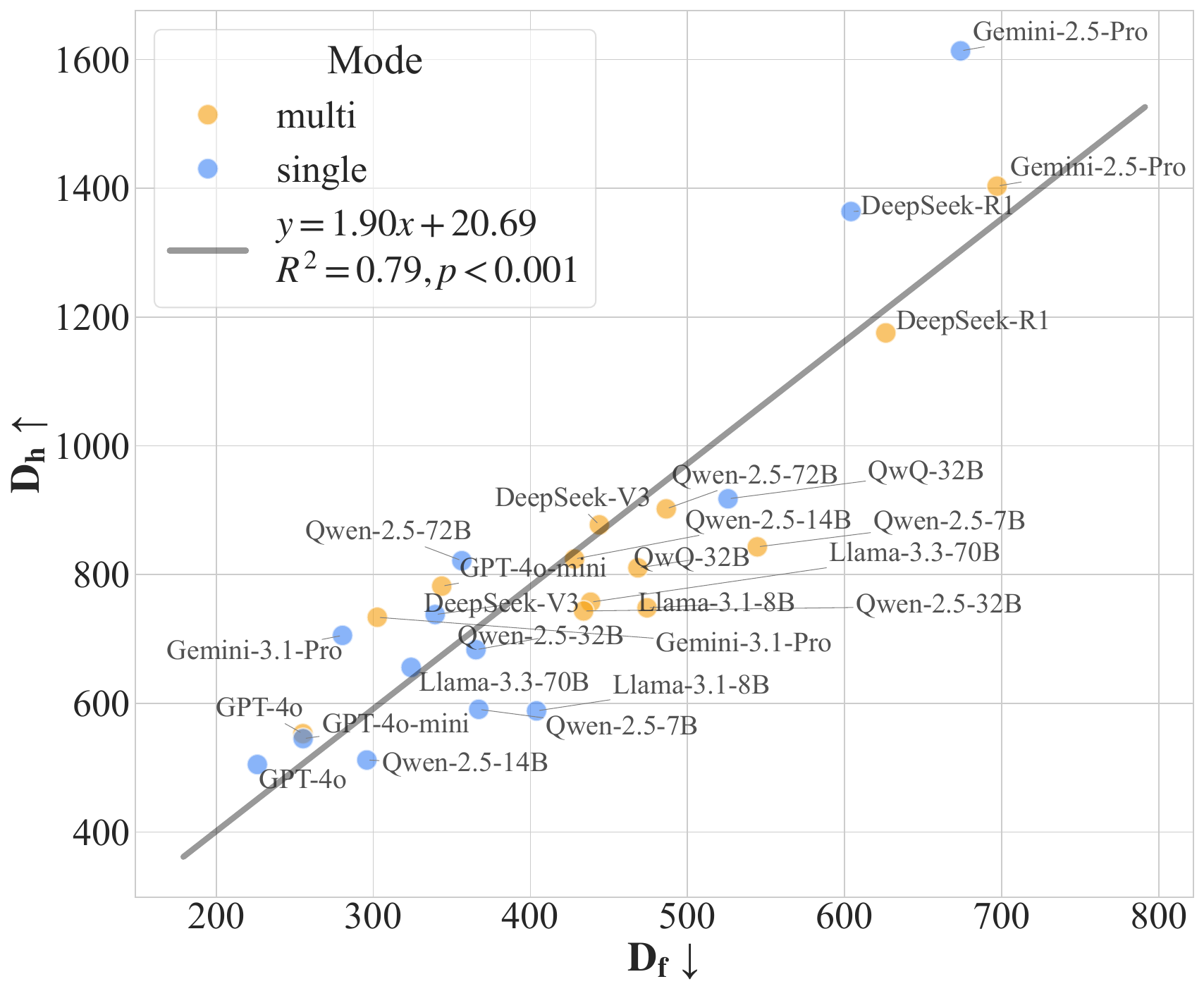}
        \vspace{-15pt}
        \caption{Information delivery efficiency}
        \label{fig:4_eff_vs_eff}
    \end{subfigure}
    \medskip\vspace{-10pt}
\caption{LLMs' performance on factual capability and information delivery efficiency.}
\label{fig:4_all_performance_plots}
\vspace{-10pt}
\end{figure*}

\section{Experiments}

\subsection{Experimental Setup}
\label{sec:experimental_setup}

We evaluate a set of popular LLMs, including the DeepSeek series: DeepSeek-V3-0324, denoted as \textbf{DeepSeek-V3}~\citep{liu2024deepseek} and DeepSeek-R1-0528, denoted as \textbf{DeepSeek-R1}~\citep{guo2025deepseek}, OpenAI’s GPT models~\citep{achiam2023gpt}: gpt-4o-mini-2024-07-18, denoted as \textbf{GPT-4o mini}, and gpt-4o-2024-08-06, denoted as \textbf{GPT-4o}, Meta’s Llama family~\citep{touvron2023llama}: Llama-3.1-8B-Instruct, denoted as \textbf{Llama-3.1-8B} and Llama-3.3-70B-Instruct, denoted as \textbf{Llama-3.3-70B}, the Qwen family~\citep{team2024qwen2}: Qwen-2.5-7B/14B/32B/72B-Instruct, denoted collectively as \textbf{Qwen-2.5-7B/14B/32B/72B}, as well as \textbf{QwQ-32B}. We also include \textbf{Gemini-3.1-Pro}~\citep{gemini31pro2026}, one of the latest state-of-the-art models, together with its predecessor \textbf{Gemini-2.5-Pro}~\citep{comanici2025gemini}.
Token counts are computed using tiktoken's GPT-4o tokeniser.\footnote{\url{https://github.com/openai/tiktoken}} Local model inference and automated evaluation are conducted on NVIDIA A800 GPUs, while API-based models are queried through their providers' endpoints. See detailed experiment settings in Appendix~\ref{exp setting}.

\begin{figure*}[t]
    \centering

    \begin{subfigure}[t]{0.7314\linewidth}
        \centering
        \includegraphics[
            width=\linewidth,
            trim={0bp 0bp 544.8bp 0bp},
            clip
        ]{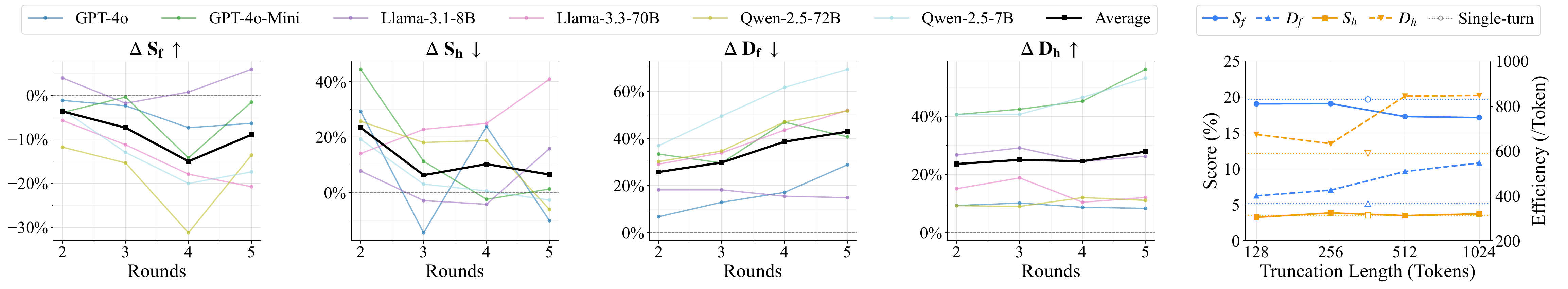}
        \caption{Effect of dialogue turns.}
        \vspace{-5pt}
        \label{fig:4_relative_diff_summary}
    \end{subfigure}
    \hfill
    \begin{subfigure}[t]{0.2386\linewidth}
        \centering
        \includegraphics[
            width=\linewidth,
            trim={1670bp 0bp 0bp 0bp},
            clip
        ]{figures/assets/dialogue_length_effects_ab.pdf}
        \caption{Effect of context length.}
        \label{fig:4_length}
    \end{subfigure}

    \vspace{-5pt}
    \caption{Effects of dialogue length on multi-turn performance across dialogue turns and context length.}
    \vspace{-10pt}
    \label{fig:4_length_effects}
\end{figure*}

\subsection{Main Results}
\label{sec:main_results}
To reveal the relations between different performance dimensions, we map all model performances into scatter plots (Figure~\ref{fig:4_all_performance_plots}). We present four primary metrics in the main text, while the full metric summary with detailed numerical results and additional metric values is provided in Table~\ref{tab:main_results} in Appendix~\ref{app:numerical_results}. A clear observation from the plots is a systematic shift in performance when moving from single-turn to multi-turn settings. Specifically, in Figure~\ref{fig:4_f1_vs_hall}, most models move toward lower Factual F1 ($\mathbf{S_f}$) and higher Hallucination F1 ($\mathbf{S_h}$), indicating weakened factual capability. Similarly, in Figure~\ref{fig:4_eff_vs_eff}, the points generally shift toward higher token cost per correct fact ($\mathbf{D_f}$), while their generated factual errors also become sparser (higher $\mathbf{D_h}$). This pronounced degradation in factuality, coupled with the challenge of efficiently delivering correct information in multi-turn dialogues, highlights the unique challenges posed by the MT-LFQA task. Within this general trend, several model-level patterns are also apparent: proprietary models such as GPT-4o and Gemini-3.1-Pro define the performance frontier, and larger models generally outperform smaller ones within the same family. The newer Gemini-3.1-Pro also shows stronger factual capability than its predecessor Gemini-2.5-Pro, achieving higher $\mathbf{S_f}$ and lower $\mathbf{S_h}$ in both single-turn and multi-turn settings. However, models optimised for Chain-of-Thought (e.g., DeepSeek-R1 compared with DeepSeek-V3, and QwQ-32B compared with Qwen-2.5-32B) show no consistent advantage in factual capability, suggesting that current CoT mechanisms may not directly translate to more factual answers.

The point distributions further reveal two underlying correlations. First, Figure~\ref{fig:4_f1_vs_hall} illustrates a moderate negative correlation between Factual F1 ($\mathbf{S_f}$) and Hallucination F1 ($\mathbf{S_h}$) ($R^2=0.58$), suggesting that as a model's factuality improves, its hallucination tendency generally decreases. This points to a potential strategy for mitigating hallucinations: enhancing the models' intrinsic knowledge. Second, Figure~\ref{fig:4_eff_vs_eff} reveals a strong positive correlation between the token cost per correct fact ($\mathbf{D_f}$) and per contradicted fact ($\mathbf{D_h}$) ($R^2=0.79$), suggesting that most models tend to be uniformly concise or verbose, rather than selectively efficient when conveying correct versus incorrect information. A notable exception appears in the single-turn efficiency plot: Gemini-2.5-Pro achieves an exceptionally high $\mathbf{D_h}$ for its given $\mathbf{D_f}$, meaning that it conveys correct facts relatively efficiently while producing contradicted facts only sparsely. However, this desirable characteristic disappears in the multi-turn setting, where its performance aligns with the general trend. Therefore, a key direction for future research is to design models or strategies that can achieve a low $\mathbf{D_f}$ while simultaneously increasing $\mathbf{D_h}$ in multi-turn dialogue. A detailed, holistic four-quadrant analysis is provided in Appendix~\ref{app:four-quadrant-analysis}.

\subsection{The Source of Multi-turn Degradation}
\label{sec:pd}

The main experiment reveals a systematic degradation when models move from the single-turn to the multi-turn setting. Compared with the single-turn setting, MT-LFQA introduces two additional challenges. First, the model conditions on a longer dialogue history before answering the final question. Second, during multi-turn dialogue, the model may introduce redundant, irrelevant, or incorrect information into its own history. We therefore examine whether the degradation stems from context length or history noise.

\vspace{-5pt}
\paragraph{Effect of dialogue length.}
To isolate the impact of length, we conduct two experiments. \textit{(i) Number of Turns.} We vary the number of dialogue turns and report the relative change of each metric $m \in \{\mathbf{S_f, S_h, D_f, D_h}\}$ against the single-turn baseline, calculated as $\Delta m=(m_{\text{multi}}-m_{\text{single}})/m_{\text{single}}$ (Figure~\ref{fig:4_relative_diff_summary}). \textit{(ii) Context Length.} For each multi-turn instance, we use Qwen-2.5-7B to regenerate the final-turn answer under different token budgets for the dialogue history preceding the final question (Figure~\ref{fig:4_length}). The results show that information delivery efficiency consistently degrades as the context lengthens, suggesting a dilution effect where models become increasingly verbose under longer histories. In contrast, factuality shows only a slight overall decrease, while hallucination exhibits no clear dependence on length. Thus, dialogue length mainly affects information delivery efficiency rather than factual capability.

\begin{figure}[h]
    \centering
    \includegraphics[width=\linewidth, trim=5pt 0pt 5pt 0pt, clip]{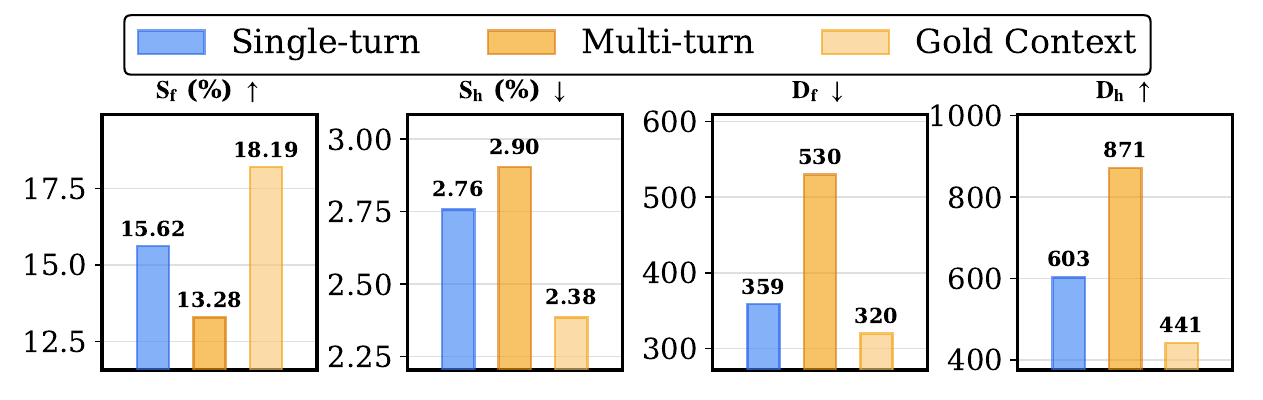}
    \vspace{-20pt}
    \caption{Impact of manually annotated gold context on \textbf{Qwen-2.5-7B}.}
    \label{fig:5_gold_context}
    \vspace{-5pt}
\end{figure}

\vspace{-5pt}
\paragraph{Noise in dialogue history.}
Having shown that dialogue length alone cannot fully explain the factual degradation, we next examine whether the degradation stems from noise accumulated in self-generated dialogue histories. We first test this hypothesis through a gold-context experiment (Figure~\ref{fig:5_gold_context}). Specifically, we randomly select 100 finance QA instances, manually construct evidence-supported answers for all non-final questions with the help of external tools, including search engines, and use these answers to substitute for the model-generated historical assistant responses while keeping the final question unchanged. Compared with both the standard multi-turn and single-turn baselines on the same subset, the gold context achieves better factual performance, lower hallucination, and higher information delivery efficiency, which suggests that conversation context is not inherently harmful: when the dialogue history is low-noise, using it as context can lead to better performance than answering the final question in isolation.

\begin{figure}[h]
    \centering
    \includegraphics[width=\linewidth, trim=5pt 0pt 5pt 0pt, clip]{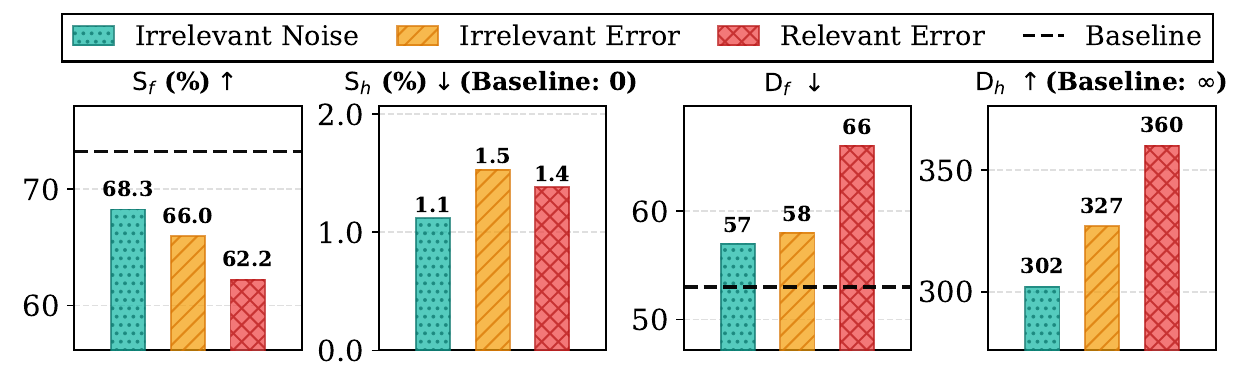}
    \vspace{-20pt}
    \caption{Performance of \textbf{GPT-4o} under different types of controlled dialogue-history noise.}
    \label{fig:5_controlled_noise}
    \vspace{-5pt}
\end{figure}

\begin{table*}[t]
\centering
\small
\vspace{-10pt}
\caption{Results across different domains. \underline{Underlined} values denote improvements over the base model under the same domain and turn setting. White rows indicate single-turn results, while gray rows indicate multi-turn results.}
\setlength{\tabcolsep}{9.5pt}
\renewcommand{\arraystretch}{0.9}

\begin{tabular}{l|cccc|cccc}
\toprule
& \multicolumn{4}{c|}{\textbf{Finance}} 
& \multicolumn{4}{c}{\textbf{Non-finance}} \\
\cmidrule(lr){2-5} \cmidrule(lr){6-9}
\textbf{Model}
& {$\mathbf{S_f}$ (\%) $\uparrow$}
& {$\mathbf{S_h}$ (\%) $\downarrow$}
& {$\mathbf{D_f}$ $\downarrow$}
& {$\mathbf{D_h}$ $\uparrow$}
& {$\mathbf{S_f}$ (\%) $\uparrow$}
& {$\mathbf{S_h}$ (\%) $\downarrow$}
& {$\mathbf{D_f}$ $\downarrow$}
& {$\mathbf{D_h}$ $\uparrow$} \\
\midrule
\rowcolor{white}
\textit{\textbf{Qwen-2.5-7B}}
& 16.50 & 2.29 & 376.19 & 616.21
& 21.87 & 4.26 & 361.73 & 575.38 \\

\rowcolor{gray!15}
& 14.60 & 2.12 & 578.12 & 888.26
& 19.52 & 4.84 & 524.34 & 815.99 \\

\midrule

\rowcolor{white}
\textit{\textbf{Fin-R1}}
& 14.51 & \underline{2.19} & 568.49 & \underline{895.72}
& \underline{22.52} & 4.76 & 487.90 & \underline{831.48} \\

\rowcolor{gray!15}
& 13.39 & 3.53 & \underline{511.55} & 825.01
& \underline{20.08} & 4.85 & \underline{467.26} & 802.78 \\

\midrule

& \multicolumn{4}{c|}{\textbf{Medical}}
& \multicolumn{4}{c}{\textbf{Non-medical}} \\
\cmidrule(lr){2-5} \cmidrule(lr){6-9}
\textbf{Model}
& {$\mathbf{S_f}$ (\%) $\uparrow$}
& {$\mathbf{S_h}$ (\%) $\downarrow$}
& {$\mathbf{D_f}$ $\downarrow$}
& {$\mathbf{D_h}$ $\uparrow$}
& {$\mathbf{S_f}$ (\%) $\uparrow$}
& {$\mathbf{S_h}$ (\%) $\downarrow$}
& {$\mathbf{D_f}$ $\downarrow$}
& {$\mathbf{D_h}$ $\uparrow$} \\
\midrule

\rowcolor{white}
\textit{\textbf{Qwen-2.5-7B}}
& 29.59 & 1.44 & 260.80 & 628.09
& 16.60 & 4.22 & 402.77 & 578.14 \\

\rowcolor{gray!15}
& 27.43 & 2.24 & 368.33 & 870.07
& 14.41 & 4.35 & 603.49 & 834.01 \\

\midrule

\rowcolor{white}
\textit{\textbf{HuatuoGPT}}
& \underline{34.54} & 2.30 & \underline{162.68} & 428.98
& \underline{18.00} & 5.60 & \underline{267.14} & 374.66 \\

\rowcolor{gray!15}
& \underline{33.85} & \underline{1.38} & \underline{143.11} & 349.02
& \underline{18.06} & 5.49 & \underline{226.94} & 308.88 \\
\bottomrule
\end{tabular}

\label{tab:domain_results_all}
\vspace{-10pt}
\end{table*}

To further verify whether the factual degradation is driven by noise in the dialogue history, we then employ a controlled noise-injection protocol (Figure~\ref{fig:5_controlled_noise}). To ensure that the final question itself is answerable by the model, we first curate a \emph{knowledge-verified} baseline by filtering for 85 instances where GPT-4o exhibits good factual mastery in the single-turn setting ($\mathbf{S_f} \ge 60\%, \mathbf{S_h} = 0$). Keeping the final question unchanged, we manipulate \emph{only the content of the previous turn} to construct three distinct history types: \emph{Irrelevant Noise} (correct but unrelated), \emph{Irrelevant Error} (incorrect and unrelated), and \emph{Relevant Error} (incorrect and related). The controlled perturbations reveal a clear degradation pattern: as historical noise becomes more severe, factuality drops monotonically while hallucinations rise sharply. Notably, \emph{relevant errors} induce the most catastrophic degradation, accompanied by a significant deterioration in information delivery efficiency. Complementing these controlled perturbations, we further analyze natural variation in self-generated histories and find that intermediate-turn history quality is positively correlated with final-turn factuality ($\mathbf{S_f}$: Pearson $r=0.36$, $p=0.0002$). We also observe a weaker but statistically significant positive correlation for hallucination ($\mathbf{S_h}$: Pearson $r=0.25$, $p=0.011$), suggesting that noisier intermediate histories are more likely to propagate factual errors into the final response. Together, these analyses show that low-quality dialogue histories, especially relevant misinformation, directly undermine factual grounding and amplify hallucination in MT-LFQA.

\vspace{-1pt}
\section{Mitigation Strategies}
\vspace{-1pt}

\label{sec:mitigation_strategies}

To address the performance degradation in MT-LFQA, we investigate mitigation strategies along two directions. The first direction focuses on \textbf{Knowledge Fortification}, aiming to enhance the model's factual grounding to make it more robust against hallucination and noise. We explore this via leveraging \textbf{extrinsic knowledge} through Retrieval-Augmented Generation (RAG) (Section~\ref{subsec:rag}) and strengthening \textbf{intrinsic knowledge} through domain-specific finetuning (Section~\ref{subsec:finetuning}). The second direction targets \textbf{Active Contextual Denoising}, where we employ \textbf{inference-time interventions} to actively filter and refine the dialogue history, thereby directly attenuating the interference from noisy contexts (Section~\ref{subsec:prompting}).

\subsection{Effect of Domain-Specific Fine-tuning}
\label{subsec:finetuning}

Given the high proportion of medical and financial instances in our benchmark and the prevalence of domain-tuned models in these areas, we examine how domain finetuning influences LFQA and multi-turn dialogue. We evaluate two Qwen-2.5-7B derivatives: \textbf{Fin-R1}\footnote{\url{https://huggingface.co/SUFE-AIFLM-Lab/Fin-R1}}~\citep{liu2025fin} for finance and \textbf{HuatuoGPT}\footnote{\url{https://huggingface.co/FreedomIntelligence/HuatuoGPT-o1-7B}}~\citep{chen2024huatuogpt} for medicine. Table~\ref{tab:domain_results_all} indicates that HuatuoGPT improves factuality and correct-information delivery efficiency in all reported settings, with larger factuality gains in the multi-turn and medical-domain comparisons. Its reliability results are mixed: $\mathbf{S_h}$ improves only for medical multi-turn responses, while $\mathbf{D_h}$ decreases throughout. Fin-R1 shows no consistent advantage in the finance subset: its $\mathbf{S_f}$ is lower in both turn settings, and improvements in the remaining metrics vary by setting. Overall, domain-tuned models can improve particular dimensions, but specialization alone does not guarantee a consistent in-domain advantage.

\subsection{Effect of RAG Strategies}
\label{subsec:rag}
To evaluate the effect of extrinsic knowledge, we test four RAG strategies: \textbf{Base} (retrieval at the final turn using the last-turn query), \textbf{Last} (retrieval at the final turn using the full dialogue), \textbf{Rounds} (retrieval at each turn using the current query), and \textbf{All} (retrieval at each turn using the full previous history). In the single-turn setting, since there is no dialogue history, only the Base settings are available. Detailed settings are provided in Appendix~\ref{exp setting}.

As shown in Table~\ref{tab:5_rag}, the baseline model without RAG confirms the performance degradation in its factual capacity ($\mathbf{S_f}$ and $\mathbf{S_h}$) from single-turn to multi-turn settings. The introduction of RAG provides a substantial improvement in factual capacity for both settings. Among these strategies, \textbf{Rounds} is the best strategy in the multi-turn setting, achieving the highest factuality ($\mathbf{S_f}$), the fewest hallucinations ($\mathbf{S_h}$), and the best correct information delivery efficiency ($\mathbf{D_f}$). Notably, this strategy is so effective that it reverses the performance degradation on some dimension: it enables the model to achieve a higher factuality score in the multi-turn setting than even the RAG-enhanced single-turn baseline. This highlights RAG's capacity not merely to mitigate noise, but to actively leverage the multi-turn structure by grounding each step with factual evidence, thus preventing the accumulation of noise that characterizes the non-RAG setting. In contrast, the \textbf{Last} and \textbf{All} strategies, which use the full dialogue history as the queries, are more likely to accumulate noise and therefore underperform.

\begin{table}[t]
\centering
\small
\setlength{\tabcolsep}{7pt}  %
\renewcommand{\arraystretch}{1} %
\caption{Comparison of w/o versus w/ RAG on \textbf{Qwen-2.5-7B}. \textbf{White rows} indicate \textbf{single-turn settings}, while \textbf{gray rows} indicate \textbf{multi-turn settings}. \textbf{Bold} numbers highlight the best results of the multi-turn RAG setting. \underline{Underlined} values denote improvements over the baseline under the same turn setting.}
\vspace{-5pt}
\label{tab:5_rag}
\begin{tabular}{l cccc}
\toprule
Setting & {$\mathbf{S_f}$ (\%) $\uparrow$} & {$\mathbf{S_h}$ (\%) $\downarrow$} & {$\mathbf{D_f}$ $\downarrow$} & {$\mathbf{D_h}$ $\uparrow$} \\
\midrule
\multicolumn{5}{l}{\textbf{w/o RAG}} \\
\rowcolor{white} Original & 19.86 & 3.52 & 367.15 & 590.67 \\
\rowcolor{gray!15} Original & 17.68 & 3.82 & 544.48 & 843.06 \\
\midrule
\multicolumn{5}{l}{\textbf{w/ RAG}} \\
\rowcolor{white} Base & \underline{40.77} & \underline{1.87} & \underline{193.70} & 575.75 \\
\rowcolor{gray!15} Base & \underline{40.19} & \underline{2.27} & \underline{304.82} & \underline{\textbf{888.61}} \\
\rowcolor{gray!15} Last & \underline{39.70} & \underline{2.58} & \underline{251.64} & 607.05 \\
\rowcolor{gray!15} Rounds & \underline{\textbf{41.01}} & \underline{\textbf{2.01}} & \underline{\textbf{230.65}} & 680.25 \\
\rowcolor{gray!15} All & \underline{37.82} & \underline{2.35} & \underline{264.58} & 564.97 \\
\bottomrule
\end{tabular}
\vspace{-5pt}
\end{table}

\subsection{Effect of Prompt-Based Strategies}
\label{subsec:prompting}

Based on previous analysis, we explored various prompt-based mitigation strategies to address the degradation. We began with a straightforward intervention, explicitly instructing the model: ``You may refer to the previous conversation for context, but rely primarily on your internal knowledge.'' This \textit{Simple Prompt} yielded slight improvements in information delivery efficiency and hallucination control but did not improve factuality, suggesting that explicit instructions alone are insufficient to filter out noise. Moving to a structural intervention, \textit{Self-Refinement}~\cite{madaan2023self} successfully reduced hallucinations; however, it induced excessive conservatism, resulting in a significant trade-off where factuality notably declined.

In contrast, context-restructuring methods are substantially more effective. In both \textit{Summarization} and \textit{Selection} strategies, the model processes the preceding dialogue history, either by compressing or filtering it, and feeds the refined context together with the current question to generate the response. These approaches successfully counteract the negative impacts of excessive or noisy context. By presenting a concise and focused context, both strategies alleviate the model's ``conservative'' constraint, thereby restoring information delivery efficiency and fostering the generation of more factual content. \textit{Context Selection} performs best: by removing noise rather than merely compressing it, it yields the strongest recovery in factual capability.

\begin{table}[t]
\small
\setlength{\tabcolsep}{5pt}  %
\renewcommand{\arraystretch}{1} %
\caption{Performance comparison of \textbf{Qwen2.5-32B} with prompt-based mitigation strategies. \textbf{Gray and white rows} indicate as above. \underline{Underlined} values denote improvements to the multi-turn baseline.}
\vspace{-5pt}
\captionsetup{width=\linewidth}
\centering
\begin{tabular}{lcccc}
\toprule
\textbf{Strategy} & {$\mathbf{S_f}$ (\%) $\uparrow$} & {$\mathbf{S_h}$ (\%) $\downarrow$} & {$\mathbf{D_f}$ $\downarrow$} & {$\mathbf{D_h}$ $\uparrow$} \\
\midrule
\multicolumn{5}{l}{\textbf{Baselines}} \\
\rowcolor{white} Original & 22.91 & 2.77 & 365 & 683 \\
\rowcolor{gray!15} Original & 22.05 & 3.12 & 434 & 743 \\
\midrule
\multicolumn{5}{l}{\textbf{Interventions}} \\
\rowcolor{gray!15} + Simple Prompt & 21.89 & \underline{3.05} & \underline{417} & \underline{764} \\
\rowcolor{gray!15} + Summarization & \underline{24.05} & 3.28 & \underline{259} & 419 \\
\rowcolor{gray!15} + Selection & \underline{24.10} & 3.19 & \underline{319} & 590 \\
\rowcolor{gray!15} + Self-Refine & 19.84 & \underline{2.90} & 525 & \underline{899} \\
\bottomrule
\end{tabular}
\vspace{-10pt}
\label{tab:advanced_mitigation}
\end{table}

\section{Conclusion}

In this work, we introduce \textbf{KnowMT-Bench}, a benchmark for evaluating knowledge-intensive \textbf{multi-turn long-form QA (MT-LFQA)}. Experiments across diverse LLMs reveal consistent multi-turn degradation in both \textit{factual capability} and \textit{information-delivery efficiency}: the efficiency decline is strongly associated with dialogue length, while factual capability degradation is mainly driven by accumulated \textit{contextual noise}, especially self-generated misinformation. We further examine mitigation strategies, showing domain-dependent gains from domain-specific finetuning and more consistent improvements from \textbf{active context denoising} and \textbf{RAG}. These findings motivate two concrete avenues for future model development in knowledge-intensive applications: (i) inference-time mechanisms that select and compress dialogue history to denoise context, and (ii) training-time objectives that preserve factual efficiency, decoupling verbosity from factual capacity and reliability.

\section*{Limitations}

KnowMT-Bench provides an early systematic evaluation of knowledge-intensive MT-LFQA, but it still has several limitations. First, our benchmark mainly focuses on knowledge-intensive domains, including medicine, finance, and law, while its coverage of other application scenarios remains limited. Future work may further extend the benchmark to broader domains and more diverse user needs. Second, our current evaluation adopts a dynamic self-generated dialogue setting to approximate multi-turn interactions. While this setting captures an important aspect of real-world use, future work can further develop more realistic interactive settings that better reflect how humans communicate with LLMs in open-ended conversation. Third, although our NLI-based evaluator is validated with human annotations and cross-model checks, automated evaluation may still contain residual biases, including verbosity bias, position effects, or domain-specific factual blind spots in medicine, finance, and law. Future work can further strengthen the evaluation with broader expert audits and evaluator ensembles.

\section*{Ethical Considerations}

KnowMT-Bench is built from publicly available and evidence-backed sources, and is designed to avoid collecting or exposing personal or sensitive user information. Since the benchmark covers high-stakes domains such as medicine, finance, and law, we emphasize that it is intended for research and evaluation purposes only, rather than for providing professional medical, financial, or legal advice. The benchmark highlights factual degradation and hallucination risks in multi-turn settings, which may help researchers develop more reliable and knowledge-grounded dialogue systems.

\section*{Acknowledgments}

This work was supported by the National Natural Science Foundation of China (Grant No. 62506318); Guangdong Provincial Department of Education Project (Grant No. 2024KQNCX028); Scientific Research Projects for the Higher-educational Institutions (Grant No. 2024312096), Education Bureau of Guangzhou Municipality; and Guangzhou-HKUST(GZ) Joint Funding Program (Grant No. 2025A03J3957), Education Bureau of Guangzhou Municipality.

Generative AI tools, including ChatGPT and Gemini, were used solely to assist with language polishing and improving the clarity of the manuscript. All scientific content and conclusions were developed and verified by the authors.

\bibliography{references}

\begin{thebibliography}{52}
\providecommand{\natexlab}[1]{#1}

\bibitem[{Achiam et~al.(2023)Achiam, Adler, Agarwal, Ahmad, Akkaya, Aleman,
  Almeida, Altenschmidt, Altman, Anadkat et~al.}]{achiam2023gpt}
Josh Achiam, Steven Adler, Sandhini Agarwal, Lama Ahmad, Ilge Akkaya,
  Florencia~Leoni Aleman, Diogo Almeida, Janko Altenschmidt, Sam Altman,
  Shyamal Anadkat, and 1 others. 2023.
\newblock Gpt-4 technical report.
\newblock \emph{arXiv preprint arXiv:2303.08774}.

\bibitem[{Adlakha et~al.(2022)Adlakha, Dhuliawala, Suleman, de~Vries, and
  Reddy}]{adlakha2022topiocqa}
Vaibhav Adlakha, Shehzaad Dhuliawala, Kaheer Suleman, Harm de~Vries, and Siva
  Reddy. 2022.
\newblock Topiocqa: Open-domain conversational question answering with topic
  switching.
\newblock \emph{Transactions of the Association for Computational Linguistics},
  10:468--483.

\bibitem[{Bai et~al.(2024)Bai, Liu, Bu, He, Liu, Zhou, Lin, Su, Ge, Zheng
  et~al.}]{bai2024mt}
Ge~Bai, Jie Liu, Xingyuan Bu, Yancheng He, Jiaheng Liu, Zhanhui Zhou, Zhuoran
  Lin, Wenbo Su, Tiezheng Ge, Bo~Zheng, and 1 others. 2024.
\newblock Mt-bench-101: A fine-grained benchmark for evaluating large language
  models in multi-turn dialogues.
\newblock In \emph{Proceedings of the 62nd Annual Meeting of the Association
  for Computational Linguistics (Volume 1: Long Papers)}, pages 7421--7454.

\bibitem[{Banatt et~al.(2024)Banatt, Cheng, Vaidyanath, and
  Hwu}]{banatt2024wilt}
Eryk Banatt, Jonathan Cheng, Skanda Vaidyanath, and Tiffany Hwu. 2024.
\newblock Wilt: A multi-turn, memorization-robust inductive logic benchmark for
  llms.
\newblock \emph{arXiv preprint arXiv:2410.10998}.

\bibitem[{Campos et~al.(2020)Campos, Otegi, Soroa, Deriu, Cieliebak, and
  Agirre}]{campos2020doqa}
Jon~Ander Campos, Arantxa Otegi, Aitor Soroa, Jan Deriu, Mark Cieliebak, and
  Eneko Agirre. 2020.
\newblock Doqa--accessing domain-specific faqs via conversational qa.
\newblock \emph{arXiv preprint arXiv:2005.01328}.

\bibitem[{{CFP Board}(2020)}]{cfp2020practicestandards}
{CFP Board}. 2020.
\newblock Practice standards reference guide.
\newblock
  \url{https://www.cfp.net/ethics/compliance-resources/2020/11/practice-standards-reference-guide}.
\newblock Certified Financial Planner Board of Standards, Inc.

\bibitem[{Chen et~al.(2024{\natexlab{a}})Chen, Chen, Liu, Jiang, and
  Wang}]{chen2024humans}
Guiming~Hardy Chen, Shunian Chen, Ziche Liu, Feng Jiang, and Benyou Wang.
  2024{\natexlab{a}}.
\newblock Humans or llms as the judge? a study on judgement biases.
\newblock \emph{arXiv preprint arXiv:2402.10669}.

\bibitem[{Chen et~al.(2024{\natexlab{b}})Chen, Zhou, Hua, Xin, Chen, Li, Zhu,
  and Liang}]{chen2024fintextqa}
Jian Chen, Peilin Zhou, Yining Hua, Loh Xin, Kehui Chen, Ziyuan Li, Bing Zhu,
  and Junwei Liang. 2024{\natexlab{b}}.
\newblock Fintextqa: A dataset for long-form financial question answering.
\newblock In \emph{Proceedings of the 62nd Annual Meeting of the Association
  for Computational Linguistics (Volume 1: Long Papers)}, pages 6025--6047.

\bibitem[{Chen et~al.(2024{\natexlab{c}})Chen, Cai, Ji, Wang, Liu, Wang, Hou,
  and Wang}]{chen2024huatuogpt}
Junying Chen, Zhenyang Cai, Ke~Ji, Xidong Wang, Wanlong Liu, Rongsheng Wang,
  Jianye Hou, and Benyou Wang. 2024{\natexlab{c}}.
\newblock Huatuogpt-o1, towards medical complex reasoning with llms.
\newblock \emph{arXiv preprint arXiv:2412.18925}.

\bibitem[{Chen et~al.(2022)Chen, Li, Smiley, Ma, Shah, and
  Wang}]{chen2022convfinqa}
Zhiyu Chen, Shiyang Li, Charese Smiley, Zhiqiang Ma, Sameena Shah, and
  William~Yang Wang. 2022.
\newblock Convfinqa: Exploring the chain of numerical reasoning in
  conversational finance question answering.
\newblock \emph{arXiv preprint arXiv:2210.03849}.

\bibitem[{Choi et~al.(2018)Choi, He, Iyyer, Yatskar, Yih, Choi, Liang, and
  Zettlemoyer}]{choi2018quac}
Eunsol Choi, He~He, Mohit Iyyer, Mark Yatskar, Wen-tau Yih, Yejin Choi, Percy
  Liang, and Luke Zettlemoyer. 2018.
\newblock Quac: Question answering in context.
\newblock \emph{arXiv preprint arXiv:1808.07036}.

\bibitem[{Comanici et~al.(2025)Comanici, Bieber, Schaekermann, Pasupat,
  Sachdeva, Dhillon, Blistein, Ram, Zhang, Rosen et~al.}]{comanici2025gemini}
Gheorghe Comanici, Eric Bieber, Mike Schaekermann, Ice Pasupat, Noveen
  Sachdeva, Inderjit Dhillon, Marcel Blistein, Ori Ram, Dan Zhang, Evan Rosen,
  and 1 others. 2025.
\newblock Gemini 2.5: Pushing the frontier with advanced reasoning,
  multimodality, long context, and next generation agentic capabilities.
\newblock \emph{arXiv preprint arXiv:2507.06261}.

\bibitem[{Duan et~al.(2024)Duan, Wei, Wang, Liu, Fang, Zhang, Lin, and
  Chen}]{duan2024botchat}
Haodong Duan, Jueqi Wei, Chonghua Wang, Hongwei Liu, Yixiao Fang, Songyang
  Zhang, Dahua Lin, and Kai Chen. 2024.
\newblock Botchat: Evaluating llms’ capabilities of having multi-turn
  dialogues.
\newblock In \emph{Findings of the Association for Computational Linguistics:
  NAACL 2024}, pages 3184--3200.

\bibitem[{Fan et~al.(2025)Fan, Ni, Merane, Salimbeni, Tian, Hermstr{\"u}wer,
  Huang, Akhtar, Geering, Dreyer et~al.}]{fan2025lexam}
Yu~Fan, Jingwei Ni, Jakob Merane, Etienne Salimbeni, Yang Tian, Yoan
  Hermstr{\"u}wer, Yinya Huang, Mubashara Akhtar, Florian Geering, Oliver
  Dreyer, and 1 others. 2025.
\newblock Lexam: Benchmarking legal reasoning on 340 law exams.
\newblock \emph{arXiv preprint arXiv:2505.12864}.

\bibitem[{Fan et~al.(2024)Fan, Chen, Hu, and Liu}]{fan2024fairmt}
Zhiting Fan, Ruizhe Chen, Tianxiang Hu, and Zuozhu Liu. 2024.
\newblock Fairmt-bench: Benchmarking fairness for multi-turn dialogue in
  conversational llms.
\newblock \emph{arXiv preprint arXiv:2410.19317}.

\bibitem[{Feng et~al.(2021)Feng, Patel, Wan, and Joshi}]{feng2021multidoc2dial}
Song Feng, Siva~Sankalp Patel, Hui Wan, and Sachindra Joshi. 2021.
\newblock Multidoc2dial: Modeling dialogues grounded in multiple documents.
\newblock \emph{arXiv preprint arXiv:2109.12595}.

\bibitem[{Feng et~al.(2020)Feng, Wan, Gunasekara, Patel, Joshi, and
  Lastras}]{feng2020doc2dial}
Song Feng, Hui Wan, Chulaka Gunasekara, Siva~Sankalp Patel, Sachindra Joshi,
  and Luis~A Lastras. 2020.
\newblock doc2dial: A goal-oriented document-grounded dialogue dataset.
\newblock \emph{arXiv preprint arXiv:2011.06623}.

\bibitem[{Fu et~al.(2023)Fu, Laskar, Chen, and TN}]{fu2023large}
Xue-Yong Fu, Md~Tahmid~Rahman Laskar, Cheng Chen, and Shashi~Bhushan TN. 2023.
\newblock Are large language models reliable judges? a study on the factuality
  evaluation capabilities of llms.
\newblock \emph{arXiv preprint arXiv:2311.00681}.

\bibitem[{{Google DeepMind}(2026)}]{gemini31pro2026}
{Google DeepMind}. 2026.
\newblock \href
  {https://storage.googleapis.com/deepmind-media/Model-Cards/Gemini-3-1-Pro-Model-Card.pdf}
  {{Gemini 3.1 Pro Model Card}}.
\newblock Technical report, Google DeepMind.

\bibitem[{Guo et~al.(2025)Guo, Yang, Zhang, Song, Zhang, Xu, Zhu, Ma, Wang, Bi
  et~al.}]{guo2025deepseek}
Daya Guo, Dejian Yang, Haowei Zhang, Junxiao Song, Ruoyu Zhang, Runxin Xu,
  Qihao Zhu, Shirong Ma, Peiyi Wang, Xiao Bi, and 1 others. 2025.
\newblock Deepseek-r1: Incentivizing reasoning capability in llms via
  reinforcement learning.
\newblock \emph{arXiv preprint arXiv:2501.12948}.

\bibitem[{Hackenburg et~al.(2025)Hackenburg, Tappin, Hewitt, Saunders, Black,
  Lin, Fist, Margetts, Rand, and Summerfield}]{hackenburg2025levers}
Kobi Hackenburg, Ben~M Tappin, Luke Hewitt, Ed~Saunders, Sid Black, Hause Lin,
  Catherine Fist, Helen Margetts, David~G Rand, and Christopher Summerfield.
  2025.
\newblock The levers of political persuasion with conversational ai.
\newblock \emph{arXiv preprint arXiv:2507.13919}.

\bibitem[{He et~al.(2024)He, Jin, Wang, Bi, Mandyam, Zhang, Zhu, Li, Xu, Lv
  et~al.}]{he2024multi}
Yun He, Di~Jin, Chaoqi Wang, Chloe Bi, Karishma Mandyam, Hejia Zhang, Chen Zhu,
  Ning Li, Tengyu Xu, Hongjiang Lv, and 1 others. 2024.
\newblock Multi-if: Benchmarking llms on multi-turn and multilingual
  instructions following.
\newblock \emph{arXiv preprint arXiv:2410.15553}.

\bibitem[{Huang et~al.(2023)Huang, Tao, Zhang, An, Jiang, Chen, Wu, and
  Feng}]{huang2023lawyer}
Quzhe Huang, Mingxu Tao, Chen Zhang, Zhenwei An, Cong Jiang, Zhibin Chen, Zirui
  Wu, and Yansong Feng. 2023.
\newblock Lawyer llama technical report.
\newblock \emph{arXiv preprint arXiv:2305.15062}.

\bibitem[{Jeong et~al.(2024)Jeong, Hwang, Yoon, Lee, and Kang}]{jeong2024olaph}
Minbyul Jeong, Hyeon Hwang, Chanwoong Yoon, Taewhoo Lee, and Jaewoo Kang. 2024.
\newblock \href {https://arxiv.org/abs/2405.12701} {Olaph: Improving factuality
  in biomedical long-form question answering}.
\newblock \emph{Preprint}, arXiv:2405.12701.

\bibitem[{Katsis et~al.(2025)Katsis, Rosenthal, Fadnis, Gunasekara, Lee, Popa,
  Shah, Zhu, Contractor, and Danilevsky}]{katsis2025mtrag}
Yannis Katsis, Sara Rosenthal, Kshitij Fadnis, Chulaka Gunasekara, Young-Suk
  Lee, Lucian Popa, Vraj Shah, Huaiyu Zhu, Danish Contractor, and Marina
  Danilevsky. 2025.
\newblock Mtrag: A multi-turn conversational benchmark for evaluating
  retrieval-augmented generation systems.
\newblock \emph{Transactions of the Association for Computational Linguistics},
  13:784--808.

\bibitem[{Kurtz and Silverman(1996)}]{kurtz1996calgary}
Suzanne~M Kurtz and Jonathan~D Silverman. 1996.
\newblock The calgary—cambridge referenced observation guides: an aid to
  defining the curriculum and organizing the teaching in communication training
  programmes.
\newblock \emph{Medical education}, 30(2):83--89.

\bibitem[{Kwan et~al.(2024)Kwan, Zeng, Jiang, Wang, Li, Shang, Jiang, Liu, and
  Wong}]{kwan2024mt}
Wai-Chung Kwan, Xingshan Zeng, Yuxin Jiang, Yufei Wang, Liangyou Li, Lifeng
  Shang, Xin Jiang, Qun Liu, and Kam-Fai Wong. 2024.
\newblock Mt-eval: A multi-turn capabilities evaluation benchmark for large
  language models.
\newblock In \emph{Proceedings of the 2024 Conference on Empirical Methods in
  Natural Language Processing}, pages 20153--20177.

\bibitem[{Laban et~al.(2025)Laban, Hayashi, Zhou, and Neville}]{laban2025llms}
Philippe Laban, Hiroaki Hayashi, Yingbo Zhou, and Jennifer Neville. 2025.
\newblock Llms get lost in multi-turn conversation.
\newblock \emph{arXiv preprint arXiv:2505.06120}.

\bibitem[{Lee et~al.(2023)Lee, Lee, Park, Hwang, Kim, Lee, and
  Lee}]{lee2023qasa}
Yoonjoo Lee, Kyungjae Lee, Sunghyun Park, Dasol Hwang, Jaehyeon Kim, Hong-in
  Lee, and Moontae Lee. 2023.
\newblock Qasa: advanced question answering on scientific articles.
\newblock In \emph{International Conference on Machine Learning}, pages
  19036--19052. PMLR.

\bibitem[{Li et~al.(2025)Li, Li, Wang, Chang, and Wu}]{li2025structflowbench}
Jinnan Li, Jinzhe Li, Yue Wang, Yi~Chang, and Yuan Wu. 2025.
\newblock Structflowbench: A structured flow benchmark for multi-turn
  instruction following.
\newblock \emph{arXiv preprint arXiv:2502.14494}.

\bibitem[{Lin(2004)}]{lin2004rouge}
Chin-Yew Lin. 2004.
\newblock Rouge: A package for automatic evaluation of summaries.
\newblock In \emph{Text summarization branches out}, pages 74--81.

\bibitem[{Liu et~al.(2024)Liu, Feng, Xue, Wang, Wu, Lu, Zhao, Deng, Zhang, Ruan
  et~al.}]{liu2024deepseek}
Aixin Liu, Bei Feng, Bing Xue, Bingxuan Wang, Bochao Wu, Chengda Lu, Chenggang
  Zhao, Chengqi Deng, Chenyu Zhang, Chong Ruan, and 1 others. 2024.
\newblock Deepseek-v3 technical report.
\newblock \emph{arXiv preprint arXiv:2412.19437}.

\bibitem[{Liu et~al.(2025)Liu, Guo, Lou, Zeng, Niu, Wang, Xu, Cai, Yang, Zhao
  et~al.}]{liu2025fin}
Zhaowei Liu, Xin Guo, Fangqi Lou, Lingfeng Zeng, Jinyi Niu, Zixuan Wang, Jiajie
  Xu, Weige Cai, Ziwei Yang, Xueqian Zhao, and 1 others. 2025.
\newblock Fin-r1: A large language model for financial reasoning through
  reinforcement learning.
\newblock \emph{arXiv preprint arXiv:2503.16252}.

\bibitem[{Madaan et~al.(2023)Madaan, Tandon, Gupta, Hallinan, Gao, Wiegreffe,
  Alon, Dziri, Prabhumoye, Yang et~al.}]{madaan2023self}
Aman Madaan, Niket Tandon, Prakhar Gupta, Skyler Hallinan, Luyu Gao, Sarah
  Wiegreffe, Uri Alon, Nouha Dziri, Shrimai Prabhumoye, Yiming Yang, and 1
  others. 2023.
\newblock Self-refine: Iterative refinement with self-feedback.
\newblock \emph{Advances in Neural Information Processing Systems},
  36:46534--46594.

\bibitem[{Manes et~al.(2024)Manes, Ronn, Cohen, Ber, Horowitz-Kugler, and
  Stanovsky}]{manes2024k}
Itay Manes, Naama Ronn, David Cohen, Ran~Ilan Ber, Zehavi Horowitz-Kugler, and
  Gabriel Stanovsky. 2024.
\newblock K-qa: A real-world medical q\&a benchmark.
\newblock In \emph{Proceedings of the 23rd Workshop on Biomedical Natural
  Language Processing}, pages 277--294.

\bibitem[{Reddy et~al.(2019)Reddy, Chen, and Manning}]{reddy2019coqa}
Siva Reddy, Danqi Chen, and Christopher~D Manning. 2019.
\newblock Coqa: A conversational question answering challenge.
\newblock \emph{Transactions of the Association for Computational Linguistics},
  7:249--266.

\bibitem[{shareAI(2023)}]{ShareGPT-Chinese-English-90k}
shareAI. 2023.
\newblock Sharegpt-chinese-english-90k bilingual human-machine qa dataset.
\newblock
  \url{https://huggingface.co/datasets/shareAI/ShareGPT-Chinese-English-90k}.

\bibitem[{Singhal et~al.(2025)Singhal, Tu, Gottweis, Sayres, Wulczyn, Amin,
  Hou, Clark, Pfohl, Cole-Lewis et~al.}]{singhal2025toward}
Karan Singhal, Tao Tu, Juraj Gottweis, Rory Sayres, Ellery Wulczyn, Mohamed
  Amin, Le~Hou, Kevin Clark, Stephen~R Pfohl, Heather Cole-Lewis, and 1 others.
  2025.
\newblock Toward expert-level medical question answering with large language
  models.
\newblock \emph{Nature Medicine}, 31(3):943--950.

\bibitem[{Sun et~al.(2024)Sun, Liu, Zhou, Huang, Song, Zhao, Zhang, Zhang, and
  Gai}]{sun2024parrot}
Yuchong Sun, Che Liu, Kun Zhou, Jinwen Huang, Ruihua Song, Wayne~Xin Zhao,
  Fuzheng Zhang, Di~Zhang, and Kun Gai. 2024.
\newblock Parrot: Enhancing multi-turn instruction following for large language
  models.
\newblock In \emph{Proceedings of the 62nd Annual Meeting of the Association
  for Computational Linguistics (Volume 1: Long Papers)}, pages 9729--9750.

\bibitem[{Team(2024)}]{team2024qwen2}
Qwen Team. 2024.
\newblock Qwen2 technical report.
\newblock \emph{arXiv preprint arXiv:2407.10671}.

\bibitem[{Touvron et~al.(2023)Touvron, Lavril, Izacard, Martinet, Lachaux,
  Lacroix, Rozi{\`e}re, Goyal, Hambro, Azhar et~al.}]{touvron2023llama}
Hugo Touvron, Thibaut Lavril, Gautier Izacard, Xavier Martinet, Marie-Anne
  Lachaux, Timoth{\'e}e Lacroix, Baptiste Rozi{\`e}re, Naman Goyal, Eric
  Hambro, Faisal Azhar, and 1 others. 2023.
\newblock Llama: Open and efficient foundation language models.
\newblock \emph{arXiv preprint arXiv:2302.13971}.

\bibitem[{Wang et~al.(2024)Wang, Chang, Qian, Chen, Chen, Jiang, Zhang,
  Nakashima, and Nagahara}]{wang2024direct}
Bowen Wang, Jiuyang Chang, Yiming Qian, Guoxin Chen, Junhao Chen, Zhouqiang
  Jiang, Jiahao Zhang, Yuta Nakashima, and Hajime Nagahara. 2024.
\newblock Direct: Diagnostic reasoning for clinical notes via large language
  models.
\newblock \emph{Advances in neural information processing systems},
  37:74999--75011.

\bibitem[{Wang et~al.(2025{\natexlab{a}})Wang, Jiang, Susumu, Miwa, Chen, and
  Nakashima}]{wang2025taming}
Bowen Wang, Zhouqiang Jiang, Yasuaki Susumu, Shotaro Miwa, Tianwei Chen, and
  Yuta Nakashima. 2025{\natexlab{a}}.
\newblock Taming the untamed: Graph-based knowledge retrieval and reasoning for
  mllms to conquer the unknown.
\newblock In \emph{2025 IEEE/CVF International Conference on Computer Vision
  (ICCV)}, pages 4732--4742. IEEE.

\bibitem[{Wang et~al.(2023)Wang, Wang, Liu, Chen, Yuan, Peng, and
  Ji}]{wang2023mint}
Xingyao Wang, Zihan Wang, Jiateng Liu, Yangyi Chen, Lifan Yuan, Hao Peng, and
  Heng Ji. 2023.
\newblock Mint: Evaluating llms in multi-turn interaction with tools and
  language feedback.
\newblock \emph{arXiv preprint arXiv:2309.10691}.

\bibitem[{Wang et~al.(2025{\natexlab{b}})Wang, Shen, Huang, Niu, and
  Ou}]{wang2025clegal}
Yizhen Wang, Xueying Shen, Zixian Huang, Lihui Niu, and Shiyan Ou.
  2025{\natexlab{b}}.
\newblock clegal-qa: a chinese legal question answering with natural language
  generation methods.
\newblock \emph{Complex \& Intelligent Systems}, 11(1):77.

\bibitem[{Wu et~al.(2023)Wu, Irsoy, Lu, Dabravolski, Dredze, Gehrmann,
  Kambadur, Rosenberg, and Mann}]{wu2023bloomberggpt}
Shijie Wu, Ozan Irsoy, Steven Lu, Vadim Dabravolski, Mark Dredze, Sebastian
  Gehrmann, Prabhanjan Kambadur, David Rosenberg, and Gideon Mann. 2023.
\newblock Bloomberggpt: A large language model for finance.
\newblock \emph{arXiv preprint arXiv:2303.17564}.

\bibitem[{Wu et~al.(2024)Wu, Zhang, Gao, Wang, Xu, Hu, and
  Chen}]{wu2024benchmarking}
Shiwei Wu, Chen Zhang, Yan Gao, Qimeng Wang, Tong Xu, Yao Hu, and Enhong Chen.
  2024.
\newblock Benchmarking large language models for conversational question
  answering in multi-instructional documents.
\newblock \emph{arXiv preprint arXiv:2410.00526}.

\bibitem[{Xu et~al.(2023)Xu, Song, Iyyer, and Choi}]{xu2023critical}
Fangyuan Xu, Yixiao Song, Mohit Iyyer, and Eunsol Choi. 2023.
\newblock A critical evaluation of evaluations for long-form question
  answering.
\newblock \emph{arXiv preprint arXiv:2305.18201}.

\bibitem[{Zhang et~al.(2025{\natexlab{a}})Zhang, Li, Long, Zhang, Lin, Yang,
  Xie, Yang, Liu, Lin, Huang, and Zhou}]{qwen3embedding}
Yanzhao Zhang, Mingxin Li, Dingkun Long, Xin Zhang, Huan Lin, Baosong Yang,
  Pengjun Xie, An~Yang, Dayiheng Liu, Junyang Lin, Fei Huang, and Jingren Zhou.
  2025{\natexlab{a}}.
\newblock Qwen3 embedding: Advancing text embedding and reranking through
  foundation models.
\newblock \emph{arXiv preprint arXiv:2506.05176}.

\bibitem[{Zhang et~al.(2025{\natexlab{b}})Zhang, Wang, Li, Ren, Zhu, and
  Naseem}]{zhang2025turnbench}
Yiran Zhang, Mo~Wang, Xiaoyang Li, Kaixuan Ren, Chencheng Zhu, and Usman
  Naseem. 2025{\natexlab{b}}.
\newblock Turnbench-ms: A benchmark for evaluating multi-turn, multi-step
  reasoning in large language models.
\newblock \emph{arXiv preprint arXiv:2506.01341}.

\bibitem[{Zheng et~al.(2023)Zheng, Chiang, Sheng, Zhuang, Wu, Zhuang, Lin, Li,
  Li, Xing et~al.}]{zheng2023judging}
Lianmin Zheng, Wei-Lin Chiang, Ying Sheng, Siyuan Zhuang, Zhanghao Wu, Yonghao
  Zhuang, Zi~Lin, Zhuohan Li, Dacheng Li, Eric Xing, and 1 others. 2023.
\newblock Judging llm-as-a-judge with mt-bench and chatbot arena.
\newblock \emph{Advances in neural information processing systems},
  36:46595--46623.

\bibitem[{Zhou and Shen(2024)}]{zhou2024processing}
Yanmengqian Zhou and Lijiang Shen. 2024.
\newblock Processing of misinformation as motivational and cognitive biases.
\newblock \emph{Frontiers in Psychology}, 15:1430953.

\end{thebibliography}

\newpage
\appendix
\section*{Appendix}

\begin{table*}[hpbt]
\centering
\small
\caption{Conversational QA and LFQA benchmark. Columns show average ground-truth length, average number of turns, and whether dialogs are multi-turn, answers are long-form, and the task is open-domain (\cmark=yes, \xmark=no). Notation: t = tokens, w = words, c = Chinese characters, -=not provided in the original paper.}
\label{tab:bc}
\begin{tabular}{lccccc}
\toprule
\textbf{Benchmark} & \textbf{\# Avg. Length} & \textbf{\# Avg. Turns} & \textbf{Multi-turn} & \textbf{Long-Form} & \textbf{Open-Domain}\\
\midrule
QuAC~\citep{choi2018quac} & 14.6 t & 7.2 & \cmark & \xmark & \xmark \\
CoQA~\citep{reddy2019coqa}  & 2.52 t & 15.97 & \cmark & \xmark & \xmark \\
DoQA~\citep{campos2020doqa} & 12.99 t & 4.48 & \cmark & \xmark & \xmark \\
Doc2Dial~\citep{feng2020doc2dial} & 21 t & 12 & \cmark & \xmark & \xmark \\
MultiDoc2Dial~\citep{feng2021multidoc2dial} & 21.6 t & 6.36 & \cmark & \xmark & \xmark \\
ConvFinQA~\citep{chen2022convfinqa} & - & 3.67 & \cmark & \xmark & \xmark \\
TopiOCQA~\citep{adlakha2022topiocqa} & 11.75 w & 12 & \cmark & \xmark & \cmark \\
InsCoQA~\citep{wu2024benchmarking} & - & 3.11 & \cmark & \xmark & \xmark \\
QASA~\citep{lee2023qasa}            &  50 t    & 1     & \xmark & \cmark & \xmark \\
FinTextQA~\citep{chen2024fintextqa} & 75 w & 1 & \xmark & \cmark & \cmark \\
K\mbox{-}QA~\citep{manes2024k} & 119.89 t & 1 & \xmark & \cmark & \cmark \\
MedLFQA~\citep{jeong2024olaph} & 132.86 t & 1 & \xmark & \cmark & \cmark \\
cLegal\mbox{-}QA~\citep{wang2025clegal} & 93 c & 1 & \xmark & \cmark & \cmark \\
\midrule
\textbf{KnowMT\mbox{-}Bench (ours)} & 75.19 w / 95.15 t & 2.98 & \cmark & \cmark & \cmark \\
\bottomrule
\end{tabular}%
\end{table*}

\section{Related Work}

\subsection{Multi-turn Dialogues Benchmarks for LLMs}

Evaluating LLMs in multi-turn dialogues is a critical and active area of research. Early benchmarks such as MT-Bench~\citep{zheng2023judging} and MT-Bench++\citep{sun2024parrot} assess general conversational quality using an LLM-as-judge approach. Subsequent works have introduced more diverse evaluation paradigms. For instance, BotChat~\citet{duan2024botchat} evaluates alignment with human conversational patterns, while others, including MT-Eval~\citep{kwan2024mt} and MT-Bench-101~\citep{bai2024mt}, propose multi-dimensional frameworks to assess specific capabilities like instruction adherence and context utilization.

Beyond general benchmarks, a range of specialized benchmarks have been proposed to probe distinct abilities within multi-turn dialogue. TurnBench-MS~\citep{zhang2025turnbench} and WIL~\citep{banatt2024wilt} are designed to assess iterative multi-step reasoning, while Multi-IF~\citep{he2024multi} and StructFlowBench~\citep{ li2025structflowbench} focus on the instruction-following ability of LLMs. MINT~\citep{wang2023mint} explicitly evaluates LLMs’ ability to incorporate external tools and language feedback during multi-turn interactions. Additionally, some benchmarks evaluate critical risks in multi-turn dialogues, with FairMT-Bench~\citep{fan2024fairmt} measuring fairness and bias propagation; similarly, mtRAGh~\citep{katsis2025mtrag} evaluates multi-turn retrieval-augmented generation systems, emphasizing retrieval quality and conversational grounding. While comprehensive, these benchmarks do not specifically focus on the factual capability of LLMs within multi-turn, knowledge-grounded dialogues. Our work addresses this critical gap by systematically assessing factual capability in such contexts.

\subsection{Long-form Question Answering Benchmarks for LLMs}

LFQA is a knowledge-base open-domain question answering task, particularly in specialized domains like medicine, finance, and law. Evaluation paradigms for LFQA have evolved over time. Some benchmarks in LFQA, such as FinTextQA~\citep{chen2024fintextqa} and cLegal-QA~\citep{wang2025clegal} rely on surface-level similarity metrics like ROUGE~\citep{lin2004rouge}, which may not correlate well with human evaluation~\citep{xu2023critical}. To improve reliability, subsequent works such as LEXam~\citep{fan2025lexam} adopted an LLM-as-judge paradigm.

A more fine-grained and interpretable approach was introduced by K-QA and MedLFQA~\citep{manes2024k, jeong2024olaph}, which first decomposes reference answers into atomic facts and then uses NLI-based methods to check for entailment and contradiction at the fact level. This paradigm offers enhanced interpretability by assessing factuality on explicit, human-understandable statements. Our benchmark builds on this NLI-based paradigm but makes some key advancements. We introduce a new dimension for \textbf{information delivery efficiency} to evaluate content effectiveness, and we extend the LFQA task to a multi-turn dialogue setting for the first time to make it more closely resemble professional consultation scenarios.

\section{Detailed Comparison to More QA Benchmark }

Table~\ref{tab:bc} summarizes widely used conversational QA and long-form QA benchmarks in terms of average answer length, number of turns, and task characteristics. Most conversational QA datasets emphasize multi-turn interaction but contain relatively short answers, while long-form QA benchmarks are typically single-turn and open-domain with longer responses. Our KnowMT-Bench bridges these two directions by combining multi-turn dialogue with long-form, open-domain answers, reflecting more realistic information-seeking scenarios.

\section{Annotation Details}
\label{app:ad}

\begin{figure*}[t]
    \centering
    \includegraphics[width=0.95\linewidth]{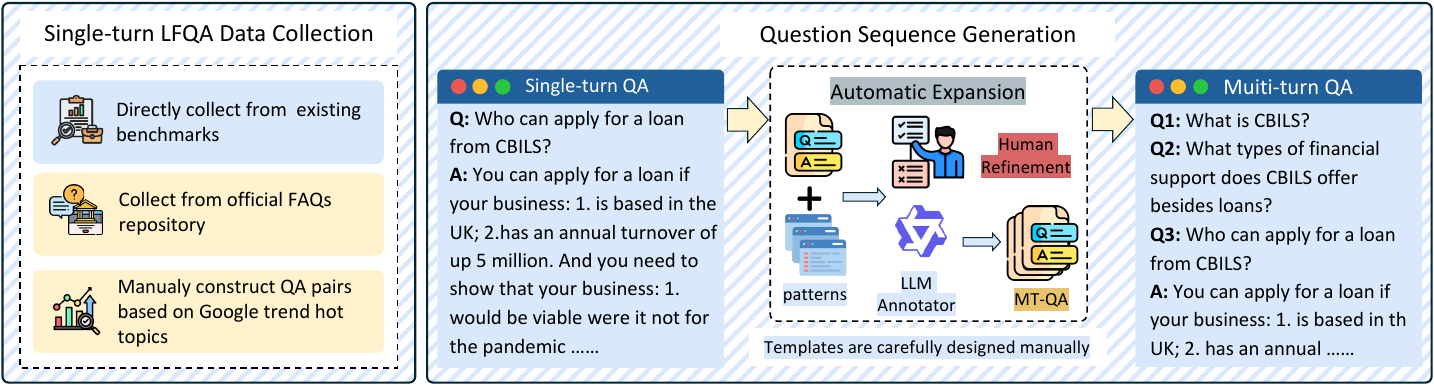}
    \caption{Overview of the dataset construction process}%
    \label{fig:app_dataset_construction}
\end{figure*}

We give a detailed introduction to the annotation details of KnowMT-Bench here; the overview of the process is illustrated in Figure~\ref{fig:app_dataset_construction}, including (i) single-turn data collection and (ii) question sequence generation
\subsection{Additional Details of Single-Turn LFQA Data Collection}
\label{app:annotation_detail_single_turn}

For the medical domain, we included all 201 QA pairs from the K-QA benchmark~\citep{manes2024k}. Since the original answers often contained both \textit{must-to-have} supporting facts and \textit{nice-to-have} details, we manually removed the redundant segments beyond the must-have facts while ensuring that the resulting ground-truth answers remained coherent and fluent. The supporting evidence was derived from authoritative sources such as institutional websites used during K-QA annotation.

\begin{tcblisting}{
  enhanced,breakable,
  colback=white,colframe=Gray!50!black,
  colbacktitle=Gray!60!black,coltitle=white,
  title=Example of Multi-Turn Questions With Answer and
   Reference,
  listing engine=listings,
  listing only,
  listing options={
    basicstyle=\ttfamily\scriptsize,
    breaklines=true, columns=fullflexible, keepspaces=true,
    showstringspaces=false
  }
}
{
      "sample_id": "9461cce2",
      "domain": ["finance"],
      "single-turn question": "Compared to living in other parts of the UK, what two additional documents are required when taxing a vehicle in Northern Ireland?",
      "multi-turn questions": [
        "How has vehicle taxation evolved in the United Kingdom over the past century?",
        "What are the current vehicle taxation policies in place in the UK today?",
        "Compared to living in other parts of the UK, what two additional documents are required when taxing a vehicle in Northern Ireland?"
      ],
      "answer": "When taxing a vehicle at a Post Office in Northern Ireland, you need to provide two additional documents: a paper copy of your insurance certificate or cover note, and an original MOT test certificate or evidence of a Temporary Exemption Certificate (TEC).",
      "must_have": [
        "When taxing a vehicle at a Post Office in Northern Ireland, a paper copy of your insurance certificate is required",
        "When taxing a vehicle at a Post Office in Northern Ireland, a cover note is required as an alternative to an insurance certificate",
        "When taxing a vehicle at a Post Office in Northern Ireland, an original MOT test certificate is required",
        "When taxing a vehicle at a Post Office in Northern Ireland, evidence of a Temporary Exemption Certificate (TEC) is required as an alternative to an MOT test certificate"
      ],
      "source": "HOT-FINANCE-TOPIC",
      "url": [
        "https://www.gov.uk/vehicle-tax"
      ]
}
\end{tcblisting}

For the legal domain, we collect 116 QA pairs from the official FAQ repository maintained by the U.S. Securities and Exchange Commission (SEC)\footnote{\url{https://www.sec.gov/answers/faqs.htm}} and 184 QA pairs from the policy-focused subset of FinTextQA~\citep{chen2024fintextqa}. Several quality-control steps were applied, including the removal of trivial answers consisting only of affirmation, negation, or phrase-level responses, as well as the manual addition of missing jurisdictional context (e.g., ``Hong Kong'' in entries from HKMA) to resolve ambiguities. The resulting curated subset thus covers major financial jurisdictions, including Hong Kong, the European Union, and the United States.

For the financial domain, to align the benchmark with topics of broad public interest, we further collect ``trending'' and ``rising'' terms from Google Trends\footnote{\url{https://trends.google.com/trends/}}, focusing on the ``Finance'' category and its subcategories in the United States and the United Kingdom over the five years preceding February 26, 2025. After de-duplication and filtering, this process yielded 272 unique finance-related topics. A word cloud visualization of these topics is shown in Fig.~\ref{fig:app_google_fin_topic_wordcloud}. We manually classified these topics into three categories: institutions and products, policies and events, and concepts (for more detail, see Appendix~\ref{sec:organizations_products}), and then constructed QA pairs through annotation. Eighteen topics were randomly selected to receive two QA pairs, yielding a total of 300 QA pairs.  

We recruited six annotators, all graduate students in either Finance or Computer Science, and instructed them to formulate complete, non-redundant questions and answers for each topic by referring to authoritative sources such as official institutional websites, Encyclop{\ae}dia Britannica, and Investopedia, and to record the source URL as supporting evidence. A rigorous verification process was then conducted to ensure accuracy, clarity, and consistency across all QA pairs, including spot-checking answers against their cited references. Overall, we obtained 201 QA pairs in the medical domain, 300 in the legal domain, and 300 in the financial domain (801 in total).

\subsection{Additional Details for Multi-Turn Dialogue Generation}
\label{app:annotation_detail_expansion_templates}

\begin{figure*}[t]
    \centering
    \begin{subfigure}[t]{0.68\textwidth}
        \centering
        \includegraphics[width=\linewidth]{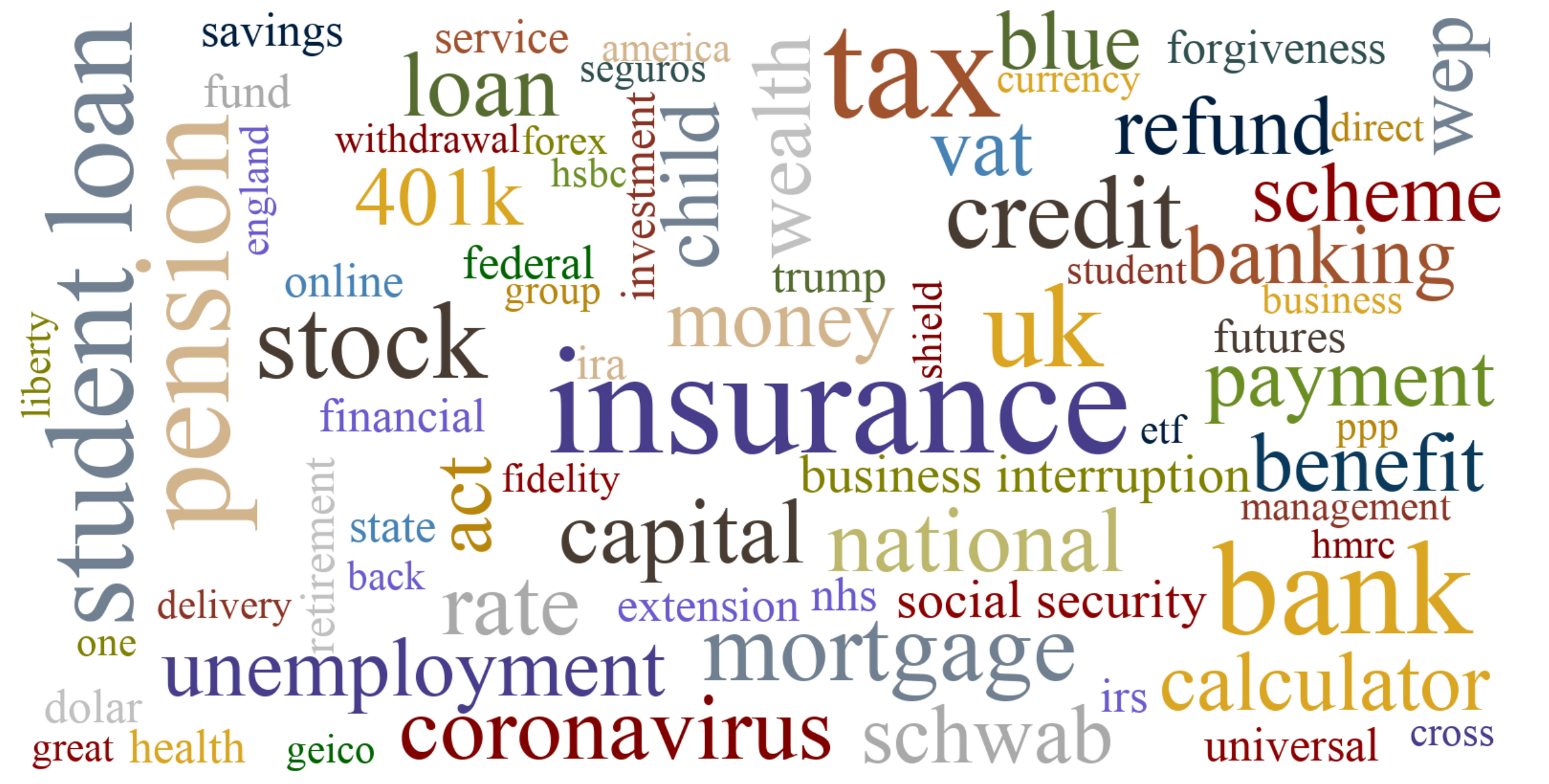}
        \caption{Financial Topics on Google Trends.}
        \label{fig:app_google_fin_topic_wordcloud}
    \end{subfigure}\hfill
    \begin{subfigure}[t]{0.28\textwidth}
        \centering
        \includegraphics[width=\linewidth]{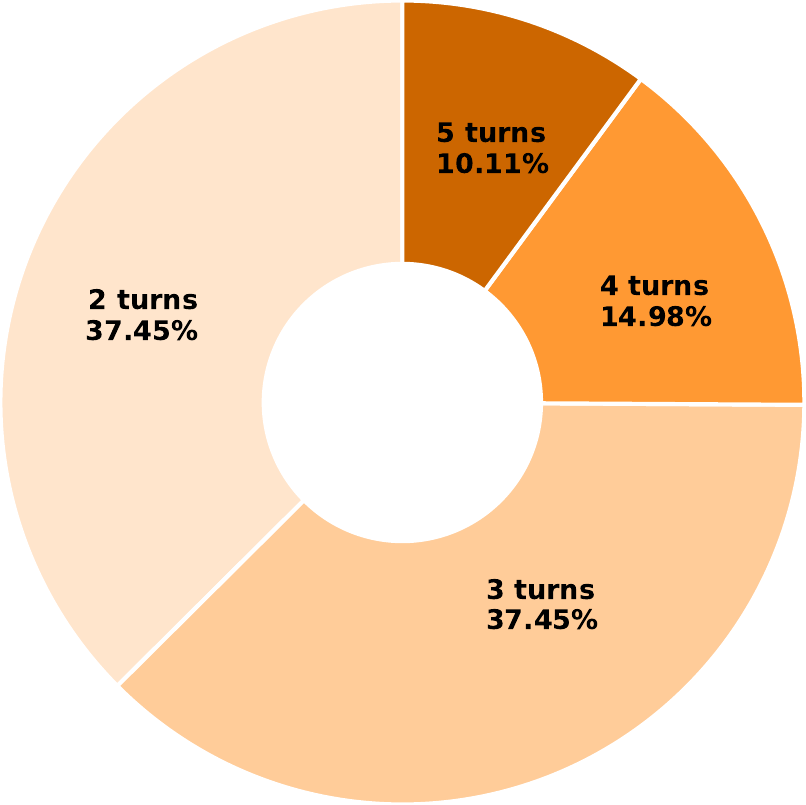}
        \caption{Distribution of turn count}
        \label{fig:app_turn_distribution}
    \end{subfigure}

    \caption{Financial topics and turn-count distribution.}
    \label{fig:app_google_fin_and_turn}
\end{figure*}

To reduce annotation effort, we designed a set of guiding expansion patterns to bootstrap multi-turn question generation, including $\mathcal{T}_2$ for two-turn dialogues ($|\mathcal{T}_2|=8$) and $\mathcal{T}_3$ for three-turn dialogues ($|\mathcal{T}_3|=10$) (see Appendix~\ref{app:questions template}, where all patterns are shown in the prompt). Using these patterns, Qwen2.5-32B-Instruct was prompted to automatically expand a single-turn seed question into two- and three-turn question sequences (prompts are provided in Appendix~\ref{app:build_question_list_prompt}). All generated question sequences were manually reviewed and revised to avoid cases where different questions essentially ask the same underlying point, while maintaining diversity and broad factual coverage. In addition, we ensured that each dialogue could naturally and logically lead to the final question in a realistic conversational flow. We then manually extended a subset of the generated sequences into four- and five-turn dialogues to ensure a natural progression and high overall quality. Figure~\ref{fig:app_turn_distribution} reports the realized turn-length distribution.

\subsection{Additional Details for Automatic Evaluation}
\label{app:annotation_detail_evaluation}

\begin{table*}[t]
\centering
\small
\caption{Summary of the main evaluation metrics.}
\label{tab:metric_summary}
\setlength{\tabcolsep}{4pt}
\begin{tabular}{lp{0.19\textwidth}p{0.48\textwidth}p{0.18\textwidth}}
\toprule
\textbf{Metric} & \textbf{Dimension} & \textbf{Meaning} & \textbf{Direction} \\
\midrule
$\mathbf{S_f}$ & Factuality & Factual F1 over supported facts and generated claims & Higher is better \\
$\mathbf{S_h}$ & Reliability & Hallucination F1 over contradicted facts and claims & Lower is better \\
$\mathbf{D_f}$ & Efficiency & Token cost per correctly entailed fact & Lower is better \\
$\mathbf{D_h}$ & Efficiency & Token cost per contradicted fact & Higher means sparser errors \\
\bottomrule
\end{tabular}
\end{table*}

\subsubsection{Validation of The Evaluation Pipeline}
\begin{figure*}[t]
    \centering
    \includegraphics[width=0.95\linewidth]{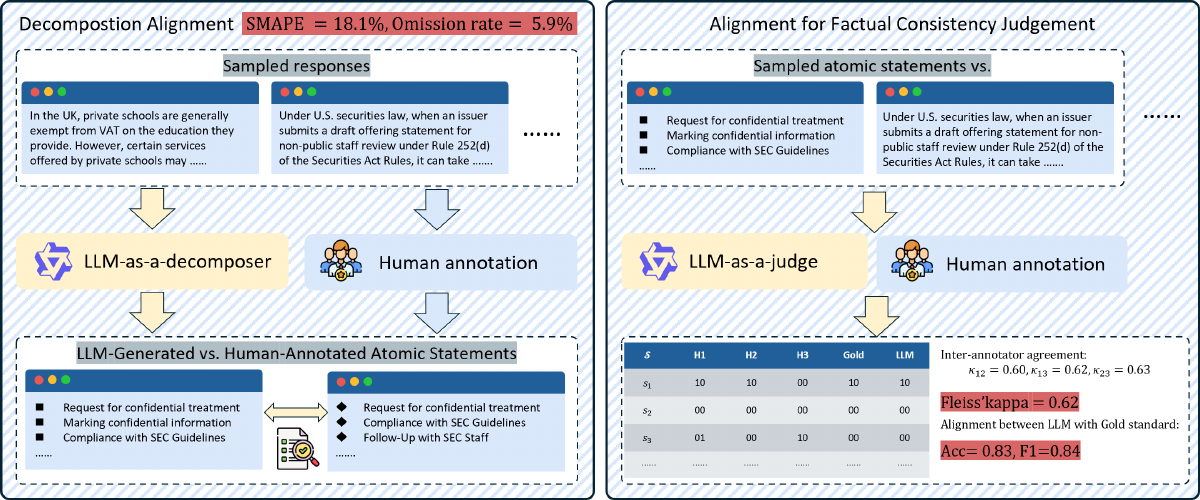}
    \caption{Human validation of our evaluation pipeline. (\textbf{left}) Alignment between LLM-based and human-annotated atomic decompositions. (\textbf{right}) Alignment between LLM-based and human annotations for factual consistency judgments.}
    \label{fig:alignment}
\end{figure*}

To validate our two-stage evaluation pipeline, we conducted a human annotation study, illustrated in Fig.~\ref{fig:alignment}. For the Atomic Decomposition stage, we sampled 100 model answers drawn from four representative LLMs of varying scales (Qwen2.5-14B-Instruct, Llama-3.1-8B-Instruct, DeepSeek-V3-0324, and GPT-4o-2024-08-06), covering single-turn or multi-turn settings. Each answer was decomposed into atomic statements by our decomposer model (Qwen2.5-32B-Instruct) and compared against human annotations. As shown in the top part of Fig.~\ref{fig:alignment}, the decomposition achieved high fidelity, with a Symmetric Mean Absolute Percentage Error (SMAPE) of \textbf{18.1\%} in statement counts and an omission rate of only \textbf{5.9\%}, indicating that discrepancies were mainly due to under-segmentation rather than semantic distortion.

\begin{table}[t]
\captionsetup{width=0.9\linewidth}
\small
\centering
\caption{Performance comparison on the Factual Consistency Judgment stage.}
\label{tab:consistency_comparison}
\vspace{-5pt}
\setlength{\tabcolsep}{3.5pt} %
\begin{tabular}{lcccc}
\toprule
\textbf{Model} & \textbf{Accuracy} & \textbf{F1} & \textbf{Precision} & \textbf{Recall} \\
\midrule
Qwen-2.5-14B & \textbf{0.83} & \textbf{0.84} & 0.84 & \textbf{0.83} \\
GPT-4o & 0.81 & 0.82 & \textbf{0.84} & 0.81 \\
DeepSeek-V3 & 0.76 & 0.78 & 0.84 & 0.76 \\
\bottomrule
\end{tabular}
\end{table}

For the Factual Consistency Judgment stage, the same 100 dialogues were decomposed into 1,687 evaluation items, consisting of atomic statements paired with the opposing full-text answers. These were labeled by our evaluator (Qwen2.5-14B-Instruct) and independently annotated by three human experts. As shown in the bottom part of Fig.~\ref{fig:alignment}, human annotators achieved substantial agreement (pairwise Cohen’s $\kappa$ values of 0.60, 0.62, and 0.63; Fleiss’ $\kappa = 0.62$). Furthermore, we constructed a gold standard by majority voting over the three human annotations. The agreement between this gold standard and individual annotators was consistently high, with Cohen’s $\kappa$ values of 0.80, 0.79, and 0.81, indicating strong alignment between the aggregated ground truth and expert annotations.

\begin{table}[h]
\centering
\small
\caption{Performance breakdown per model, as judged by the Qwen-2.5-14B evaluator. OR refers to Omission Rate.}
\label{tab:bias_check_performance}
\setlength{\tabcolsep}{2.8pt}
\begin{tabular}{lcccccc}
\toprule
\textbf{Model} & \textbf{Acc.} & \textbf{F1} & \textbf{Prec.} & \textbf{Rec.} & \textbf{SMAPE(\%)} & \textbf{OR(\%)} \\
\midrule
DS-V3 & 0.85 & 0.85 & 0.87 & 0.85 & 13.12 & 6.34\\
L3-8B   & 0.84 & 0.85 & 0.86 & 0.84 & 20.66 & 2.54\\
GPT-4o      & 0.83 & 0.83 & 0.83 & 0.83 & 19.70 & 7.52\\
Q2.5-14B & 0.81 & 0.81 & 0.81 & 0.81 & 18.10 & 5.90\\
\bottomrule
\end{tabular}
\end{table}

We then assessed the performance of various models against this gold standard, as detailed in Table~\ref{tab:consistency_comparison}. The results show that Qwen-2.5-14B achieves the most favorable performance among the candidates, leading in both accuracy and F1-score. Consequently, considering its strong performance and computational efficiency, we selected Qwen-2.5-14B as the designated evaluator for our pipeline. We conjecture that this outcome may be attributed to the nature of the task; while larger models possess more powerful general reasoning abilities, they might be prone to overly complex inference paths for a constrained judgment task, potentially introducing instability. A well-calibrated, medium-sized model, such as Qwen-2.5-14B, may follow a more direct and consistent reasoning process, rendering it more reliable for this specific application.

\begin{table}[h]
\centering
\small
\caption{Pairwise significance tests on model accuracy for evaluator impartiality. We report p-values from Mann-Whitney U tests. \textit{Note:} ns denotes p $\geq$ 0.01.}
\label{tab:mwu_significance}
\setlength{\tabcolsep}{5pt}
\begin{tabular}{lcccc}
\toprule
\textbf{Model} & \textbf{DS-V3} & \textbf{L3-8B} & \textbf{GPT-4o} & \textbf{Q2.5-14B} \\
\midrule
DS-V3    & -       & .761 ns   & .445 ns   & .183 ns   \\
L3-8B    & .761 ns & -         & .674 ns   & .347 ns   \\
GPT-4o   & .445 ns & .674 ns   & -         & .602 ns   \\
Q2.5-14B & .183 ns & .347 ns   & .602 ns   & -         \\
\bottomrule
\end{tabular}
\end{table}

\subsubsection{Investigating Bias of Decomposer and Evaluator}

A critical consideration for our pipeline is whether the chosen evaluator, Qwen-2.5-14B, exhibits any bias, particularly a self-preference for models within its own family. To investigate this, we conducted a detailed analysis of the evaluation outcomes for each target model, with performance metrics presented in Table~\ref{tab:bias_check_performance}. The metrics are tightly clustered across all models, with F1-scores ranging from 0.81 to 0.85. Notably, the model from the evaluator's own family, Qwen2.5-14B, does not receive a disproportionately high score.

To further validate this observation with a method robust to non-normal data distributions, we performed pairwise non-parametric Mann-Whitney U tests on the models' accuracy scores. As the answers for each question were generated by a single, randomly assigned model, the groups of scores for each model are independent, making this test appropriate. The results, summarized in Tab.~\ref{tab:mwu_significance}, show that all comparisons yield p-values well above the 0.01 threshold. Furthermore, the calculated effect sizes for all pairs were negligible (\(r < 0.03\)), indicating that the observed differences lack practical importance. In addition, a Kruskal-Wallis test yields \(H = 1.9552, p = 5.82\times10^{-1}\), indicating no significant overall difference among models. These statistical evidences strongly support the conclusion that our evaluation pipeline operates impartially and does not systematically favor any specific model architecture or family.

\begin{table}[t]
    \centering
    \caption{Pairwise significance tests (p-values from Mann-Whitney U tests) on model SMAPE for Decomposer. \textit{Note:} ns denotes p $\geq$ 0.01.}
    \setlength{\tabcolsep}{2.2pt} %
    \label{tab:mwu_significance2}
    \begin{tabular}{lcccc}
    \toprule
    \textbf{Model} & \textbf{DS-V3} & \textbf{L3-8B} & \textbf{GPT-4o} & \textbf{Q2.5-14B} \\
    \midrule
    DS-V3    & -       & .477 ns   & .237 ns   & .104 ns   \\
    L3-8B    & .477 ns & -         & .086 ns   & .030 ns \\
    GPT-4o   & .237 ns & .086 ns   & -         & .841 ns   \\
    Q2.5-14B & .104 ns & .030 ns     & .841 ns   & -         \\
    \bottomrule
    \end{tabular}
\end{table}

Analogous to the evaluator check, we test whether the Qwen-based Decomposer systematically favors or penalizes some models when extracting atomic facts from generated answers. For each model, we compute decomposition count error against human gold using SMAPE, then perform (1) a Kruskal–Wallis test across models and (2) pairwise Mann-Whitney U tests (see the results in Table~\ref{tab:mwu_significance2}). The Kruskal-Wallis test yields \(H = 6.1147, p = 1.062\times10^{-1}\), showing no significant global difference across model families. Pairwise tests again indicate that no pair is significant at the stricter 0.01 level, and only one pair reaches 0.05 with a small effect size. This suggests that any between-model variation in decomposition error is minor, and we find no strong evidence that the Decomposer systematically favors or penalizes specific model families.

\subsubsection{Investigating Robustness of Decomposer and Evaluator}

To test the robustness of our evaluation pipeline. We conduct cross-model validation by replacing the original LLM with GPT-4o at two different stages of the pipeline, the decomposer and the evaluator, and then re-evaluate four representative LLMs (GPT-4o, DeepSeek-V3, Llama-3.3-70B, and Qwen-2.5-72B) under both single-turn and multi-turn settings. The results are shown in Table~\ref{tab:cross_model_summary} and Table~\ref{tab:cross_model_detailed}.

\paragraph{Stability of the Decomposer} 

We first test the stability of the Atomic Fact Decomposition stage by replacing the original Qwen-2.5-32B Decomposer with GPT-4o, while keeping the Evaluator fixed. We find near-perfect agreement: (i) Pearson correlation \(r = 0.996\). This extremely high value indicates that the atomic facts extracted by the Qwen-based decomposer are highly aligned with those extracted by GPT-4o, and that the two decomposers induce nearly identical variation in the final scores. (ii) Spearman rank correlation \(\rho = 1.000\), which confirms that changing the Decomposer does not significantly alter the relative comparison between models.

\paragraph{Robustness of the Evaluator}

We further replace the original Qwen-2.5-14B Evaluator with GPT-4o, while keeping the Decomposer fixed, and compare the resulting factual F1 scores (\(\mathbf{S_f}\)). We observe strong statistical consistency between the two Evaluators:
(i) Pearson correlation \(r = 0.982\), which indicates a strong linear relationship between the two sets of scores. (ii) Spearman rank correlation \(\rho = 0.952\), which shows that the \textbf{relative ranking} of models is almost identical under the two Evaluators. Specifically, as shown in Table~\ref{tab:cross_model_summary}, both Evaluators (Qwen-based and GPT-4o-based) consistently agree that GPT-4o is the top-tier model, and they preserve the overall separation between the proprietary model and the open-source models in both single-turn and multi-turn evaluation. The rankings of the open-source models only exhibit minor permutations, without changing our qualitative conclusions.

\begin{table*}
\centering
\small
\setlength{\tabcolsep}{6pt}
\renewcommand{\arraystretch}{1.15}
\caption{Model Ranking (\(\mathbf{S_f}\)) Consistency across Evaluators and Decomposer Variants. OR refers to Original Rank.}
\label{tab:cross_model_summary}
\resizebox{\textwidth}{!}{
\begin{tabular}{clccc}
\toprule
\textbf{OR} & \textbf{Model (Mode)} & $\mathbf{S_f}$ & \textbf{GPT-4o Eval (Rank)} & \textbf{GPT-4o Decomp (Rank)} \\
\midrule
1 & GPT-4o (Single) & 28.99\% & 38.06\% (1) & 29.54\% (1) \\
2 & GPT-4o (Multi) & 28.08\% & 37.86\% (2) & 29.24\% (2) \\
3 & DeepSeek-V3 (Single) & 24.33\% & 33.82\% (3) & 24.51\% (3) \\
4 & Qwen-2.5-72B (Single) & 24.30\% & 32.37\% (5) & 24.47\% (4) \\
5 & Llama-3.3-70B (Single) & 24.06\% & 32.95\% (4) & 24.41\% (5) \\
6 & Llama-3.3-70B (Multi) & 21.37\% & 29.26\% (6) & 21.50\% (6) \\
7 & Qwen-2.5-72B (Multi) & 20.31\% & 28.54\% (8) & 20.97\% (7) \\
8 & DeepSeek-V3 (Multi) & 19.49\% & 29.19\% (7) & 19.54\% (8) \\
\bottomrule
\end{tabular}
}
\end{table*}

In summary, our main conclusion, that models’ factual ability degrades in multi-turn dialogue, remains stable under all validation configurations. As shown in Table~\ref{tab:cross_model_detailed}, regardless of whether we replace the Evaluator or the Decomposer, all models (including GPT-4o) exhibit a consistent drop in factual F1 (\(\mathbf{S_f}\))
when moving from the single-turn to the multi-turn setting. Furthermore, the evaluation framework of KnowMT-Bench is not tied to a specific architecture. While we chose the Qwen family primarily for reproducibility and computational efficiency, our cross-model validation demonstrates that using a stronger proprietary model (GPT-4o) in place of Qwen leads to the same qualitative conclusions.

\begin{table*}[t]
\centering
\setlength{\tabcolsep}{12pt}
\renewcommand{\arraystretch}{1}
\caption{Cross-Model Validation – Detailed Metrics for Evaluator and Decomposer Variants.}
\vspace{-5pt}
\label{tab:cross_model_detailed}
\begin{tabular}{lllrrrr}
\toprule
\textbf{Model} & \textbf{Setting} & \textbf{Configuration} & $\mathbf{S_f}$ (\%) & $\mathbf{S_h}$ (\%) & $\mathbf{D_f}$ & $\mathbf{D_h}$ \\
\midrule
GPT-4o          & Single & Original            & 28.99 & 2.29 & 226 & 505 \\
GPT-4o         & Single & w/ GPT-4o Eval      & 38.06 & 3.36 & 179 & 434 \\
GPT-4o         & Single & w/ GPT-4o Decomp    & 29.54 & 2.55 & 226 & 507 \\
\midrule
GPT-4o          & Multi  & Original            & 28.08 & 2.47 & 255 & 553 \\
GPT-4o         & Multi  & w/ GPT-4o Eval      & 37.86 & 3.11 & 214 & 532 \\
GPT-4o         & Multi  & w/ GPT-4o Decomp    & 29.24 & 2.26 & 253 & 495 \\
\midrule
DeepSeek-V3     & Single & Original            & 24.33 & 3.24 & 339 & 738 \\
DeepSeek-V3    & Single & w/ GPT-4o Eval      & 33.82 & 4.19 & 285 & 722 \\
DeepSeek-V3    & Single & w/ GPT-4o Decomp    & 24.51 & 3.32 & 340 & 739 \\
\midrule
DeepSeek-V3     & Multi  & Original            & 19.49 & 3.45 & 444 & 877 \\
DeepSeek-V3    & Multi  & w/ GPT-4o Eval      & 29.19 & 4.32 & 373 & 837 \\
DeepSeek-V3    & Multi  & w/ GPT-4o Decomp    & 19.54 & 3.40 & 445 & 877 \\
\midrule
Llama-3.3-70B   & Single & Original            & 24.06 & 3.48 & 324 & 656 \\
Llama-3.3-70B  & Single & w/ GPT-4o Eval      & 32.95 & 5.12 & 272 & 628 \\
Llama-3.3-70B  & Single & w/ GPT-4o Decomp    & 24.41 & 3.58 & 326 & 653 \\
\midrule
Llama-3.3-70B   & Multi  & Original            & 21.37 & 4.26 & 438 & 757 \\
Llama-3.3-70B  & Multi  & w/ GPT-4o Eval      & 29.26 & 6.00 & 383 & 732 \\
Llama-3.3-70B  & Multi  & w/ GPT-4o Decomp    & 21.50 & 4.16 & 435 & 775 \\
\midrule
Qwen-2.5-72B    & Single & Original            & 24.30 & 2.41 & 356 & 822 \\
Qwen-2.5-72B   & Single & w/ GPT-4o Eval      & 32.37 & 3.92 & 334 & 791 \\
Qwen-2.5-72B   & Single & w/ GPT-4o Decomp    & 24.47 & 2.49 & 366 & 822 \\
\midrule
Qwen-2.5-72B    & Multi  & Original            & 20.31 & 2.81 & 487 & 902 \\
Qwen-2.5-72B   & Multi  & w/ GPT-4o Eval      & 28.54 & 4.05 & 434 & 872 \\
Qwen-2.5-72B   & Multi  & w/ GPT-4o Decomp    & 20.97 & 2.86 & 484 & 900 \\
\bottomrule
\end{tabular}
\end{table*}

\vspace{-10pt}

\section{Quality Assurance and Annotation Workload}
\label{sec:qa_workload}
We designed a workflow to ensure data usability and reliability across single-turn and multi-turn collections. The workflow combined automatic generation with staged human verification: the model produced initial outputs, annotators reviewed and rewrote these outputs to ensure realistic conversational flow and to remove redundant follow-ups, and a subset of items was manually extended where longer dialogues were needed. At each stage we recorded human effort and time cost. All annotation and verification was performed by individuals with backgrounds in finance or computer science and included independent cross-checks to ensure consistency.

\subsection{Single-turn LFQA Annotation (300 Finance Topics)}
\label{sec:qa_singleturn_workload}
We collected 300 new daily financial QA pairs (1–2 per topic) informed by Google Trends topics. Annotators retrieved authoritative sources (for example, official websites or expert-verified encyclopedic sites) and produced full single-turn LFQA answers. Each QA pair required on average 10–20 minutes of work; a few difficult items took more than 30 minutes. Six graduate annotators participated; the total effort was about \textbf{80 person-hours}. All 300 pairs were manually checked against authoritative sources in a final review pass.

For items adapted from external collections (e.g., SEC FAQ or FinTextQA), annotators performed alignment edits such as removing trivial yes/no items and adding jurisdictional context when necessary. These edits ensured that single-turn items matched the dataset's task definition and remained non-redundant and answer-complete.

\subsection{Multi-turn Dialogue Generation and Review}
\label{sec:qa_multiturn_workload}
Multi-turn sequences were produced by a two-step process. First, an automatic generator produced initial 2–3-turn expansions for each single-turn question. Human annotators then reviewed and rewrote these initial expansions to ensure realistic conversational flow and to remove redundant or irrelevant follow-ups. A subset of the reviewed dialogues was further extended to 4–5 turns by annotators to create longer conversational examples. All dialogues underwent manual review to verify intent preservation and to avoid inadvertent information leakage across turns. Three annotators participated in this stage; their results were cross-checked to maintain consistency.

\subsection{Atomic fact construction and alignment}
\label{sec:qa_atomic_alignment}
We decomposed ground-truth answers into atomic factual statements (facts) through an annotation-assisted process, followed by manual alignment. The decomposition step produced concise, independently verifiable factual units. These units were reviewed in two manual rounds to identify missing information, over-bundling (multiple claims combined), or unsupported content. Round 1 focused on individual corrections and completeness checks and required about \textbf{200 person-hours} in total, with an average of 10–20 minutes per item. Round 2 consisted of cross-review among annotators to resolve disagreements and reach consensus, taking an additional \textbf{100 person-hours}.

\subsection{Human benchmark for consistency evaluation}
\label{sec:qa_consistency_human}
Three annotators independently labeled atomic statement pairs (atomic statement vs.~free-form answer). 
From 100 dialogues, we obtained 1,687 atomic evaluation items (Fig.~\ref{fig:alignment}). Inter-annotator agreement was substantial: pairwise Cohen's~$\kappa$ of \textbf{0.60}, \textbf{0.62}, and \textbf{0.63}; Fleiss'~$\kappa=\,$\textbf{0.62}. Gold labels were created by majority vote.  Agreement between individual annotators and the gold labels was higher: Cohen's~$\kappa$ of \textbf{0.80}, \textbf{0.79}, and \textbf{0.81} (mean \textbf{0.80}), showing the gold labels are \emph{highly consistent} with each expert judgment.

\subsection{Summary}
\label{sec:qa_summary}
We adopted a streamlined, model-assisted annotation workflow. For multi-turn data, the model first generated initial 2–3-turn expansions; annotators then reviewed, rewrote, and selectively extended dialogues to 4–5 turns. Single-turn LFQA items were written from authoritative sources and manually validated. Atomic fact decomposition was performed and followed by two rounds of manual alignment and cross-review. At every stage, independent checks were performed by annotators with finance or computer science backgrounds. The combined process yields a high-quality benchmark suitable for automated factuality evaluation and for reproducible extension.

\section{Empirical Validation Of Dialogue Naturalness}
\label{app:naturalness_validation}

While our data construction pipeline incorporates rigorous human-in-the-loop review (as detailed in Appendix~\ref{app:ad}), relying on LLM-assisted expansion for multi-turn sequences entails a potential risk of introducing synthetic artifacts or stylistic biases. To assess the naturalness of our generated dialogues and ensure they align with real-world interaction patterns, we conducted a \textbf{Complex Scenario Suitability Study} following the protocol established by StructFlowBench~\citep{li2025structflowbench}.

To ensure an equitable comparison, we aligned our evaluation protocol with established baselines. Given that the reference benchmark, StructFlowBench~\citep{li2025structflowbench} and MT-Bench-101~\citep{bai2024mt}, utilize GPT-4o for dialogue generation, we similarly selected the dialogue generated by GPT-4o within \textsc{KnowMT-Bench} for this assessment. Following the experimental setup of StructFlowBench, we randomly sampled 50 dialogues from each benchmark and employed GPT-4o as an impartial evaluator. The assessment was conducted across three distinct dimensions derived from the StructFlowBench rubric, rated on a Likert scale of 1--5:

\begin{itemize}
    \item \textbf{Logical Coherence:} Measures whether the dialogue maintains semantic consistency and logical progression without abrupt or unjustified contextual shifts.
    \item \textbf{Goal Clarity:} Assesses the transparency of the user's intent and the system's adherence to the task objective throughout the interaction.
    \item \textbf{Transition Naturalness:} Evaluates the fluidity of inter-turn transitions, specifically detecting mechanical or forced phrasings typical of synthetic text.
\end{itemize}

Furthermore, to quantify the distributional alignment with high-quality data, we also introduce the \textbf{Confusion Factor (CF)} from StructFlowBench. This composite indicator is defined as the proportion of dialogue samples that achieve an average score of $\ge 4.0$ across the three dimensions, serving as a robust proxy for the density of high-fidelity interactions.

The comparative results are presented in Table~\ref{tab:naturalness_study}. \textsc{KnowMT-Bench} achieves an overall mean score of 4.43, which is comparable to StructFlowBench (4.47) and MT-Bench-101 (4.51). Crucially, in terms of the Confusion Factor, our benchmark scores \textbf{0.78}, surpassing MT-Bench-101 (0.74) and approaching the specialized StructFlowBench (0.82). 

These findings substantiate that the multi-turn sequences in \textsc{KnowMT-Bench}, despite utilizing model-assisted expansion, possess high logical coherence and transition fluidity. The high CF score further confirms that our human verification protocols effectively filtered out low-quality synthetic artifacts, yielding a benchmark that faithfully reflects the complexity and naturalness of real-world knowledge-intensive interaction between humans and LLMs.

\begin{table*}[t]
\centering
\caption{Results of the Complex Scenario Suitability Study. We compare \textsc{KnowMT-Bench} against StructFlowBench and MT-Bench-101 across three qualitative dimensions and the composite Confusion Factor (CF). All scores are on a 1-5 scale except for CF, which is a ratio.}
\label{tab:naturalness_study}
\begin{tabular}{lccccc}
\toprule
\textbf{Dataset} & \textbf{Logical} & \textbf{Goal} & \textbf{Transition} & \textbf{Overall} & \textbf{Confusion} \\
 & \textbf{Coherence} & \textbf{Clarity} & \textbf{Naturalness} & \textbf{Mean} & \textbf{Factor (CF)} \\
\midrule
StructFlowBench~\citep{li2025structflowbench} & 4.78 & 4.46 & 4.18 & 4.47 & 0.82 \\
MT-Bench-101~\citep{bai2024mt} & 4.50 & 4.52 & 4.50 & 4.51 & 0.74 \\
\midrule
\textbf{KnowMT-Bench (Ours)} & 4.72 & 4.46 & 4.12 & 4.43 & 0.78 \\
\bottomrule
\end{tabular}%
\end{table*}

\section{Detailed experiment setting}
\label{exp setting}
\paragraph{Basic Settings}  
Based on the dataset construction process outlined in the previous section, we generat both multi-turn and single-turn dialogues based on questions. For multi-turn experiments, we use each model's \emph{chat template} to format dialogue history and we set \textit{max new tokens} to \textbf{1024} for each round. For Gemini, we restricted the chain-of-thought (CoT) output length to \textbf{256}. To ensure reproducibility, we applied \textbf{greedy decoding} (temperature=0, top\_p=1), disabling sampling and beam search. For models with CoT reasoning, we standardized answer extraction: (i) if a clear final answer is present, only that answer is retained for evaluation; (ii) if no explicit answer is generated (for example, due to truncation), the entire reasoning output is treated as the answer. This policy was applied uniformly in both multi-turn and single-turn settings.

\paragraph{RAG Settings}  
For retrieval-augmented generation tasks~\citep{chen2024fintextqa, wang2025taming}, we adopt a two-stage retrieve--then--rerank pipeline. We construct a \textit{retrieval corpus} by pooling reference documents from \textbf{all} QA instances, and use this corpus as the retrieval candidate pool. Texts in this pool were segmented using the \textbf{SentenceSplitter} from llama-index\footnote{\url{https://github.com/run-llama/llama_index}} with a chunk size of 512 and an overlap of 128, ensuring consistent coverage of context across segments. We use Qwen3-Embedding-0.6B\footnote{\url{https://huggingface.co/Qwen/Qwen3-Embedding-0.6B}} as the embedding model, and Qwen3-Reranker-0.6B\footnote{\url{https://huggingface.co/Qwen/Qwen3-Reranker-0.6B}} as the reranker~\citep{qwen3embedding}. We chose these two models because some of our experimental setups require the entire dialogue history as the query, which imposes relatively high GPU memory demands, hence, we opted for smaller but more recent models that strike a strong balance between efficiency and performance. We first retrieve \textbf{15} candidate chunks from this pool using FAISS’s IndexFlatL2 over L2-normalized embeddings, then rerank and select the top \textbf{5} chunks for prompt construction. Retrieved chunks were formatted as numbered triple-quoted blocks and concatenated with the user query, and the models were instructed to answer strictly based on this context.

\section{Additional Experiments}
\label{app:additional-experiments}

\begin{table*}[t]
    \small
    \caption{Results for all models. Sections explicitly indicate \textbf{Single-turn}, \textbf{Multi-turn}, and the separate \textbf{Diff} block (Multi vs.\ Single). \textbf{Bold} values mark, \emph{within each block}, the best score per column (for $\uparrow$ higher is better; for $\downarrow$ lower is better). Green in \textbf{Diff} indicates improvement and red indicates degradation.}
    \renewcommand{\arraystretch}{0.8} %
    \label{tab:main_results}
    \centering
    \setlength{\tabcolsep}{7pt}
\begin{tabular}{lccc|ccc|cc}
    \toprule
    & \multicolumn{3}{c|}{\textbf{Factuality}} & 
      \multicolumn{3}{c|}{\textbf{Hallucination}} & 
      \multicolumn{2}{c}{\textbf{Efficiency}} \\
    \textbf{Model} & {$\mathbf{R_f}$ (\%) $\uparrow$} & {$\mathbf{P_f}$ (\%) $\uparrow$} & {$\mathbf{S_f}$ (\%) $\uparrow$} & 
    {$\mathbf{R_m}$ (\%) $\downarrow$} & {$\mathbf{P_{fc}}$ (\%) $\downarrow$} & {$\mathbf{S_h}$ (\%) $\downarrow$} & 
    {$\mathbf{D_f}$ $\downarrow$} & {$\mathbf{D_h}$ $\uparrow$} \\
    \midrule
    \multicolumn{9}{l}{\textbf{Single-turn}} \\
    Gemini-2.5-Pro & 56.95 & 18.10 & 23.40 & 11.30 & \textbf{3.33} & 3.03 & 674 & \textbf{1614} \\
        Gemini-3.1-Pro & \textbf{57.37} & 24.27 & 28.24 & 7.71 & 3.39 & 2.37 & 280 & 706 \\

    GPT-4o & 41.94 & \textbf{32.22} & \textbf{28.99} & 6.97 & 3.47 & \textbf{2.29} & \textbf{226} & 505 \\
    GPT-4o-mini & 39.49 & 30.72 & 26.89 & 6.75 & 3.99 & 2.38 & 255 & 545 \\
    DeepSeek-R1 & 51.61 & 17.62 & 21.55 & 14.12 & 4.45 & 4.18 & 604 & 1364 \\
    DeepSeek-V3 & 49.02 & 21.02 & 24.33 & 9.63 & 3.85 & 3.24 & 339 & 738 \\
    Qwen-2.5-72B & 43.51 & 22.89 & 24.30 & 8.15 & 3.51 & 2.41 & 356 & 822 \\
    Llama-3.3-70B & 41.44 & 23.92 & 24.06 & 9.17 & 4.95 & 3.48 & 324 & 656 \\
    Qwen-2.5-32B & 38.24 & 23.61 & 22.91 & 8.78 & 3.41 & 2.77 & 365 & 683 \\
    QwQ-32B & 38.67 & 23.95 & 21.34 & \textbf{6.46} & 5.38 & 2.76 & 526 & 918 \\
    Qwen-2.5-14B & 36.94 & 24.23 & 22.77 & 9.41 & 4.26 & 3.32 & 296 & 512 \\
    Llama-3.1-8B & 33.28 & 20.63 & 18.42 & 8.64 & 11.95 & 4.03 & 404 & 588 \\
    Qwen-2.5-7B & 34.27 & 20.87 & 19.86 & 8.95 & 4.71 & 3.52 & 367 & 591 \\
\midrule
    \multicolumn{9}{l}{\textbf{Multi-turn}} \\
    Gemini-2.5-Pro & 51.65 & 17.66 & 22.09 & 12.89 & 4.77 & 4.32 & 697 & \textbf{1403} \\
        Gemini-3.1-Pro & \textbf{55.38} & 24.12 & 27.89 & 8.79 & \textbf{3.30} & \textbf{2.36} & 303 & 734 \\

    GPT-4o & 42.59 & \textbf{29.17} & \textbf{28.08} & \textbf{7.18} & 3.55 & 2.47 & \textbf{255} & 553 \\
    GPT-4o-mini & 41.11 & 26.88 & 25.80 & 7.88 & 4.09 & 2.84 & 344 & 782 \\
    DeepSeek-R1 & 50.58 & 15.44 & 19.89 & 14.08 & 4.71 & 4.45 & 626 & 1175 \\
    DeepSeek-V3 & 48.92 & 15.82 & 19.49 & 10.20 & 4.00 & 3.45 & 444 & 877 \\
    Qwen-2.5-72B & 44.29 & 17.78 & 20.31 & 9.26 & 3.35 & 2.81 & 487 & 902 \\
    Llama-3.3-70B & 41.11 & 20.45 & 21.37 & 10.51 & 5.49 & 4.26 & 438 & 757 \\
    Qwen-2.5-32B & 40.83 & 20.72 & 22.05 & 9.35 & 3.65 & 3.12 & 434 & 743 \\
    QwQ-32B & 39.66 & 19.77 & 19.31 & 10.11 & 5.40 & 3.91 & 468 & 810 \\
    Qwen-2.5-14B & 38.78 & 21.21 & 21.58 & 10.04 & 4.11 & 3.17 & 428 & 824 \\
    Llama-3.1-8B & 35.25 & 19.06 & 18.67 & 9.50 & 7.14 & 4.16 & 474 & 748 \\
    Qwen-2.5-7B & 36.30 & 16.18 & 17.68 & 10.74 & 4.36 & 3.82 & 544 & 843 \\
\midrule
    \multicolumn{9}{l}{\textbf{Diff (Multi vs.\ Single)}} \\
    Gemini-2.5-Pro & {\textcolor{red}{-9.3\%}} & {\textcolor{red}{-2.4\%}} & {\textcolor{red}{-5.6\%}} & {\textcolor{red}{+14.0\%}} & {\textcolor{red}{+43.2\%}} & {\textcolor{red}{+42.3\%}} & {\textcolor{red}{+3.4\%}} & {\textcolor{red}{-13.0\%}} \\
        Gemini-3.1-Pro & {\textcolor{red}{-3.5\%}} & {\textcolor{red}{-0.6\%}} & {\textcolor{red}{-1.2\%}} & {\textcolor{red}{+14.0\%}} & {\textcolor{green}{-2.7\%}} & {\textcolor{green}{-0.5\%}} & {\textcolor{red}{+7.9\%}} & {\textcolor{green}{+4.0\%}} \\

    GPT-4o & {\textcolor{green}{+1.6\%}} & {\textcolor{red}{-9.5\%}} & {\textcolor{red}{-3.1\%}} & {\textcolor{red}{+2.9\%}} & {\textcolor{red}{+2.2\%}} & {\textcolor{red}{+7.8\%}} & {\textcolor{red}{+12.8\%}} & {\textcolor{green}{+9.5\%}} \\
    GPT-4o-mini & {\textcolor{green}{+4.1\%}} & {\textcolor{red}{-12.5\%}} & {\textcolor{red}{-4.1\%}} & {\textcolor{red}{+16.6\%}} & {\textcolor{red}{+2.5\%}} & {\textcolor{red}{+19.6\%}} & {\textcolor{red}{+34.6\%}} & {\textcolor{green}{+43.5\%}} \\
    DeepSeek-R1 & {\textcolor{red}{-2.0\%}} & {\textcolor{red}{-12.3\%}} & {\textcolor{red}{-7.7\%}} & {\textcolor{green}{-0.3\%}} & {\textcolor{red}{+5.8\%}} & {\textcolor{red}{+6.3\%}} & {\textcolor{red}{+3.7\%}} & {\textcolor{red}{-13.8\%}} \\
    DeepSeek-V3 & {\textcolor{red}{-0.2\%}} & {\textcolor{red}{-24.7\%}} & {\textcolor{red}{-19.9\%}} & {\textcolor{red}{+5.9\%}} & {\textcolor{red}{+4.0\%}} & {\textcolor{red}{+6.5\%}} & {\textcolor{red}{+30.8\%}} & {\textcolor{green}{+18.9\%}} \\
    Qwen-2.5-72B & {\textcolor{green}{+1.8\%}} & {\textcolor{red}{-22.3\%}} & {\textcolor{red}{-16.4\%}} & {\textcolor{red}{+13.6\%}} & {\textcolor{green}{-4.6\%}} & {\textcolor{red}{+16.6\%}} & {\textcolor{red}{+36.5\%}} & {\textcolor{green}{+9.8\%}} \\
    Llama-3.3-70B & {\textcolor{red}{-0.8\%}} & {\textcolor{red}{-14.5\%}} & {\textcolor{red}{-11.2\%}} & {\textcolor{red}{+14.6\%}} & {\textcolor{red}{+10.9\%}} & {\textcolor{red}{+22.4\%}} & {\textcolor{red}{+35.3\%}} & {\textcolor{green}{+15.4\%}} \\
    Qwen-2.5-32B & {\textcolor{green}{+6.8\%}} & {\textcolor{red}{-12.2\%}} & {\textcolor{red}{-3.8\%}} & {\textcolor{red}{+6.6\%}} & {\textcolor{red}{+7.1\%}} & {\textcolor{red}{+12.6\%}} & {\textcolor{red}{+18.8\%}} & {\textcolor{green}{+8.8\%}} \\
    QwQ-32B & {\textcolor{green}{+2.6\%}} & {\textcolor{red}{-17.5\%}} & {\textcolor{red}{-9.5\%}} & {\textcolor{red}{+56.5\%}} & {\textcolor{red}{+0.4\%}} & {\textcolor{red}{+41.4\%}} & {\textcolor{green}{-10.9\%}} & {\textcolor{red}{-11.7\%}} \\
    Qwen-2.5-14B & {\textcolor{green}{+5.0\%}} & {\textcolor{red}{-12.5\%}} & {\textcolor{red}{-5.2\%}} & {\textcolor{red}{+6.6\%}} & {\textcolor{green}{-3.6\%}} & {\textcolor{green}{-4.5\%}} & {\textcolor{red}{+44.7\%}} & {\textcolor{green}{+61.0\%}} \\
    Llama-3.1-8B & {\textcolor{green}{+5.9\%}} & {\textcolor{red}{-7.6\%}} & {\textcolor{green}{+1.3\%}} & {\textcolor{red}{+9.9\%}} & {\textcolor{green}{-40.3\%}} & {\textcolor{red}{+3.2\%}} & {\textcolor{red}{+17.4\%}} & {\textcolor{green}{+27.2\%}} \\
    Qwen-2.5-7B & {\textcolor{green}{+5.9\%}} & {\textcolor{red}{-22.5\%}} & {\textcolor{red}{-11.0\%}} & {\textcolor{red}{+20.0\%}} & {\textcolor{green}{-7.4\%}} & {\textcolor{red}{+8.5\%}} & {\textcolor{red}{+48.3\%}} & {\textcolor{green}{+42.7\%}} \\
\bottomrule
    \end{tabular}
    \end{table*}

\subsection{Numerical Results in Main Experiment}
\label{app:numerical_results}

Table~\ref{tab:main_results} presents the detailed numerical results for all models across the three evaluation dimensions. 
A general trend observed is a decline in performance for most models across a majority of metrics when transitioning from single-turn to multi-turn conversations. 
Notably, Factual Recall ($\mathbf{R_f}$) and the token cost per hallucinated fact ($\mathbf{D_h}$) are exceptions, showing improvements for most models.
This may suggest that while multi-turn interactions prompt models to be more comprehensive and cover more ground-truth facts, this often comes at the cost of reduced precision and greater verbosity, which in turn dilutes the density of factual errors.

\subsection{Additional Analysis of Dialogue Structure and Controlled Noise}
\label{app:structure-controlled-noise}

\begin{figure*}[t]
    \centering
    \includegraphics[width=0.95\textwidth]{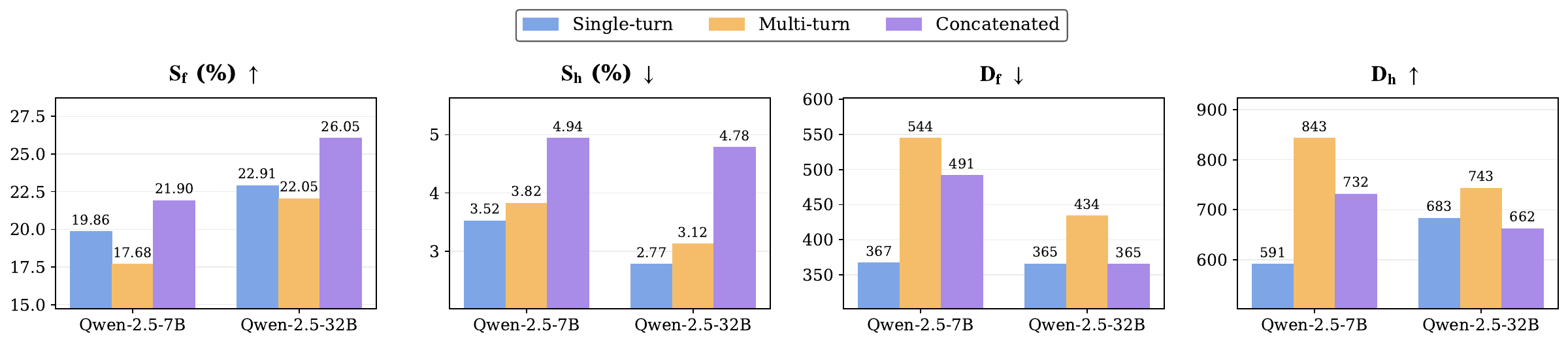}
    \caption{Comparison between standard multi-turn and a concatenated-history setting.}
    \label{fig:5_concatenated}
    \vspace{-5pt}
\end{figure*}

\paragraph{Effect of multi-turn structure.}
To examine the effect of the conversational format, we compare the standard multi-turn setting against a \emph{concatenated-history} variant. Specifically, for each dialogue, we integrate all previous turns and the final-turn question into a single long prompt, instructing the model to regenerate the final-turn answer. As shown in Figure~\ref{fig:5_concatenated}, this variant exhibits a distinct behavioural shift: it achieves the highest factual performance, surpassing even the single-turn baseline, and significantly improves efficiency. However, this comes at the cost of increased hallucination. This suggests that the multi-turn template imposes a conservative constraint, making the model more cautious and verbose. Removing this structure yields more efficient fact delivery but at a higher error risk, indicating that the format primarily shapes generation strategy.

\begin{figure*}[h]
    \centering
    \includegraphics[width=0.95\textwidth]{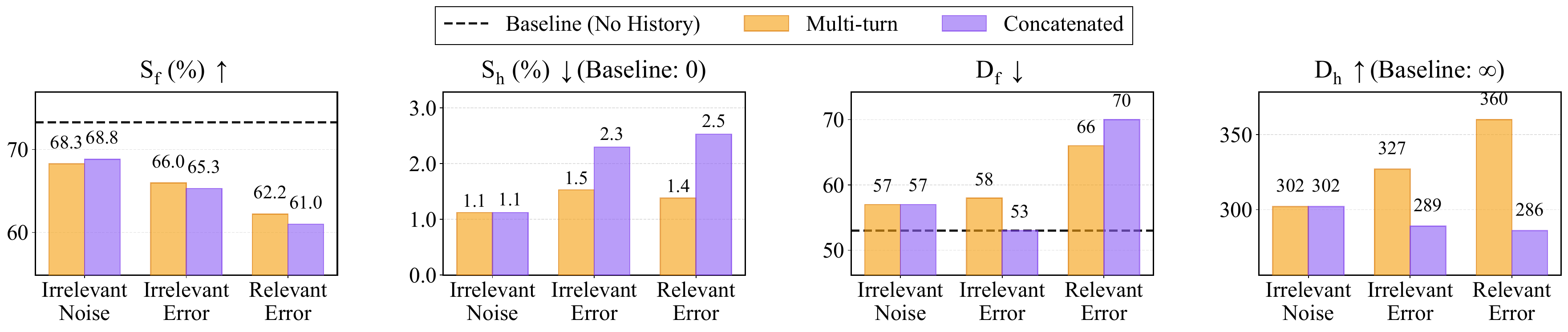}
    \caption{Performance degradation under controlled noise, comparing standard multi-turn and concatenated-history settings.}
    \label{fig:app_controlled_noise_with_concat}
    \vspace{-5pt}
\end{figure*}

\paragraph{Controlled noise injection.}
We also report the concatenated-history version of the controlled noise-injection experiment, which further isolates how noise affects known facts. We curate a ``knowledge-verified'' baseline by filtering for 85 instances where GPT-4o exhibits good factual mastery in the single-turn setting ($\mathbf{S_f} \ge 60\%, \mathbf{S_h} = 0$). Keeping the final question, we manipulate \emph{only the content of the previous turn} to construct three distinct history types: \emph{Irrelevant Noise} (correct but unrelated), \emph{Irrelevant Error} (incorrect and unrelated), and \emph{Relevant Error} (incorrect and related). As shown in Figure~\ref{fig:app_controlled_noise_with_concat}, we evaluate these perturbations under both the standard multi-turn and concatenated-history settings. The results reveal a clear degradation pattern: as history noise becomes more severe, factuality drops monotonically while hallucinations rise sharply. Notably, \emph{relevant errors} induce the most catastrophic degradation, accompanied by a significant deterioration in information delivery efficiency. This suggests that models become confused and verbose when struggling to reconcile conflicting context with their internal knowledge, confirming that contextual noise, especially relevant misinformation, is a key determinant of knowledge grounding.

\subsection{Bidirectional History Substitution}
\label{app:history-substitution}
We further examine natural variations in history quality by conducting a bidirectional substitution experiment. In the upgrading direction, conditioning weaker models on histories generated by a stronger model (GPT-4o) significantly improves factuality and reliability, while also increasing information density. This confirms that a cleaner context directly boosts performance. In the downgrading direction, where GPT-4o is conditioned on weaker histories (Qwen-2.5-7B), we observe a distinct disruption effect: the strong model becomes unconstrained and verbose, leading to higher factuality but at the severe cost of significantly increased hallucinations. This demonstrates that high-quality history acts as a necessary guardrail for reliability; without it, even strong models can degrade into aggressive but ungrounded generation. Additionally, the substitution results support our dynamic evaluation design: allowing each model to generate its own dialogue history better reflects its actual multi-turn behavior than conditioning it on manually annotated histories or histories produced by another model.

\begin{figure*}[h]
    \centering
    \includegraphics[width=0.95\textwidth]{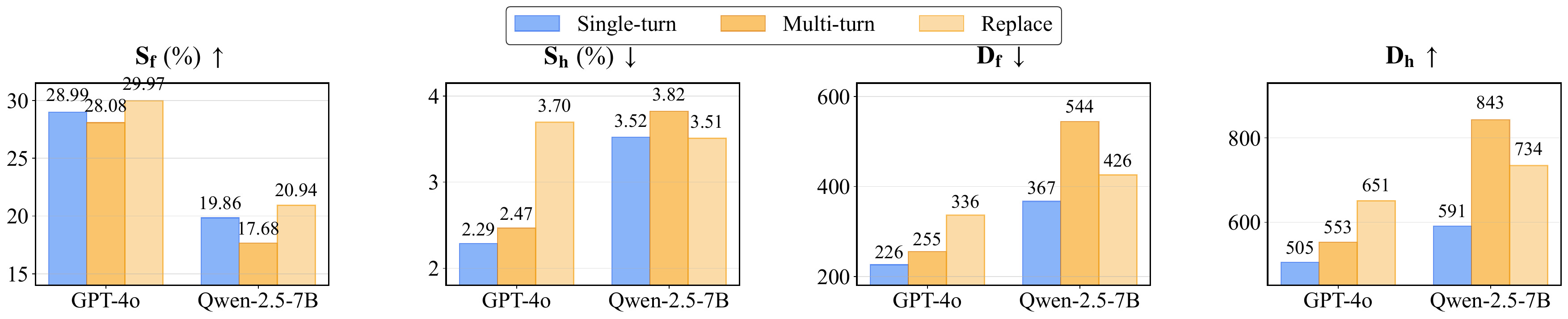}
    \caption{Bidirectional history substitution experiment between a strong model (GPT-4o) and a weaker model (Qwen-2.5-7B). It also underscores that our dynamic evaluation setting more faithfully captures real model performance.}
    \label{fig:5_replace}
    \vspace{-5pt}
\end{figure*}

\subsection{Four-Quadrant Analysis}
\label{app:four-quadrant-analysis}

\begin{figure}[t]
    \centering
    \includegraphics[width=0.48\textwidth]{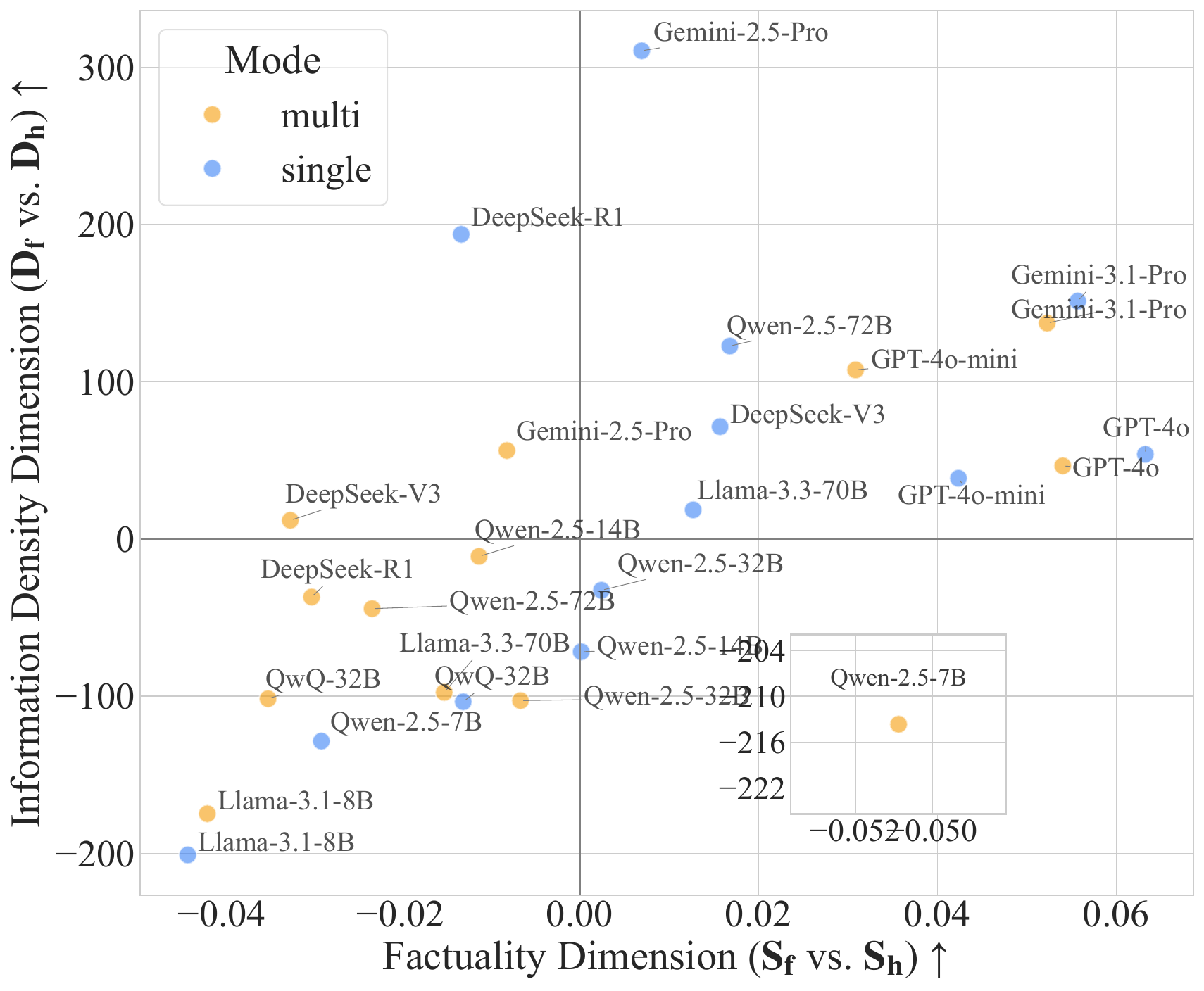}
    \caption{Model performance mapped onto the four-quadrant framework. The x-axis captures overall factuality, derived from the correlation between $\mathbf{S_f}$ and $\mathbf{S_h}$. The y-axis measures performance against the efficiency trade-off baseline, derived from the correlation between $\mathbf{D_f}$ and $\mathbf{D_h}$.}
    \label{fig:quadrant_full}
\end{figure}

While the main text analyzes performance through separate scatter plots (Figure~\ref{fig:4_all_performance_plots}), the four-quadrant framework in Figure~\ref{fig:quadrant_full} offers a synthesized, holistic view. This framework is constructed from the two empirically established regression baselines discussed in Section~\ref{sec:main_results}.

The framework defines two new orthogonal dimensions based on these trends. The \textbf{Factuality Dimension (x-axis)} is derived from the synergistic relationship between $\mathbf{S_f}$ and $\mathbf{S_h}$ (Figure~\ref{fig:4_all_performance_plots}a). The goal is to create a single score where movement towards the plot's bottom-right region (higher $\mathbf{S_f}$, lower $\mathbf{S_h}$) is considered an improvement. Since the data's primary trend, captured by the top-left-to-bottom-right regression line, aligns with this desired trajectory, we use it as a directional basis. To quantify progress along this trajectory, we construct a baseline that is \textbf{perpendicular} to the regression line and passes through the data centroid. The 'Factuality Dimension' score is then the signed distance of each data point \textbf{to} this perpendicular baseline, with points on the bottom-right side receiving higher scores. This metric thus holistically captures a model's overall factuality.

The \textbf{Information Delivery Efficiency Dimension (y-axis)} is derived from the relationship between $\mathbf{D_f}$ and $\mathbf{D_h}$ (Figure~\ref{fig:4_eff_vs_eff}). Its value is the vertical residual: the signed distance from a data point to the regression line along the $\mathbf{D_h}$ axis. Points above the line have positive values. A positive score thus quantifies the degree to which a model's token cost per contradicted fact ($\mathbf{D_h}$) exceeds the expectation set by its cost per correct fact ($\mathbf{D_f}$), measuring its ability to push the efficiency trade-off boundary. The detailed mathematical derivation of these dimensions is provided in Appendix~\ref{app:metric_correlation_analysis}.

This synthesized view crystallizes the model behaviors discussed previously. For instance, \textbf{GPT-4o} occupies the right-hand side of the plot, confirming its strong factuality profile (high x-axis value) and its adherence to the established efficiency trade-off (y-axis value near zero). The Gemini results show that the newer \textbf{Gemini-3.1-Pro} has stronger factual capability than its predecessor \textbf{Gemini-2.5-Pro} in both single-turn and multi-turn settings, while \textbf{Gemini-2.5-Pro} (in the single-turn setting) remains distinguished by a high positive y-axis value because its $\mathbf{D_h}$ is exceptionally high for its given $\mathbf{D_f}$. The \textbf{Qwen family} exhibits clear scaling effects, with larger models generally moving towards the upper-right. The transition to the more demanding multi-turn dialogue setting, however, challenges the models, causing a noticeable shift for most towards the lower-left, underscoring a degradation in both their overall factuality and their performance relative to the efficiency baseline.

\subsection{On the Non-Triviality of Observed Metric Correlations}
\label{app:metric_correlation_analysis}

We demonstrate that the empirically observed correlations specifically, the negative correlation between factuality ($\mathbf{S_f}$) and hallucination ($\mathbf{S_h}$), and the positive correlation between efficiency metrics ($\mathbf{D_f}$ and $\mathbf{D_h}$)—are non-trivial findings about model behavior, not mathematical artifacts of the metric definitions.

\paragraph{Factuality vs. Hallucination ($\mathbf{S_f}$ vs. $\mathbf{S_h}$)}
For each instance $j$, consider $s_f^{(j)} = \mathrm{HM}(p_f^{(j)}, r_f^{(j)})$ and $s_h^{(j)} = \mathrm{HM}(p_{fc}^{(j)}, r_m^{(j)})$; the dataset-level metrics are $\mathbf{S_f}=\langle s_f^{(j)}\rangle_{\mathcal D}$ and $\mathbf{S_h}=\langle s_h^{(j)}\rangle_{\mathcal D}$. At the instance level, their components are constrained because a ground-truth fact cannot be simultaneously supported and contradicted ($\mathcal{F}_j^+ \cap \mathcal{F}_j^- = \varnothing$), and a generated statement cannot be simultaneously correct and false ($\mathcal{S}_j^+ \cap \mathcal{S}_j^- = \varnothing$). This imposes constraints such as $r_f^{(j)} + r_m^{(j)} \le 1$ and $p_f^{(j)} + p_{fc}^{(j)} \le 1$.

However, these "sum-to-at-most-one" constraints on the components do not enforce a necessary monotonic relationship between the final F1 scores, $S_f$ and $S_h$. We demonstrate this with a minimal counterexample. Consider a dataset with one instance ($|\mathcal{D}|=1$), 10 ground-truth facts ($|\mathcal{F}|=10$), and a model generating 10 statements ($|\mathcal{S}|=10$).
\begin{itemize}
    \item \textbf{Case A:} The model correctly covers 5 facts with 5 statements and makes no contradictions ($|\mathcal{F}^+|=5, |\mathcal{S}^+|=5, |\mathcal{F}^-|=0, |\mathcal{S}^-|=0$). This yields $\mathbf{R_f}=0.5, \mathbf{P_f}=0.5 \implies \mathbf{S_f}=0.5$, and $\mathbf{R_m}=0, \mathbf{P_{fc}}=0 \implies \mathbf{S_h}=0$.
    \item \textbf{Case B:} The model's factuality remains the same ($|\mathcal{F}^+|=5, |\mathcal{S}^+|=5 \implies \mathbf{S_f}=0.5$), but its other 5 statements are now false and contradict 5 distinct ground-truth facts ($|\mathcal{F}^-|=5, |\mathcal{S}^-|=5$). This yields $\mathbf{R_m}=0.5, \mathbf{P_{fc}}=0.5 \implies \mathbf{S_h}=0.5$.
\end{itemize}
Since $\mathbf{S_f}$ can remain constant while $\mathbf{S_h}$ varies, no deterministic relationship (e.g., $\mathbf{S_h} = f(\mathbf{S_f})$) is imposed by the metric design. Therefore, the empirically observed negative correlation reflects a genuine behavioral pattern of current models, not an algebraic necessity.

\paragraph{Information Delivery Efficiency ($D_f$ vs. $D_h$)}
A similar analysis applies to the efficiency metrics, $\mathbf{D_f}(j) = \frac{T(a_j)}{|\mathcal{F}_j^+|}$ and $\mathbf{D_h}(j) = \frac{T(a_j)}{|\mathcal{F}_j^-|}$. While they share the same numerator (token count $T(a_j)$), their denominators are controlled by disjoint sets of ground-truth facts, $|\mathcal{F}_j^+|$ and $|\mathcal{F}_j^-|$, which are only loosely constrained by $|\mathcal{F}_j^+| + |\mathcal{F}_j^-| \le |\mathcal{F}_j|$. A necessary positive correlation can be falsified by demonstrating that one metric can be held constant while the other varies.

Consider a fixed-length response with $T(a)=100$ tokens and a context of $|\mathcal{F}|=10$ facts.
\begin{itemize}
    \item Let's fix the number of correctly covered facts at $|\mathcal{F}^+|=5$. This fixes $\mathbf{D_f} = 100/5 = 20$. The number of contradicted facts, $|\mathcal{F}^-|$, can still vary from $0$ to $5$. As $|\mathcal{F}^-|$ changes, $\mathbf{D_h}$ takes values from $\infty$ (or the smoothed maximum) down to $100/5=20$. Thus, $\mathbf{D_f}$ is constant while $\mathbf{D_h}$ varies.
    \item Conversely, let's fix the number of contradicted facts at $|\mathcal{F}^-|=2$. This fixes $\mathbf{D_h} = 100/2=50$. The number of correct facts, $|\mathcal{F}^+|$, can still vary from $0$ to $8$. As $|\mathcal{F}^+|$ changes, $\mathbf{D_f}$ varies from $\infty$ down to $100/8=12.5$. Thus, $\mathbf{D_h}$ is constant while $\mathbf{D_f}$ varies.
\end{itemize}
This independence shows that the shared numerator $T(a_j)$ is a potential confounding variable but does not create a deterministic relationship. The observed strong positive correlation between $\mathbf{D_f}$ and $\mathbf{D_h}$ is therefore an empirical finding about models' tendency towards uniform verbosity, not a mathematical artifact.

\paragraph{Effect of Smoothing}
Our smoothing procedure for zero-denominator cases (e.g., when $|\mathcal{F}_j^+|=0$) replaces the undefined value with a dataset-level maximum. This imputes a constant for the metric on that specific instance, which does not establish a functional link between metrics. In summary, any observed systematic correlation between ($\mathbf{S_f}$, $\mathbf{S_h}$) or ($\mathbf{D_f}$, $\mathbf{D_h}$) should be interpreted as an empirical pattern reflecting inherent trade-offs in model behavior, not as a mechanical coupling arising from the metric design.

\begin{table*}[t!]
\small
\caption{Comparison of model performance using traditional n-gram-based metrics and our proposed core metrics. This table provides the detailed numerical results that complement the scatter plot analysis. \textcolor{green}{Green} indicates improvement, while \textcolor{red}{red} indicates degradation.}
\label{tab:comparison_traditional_metrics}
\centering
\setlength{\tabcolsep}{4pt} %
\begin{tabular}{lcccc|cccc}
\toprule
& \multicolumn{4}{c|}{\textbf{N-gram Based Metrics}} & \multicolumn{4}{c}{\textbf{Our Proposed Core Metrics}} \\
\textbf{Model} & $\mathbf{BLEU-4}$ $\uparrow$ & $\mathbf{R-1}$ $\uparrow$ & $\mathbf{R-2}$ $\uparrow$ & $\mathbf{R-L}$ $\uparrow$ & $\mathbf{S_f}$ (\%) $\uparrow$ & $\mathbf{S_h}$ (\%) $\downarrow$ & $\mathbf{D_f}$ $\downarrow$ & $\mathbf{D_h}$ $\uparrow$ \\
\midrule
\multicolumn{9}{l}{\textbf{Single-turn}} \\
Gemini-2.5-Pro & 1.51 & 13.33 & 3.72 & 12.43 & 23.40 & 3.03 & 674 & 1614 \\
Gemini-3.1-Pro & 3.76 & 20.44 & 6.87 & 18.65 & 28.24 & 2.37 & 280 & 706 \\

GPT-4o & 5.43 & 25.98 & 8.65 & 23.53 & 28.99 & 2.29 & 226 & 505 \\
GPT-4o-mini & 4.79 & 24.98 & 7.89 & 22.55 & 26.89 & 2.38 & 255 & 545 \\
DeepSeek-R1 & 1.56 & 14.19 & 3.77 & 13.22 & 21.55 & 4.18 & 604 & 1364 \\
DeepSeek-V3 & 2.28 & 17.90 & 4.62 & 16.48 & 24.33 & 3.24 & 339 & 738 \\
Qwen-2.5-72B & 3.50 & 21.79 & 6.73 & 19.88 & 24.30 & 2.41 & 356 & 822 \\
Llama-3.3-70B & 4.20 & 23.16 & 7.55 & 20.99 & 24.06 & 3.48 & 324 & 656 \\
Qwen-2.5-32B & 3.46 & 21.73 & 6.19 & 19.57 & 22.91 & 2.77 & 365 & 683 \\
QwQ-32B & 2.18 & 17.43 & 4.39 & 16.10 & 21.34 & 2.76 & 526 & 918 \\
Qwen-2.5-14B & 3.73 & 22.39 & 6.40 & 20.17 & 22.77 & 3.32 & 296 & 512 \\
Llama-3.1-8B & 3.72 & 23.53 & 7.75 & 21.20 & 18.42 & 4.03 & 404 & 588 \\
Qwen-2.5-7B & 3.23 & 21.29 & 6.06 & 19.30 & 19.86 & 3.52 & 367 & 591 \\
\midrule
\multicolumn{9}{l}{\textbf{Multi-turn}} \\
Gemini-2.5-Pro & 1.38 & 13.22 & 3.45 & 12.29 & 22.09 & 4.32 & 697 & 1403 \\
Gemini-3.1-Pro & 3.02 & 19.00 & 5.76 & 17.31 & 27.89 & 2.36 & 303 & 734 \\

GPT-4o & 4.11 & 23.15 & 7.16 & 21.01 & 28.08 & 2.47 & 255 & 553 \\
GPT-4o-mini & 3.36 & 21.55 & 6.43 & 19.61 & 25.80 & 2.84 & 344 & 782 \\
DeepSeek-R1 & 1.28 & 12.81 & 3.08 & 11.97 & 19.89 & 4.45 & 626 & 1175 \\
DeepSeek-V3 & 1.38 & 14.02 & 2.99 & 13.05 & 19.49 & 3.45 & 444 & 877 \\
Qwen-2.5-72B & 2.18 & 17.99 & 5.24 & 16.57 & 20.31 & 2.81 & 487 & 902 \\
Llama-3.3-70B & 3.06 & 21.02 & 6.44 & 19.21 & 21.37 & 4.26 & 438 & 757 \\
Qwen-2.5-32B & 2.55 & 18.86 & 5.35 & 17.28 & 22.05 & 3.12 & 434 & 743 \\
QwQ-32B & 2.09 & 17.18 & 4.00 & 15.69 & 19.31 & 3.91 & 468 & 810 \\
Qwen-2.5-14B & 2.91 & 19.69 & 5.74 & 17.97 & 21.58 & 3.17 & 428 & 824 \\
Llama-3.1-8B & 2.86 & 21.72 & 6.64 & 19.85 & 18.67 & 4.16 & 474 & 748 \\
Qwen-2.5-7B & 2.17 & 17.84 & 4.95 & 16.52 & 17.68 & 3.82 & 544 & 843 \\
\midrule
\multicolumn{9}{l}{\textbf{Diff (Multi vs.\ Single)}} \\
Gemini-2.5-Pro & {\textcolor{red}{-8.5\%}} & {\textcolor{red}{-0.9\%}} & {\textcolor{red}{-7.3\%}} & {\textcolor{red}{-1.1\%}} & {\textcolor{red}{-5.6\%}} & {\textcolor{red}{+42.3\%}} & {\textcolor{red}{+3.4\%}} & {\textcolor{red}{-13.0\%}} \\
Gemini-3.1-Pro & {\textcolor{red}{-19.5\%}} & {\textcolor{red}{-7.0\%}} & {\textcolor{red}{-16.1\%}} & {\textcolor{red}{-7.2\%}} & {\textcolor{red}{-1.2\%}} & {\textcolor{green}{-0.5\%}} & {\textcolor{red}{+7.9\%}} & {\textcolor{green}{+4.0\%}} \\

GPT-4o & {\textcolor{red}{-24.3\%}} & {\textcolor{red}{-10.9\%}} & {\textcolor{red}{-17.3\%}} & {\textcolor{red}{-10.7\%}} & {\textcolor{red}{-3.1\%}} & {\textcolor{red}{+7.8\%}} & {\textcolor{red}{+12.8\%}} & {\textcolor{green}{+9.5\%}} \\
GPT-4o-mini & {\textcolor{red}{-29.8\%}} & {\textcolor{red}{-13.8\%}} & {\textcolor{red}{-18.6\%}} & {\textcolor{red}{-13.1\%}} & {\textcolor{red}{-4.1\%}} & {\textcolor{red}{+19.6\%}} & {\textcolor{red}{+34.6\%}} & {\textcolor{green}{+43.5\%}} \\
DeepSeek-R1 & {\textcolor{red}{-17.9\%}} & {\textcolor{red}{-9.7\%}} & {\textcolor{red}{-18.4\%}} & {\textcolor{red}{-9.5\%}} & {\textcolor{red}{-7.7\%}} & {\textcolor{red}{+6.3\%}} & {\textcolor{red}{+3.7\%}} & {\textcolor{red}{-13.8\%}} \\
DeepSeek-V3 & {\textcolor{red}{-39.4\%}} & {\textcolor{red}{-21.7\%}} & {\textcolor{red}{-35.3\%}} & {\textcolor{red}{-20.8\%}} & {\textcolor{red}{-19.9\%}} & {\textcolor{red}{+6.5\%}} & {\textcolor{red}{+30.8\%}} & {\textcolor{green}{+18.9\%}} \\
Qwen-2.5-72B & {\textcolor{red}{-37.8\%}} & {\textcolor{red}{-17.4\%}} & {\textcolor{red}{-22.2\%}} & {\textcolor{red}{-16.6\%}} & {\textcolor{red}{-16.4\%}} & {\textcolor{red}{+16.6\%}} & {\textcolor{red}{+36.5\%}} & {\textcolor{green}{+9.8\%}} \\
Llama-3.3-70B & {\textcolor{red}{-27.1\%}} & {\textcolor{red}{-9.2\%}} & {\textcolor{red}{-14.7\%}} & {\textcolor{red}{-8.5\%}} & {\textcolor{red}{-11.2\%}} & {\textcolor{red}{+22.4\%}} & {\textcolor{red}{+35.3\%}} & {\textcolor{green}{+15.4\%}} \\
Qwen-2.5-32B & {\textcolor{red}{-26.4\%}} & {\textcolor{red}{-13.2\%}} & {\textcolor{red}{-13.5\%}} & {\textcolor{red}{-11.7\%}} & {\textcolor{red}{-3.8\%}} & {\textcolor{red}{+12.6\%}} & {\textcolor{red}{+18.8\%}} & {\textcolor{green}{+8.8\%}} \\
QwQ-32B & {\textcolor{red}{-4.1\%}} & {\textcolor{red}{-1.4\%}} & {\textcolor{red}{-8.7\%}} & {\textcolor{red}{-2.6\%}} & {\textcolor{red}{-9.5\%}} & {\textcolor{red}{+41.4\%}} & {\textcolor{green}{-10.9\%}} & {\textcolor{red}{-11.7\%}} \\
Qwen-2.5-14B & {\textcolor{red}{-21.9\%}} & {\textcolor{red}{-12.0\%}} & {\textcolor{red}{-10.4\%}} & {\textcolor{red}{-10.9\%}} & {\textcolor{red}{-5.2\%}} & {\textcolor{green}{-4.5\%}} & {\textcolor{red}{+44.7\%}} & {\textcolor{green}{+61.0\%}} \\
Llama-3.1-8B & {\textcolor{red}{-23.0\%}} & {\textcolor{red}{-7.7\%}} & {\textcolor{red}{-14.4\%}} & {\textcolor{red}{-6.4\%}} & {\textcolor{green}{+1.3\%}} & {\textcolor{red}{+3.2\%}} & {\textcolor{red}{+17.4\%}} & {\textcolor{green}{+27.2\%}} \\
Qwen-2.5-7B & {\textcolor{red}{-33.0\%}} & {\textcolor{red}{-16.2\%}} & {\textcolor{red}{-18.4\%}} & {\textcolor{red}{-14.4\%}} & {\textcolor{red}{-11.0\%}} & {\textcolor{red}{+8.5\%}} & {\textcolor{red}{+48.3\%}} & {\textcolor{green}{+42.7\%}} \\
\bottomrule
\end{tabular}
\end{table*}

\subsection{Significance in Main Experiment}
\label{app: significane}

To formally test the directional changes observed in the multi-turn setting (Section~\ref{sec:main_results}), we conduct a series of one-sided Wilcoxon signed-rank tests across all evaluation metrics. For each metric, we pre-specify an expected direction that reflects the qualitative trends discussed in the main text: a degradation in factuality ($\mathbf{S_f}$), hallucination ($\mathbf{S_h}$), and efficiency for correct facts ($\mathbf{D_f}$), and a potential \emph{improvement} in the sparsity of hallucinated facts ($\mathbf{D_h}$). Concretely, the null hypothesis ($H_0$) states that there is no systematic difference between single-turn and multi-turn interactions, whereas the alternative hypothesis ($H_1$) specifies that the median paired difference follows a given direction: Multi-turn $<$ Single-turn for $\mathbf{S_f}$, and Multi-turn $>$ Single-turn for $\mathbf{S_h}$, $\mathbf{D_f}$, and $\mathbf{D_h}$. From an efficiency perspective, larger $\mathbf{D_f}$ means that the model spends more tokens per \emph{correct} fact (worse efficiency), whereas larger $\mathbf{D_h}$ means that each \emph{hallucinated} fact is more token-expensive and thus sparser per token, which is preferable from a safety/robustness standpoint. Table~\ref{tab:fig4_wilcoxon_summary} summarizes the alignment of each model with these metric-specific hypotheses. At a high level, degradation in $\mathbf{S_f}$ and $\mathbf{D_f}$ is statistically significant for almost all models; for hallucination, we observe a consistent tendency for $\mathbf{S_h}$ to increase and for $\mathbf{D_h}$ to increase as well (hallucinations becoming sparser), although the strength of statistical evidence differs across metrics, as detailed below.

\begin{table*}[t]
\centering
\small
\caption{Overall results of the Wilcoxon signed-rank tests for Figure~3. Here, ``Correct'' means that the observed difference between Multi-turn and Single-turn is aligned with the pre-defined expected direction: Multi-turn $<$ Single-turn for $\mathbf{S_f}$ (degradation in factuality) and Multi-turn $>$ Single-turn for $\mathbf{S_h}$ and $\mathbf{D_f}$ (degradation in hallucination and efficiency for correct facts) as well as for $\mathbf{D_h}$ (an improvement, i.e., larger token cost per hallucinated fact). ``Significant'' means one-sided $p<0.05$. Most models move in the expected direction and are significant on $\mathbf{S_f}$, $\mathbf{D_f}$, and $\mathbf{D_h}$, whereas for $\mathbf{S_h}$ the direction is usually correct but the significance is weaker, mainly because $\mathbf{S_h}$ values are close to 0 for many models under both settings.}
\resizebox{1.0\textwidth}{!}{%
\begin{tabular}{lcccc}
\toprule
Metric & Correct \& significant & Correct \& not significant & Opposite \& significant & Opposite \& not significant \\
\midrule
$S_f$ & 9/13 & 3/13 & 0/13 & 1/13 \\
$S_h$ & 3/13 & 8/13 & 0/13 & 2/13 \\
$D_f$ & 11/13 & 1/13 & 0/13 & 1/13 \\
$D_h$ & 10/13 & 0/13 & 0/13 & 3/13 \\
\bottomrule
\end{tabular}}
\label{tab:fig4_wilcoxon_summary}
\end{table*}

\vspace{-10pt}
\paragraph{Factuality ($\mathbf{S_f}$)}
The detailed results in Table~\ref{tab:fig4_wilcoxon_Sf} confirm a statistically significant decline in factual coverage. Among the 13 models, 12 show lower $\mathbf{S_f}$ in the multi-turn setting, and this drop is significant at one-sided $p<0.05$ for 9 of them. Models such as DeepSeek-V3, Qwen-2.5-72B, and Llama-3.3-70B exhibit highly significant declines ($p<0.001$), reinforcing the conclusion that maintaining comprehensive factual recall becomes increasingly difficult as conversation depth grows.

\begin{table*}[t]
\centering
\small
\caption{Wilcoxon signed-rank test results for $\mathbf{S_f}$ (one-sided). The expected direction is Multi-turn $<$ Single-turn. The column ``Aligned with expectation?'' indicates whether the mean difference between Multi-turn and Single-turn follows this expected direction. The ``Significance'' column uses *, ** and *** to denote one-sided $p<0.05$, $p<0.01$ and $p<0.001$, respectively.  MT and ST denote multi-turn and single-turn.}
\begin{tabular}{lrrrrcl}
\toprule
Model & MT mean & ST mean & Diff (M-S) & Fit expectation & $p$-value (one-sided) & Significance \\
\midrule
Gemini-2.5-Pro & 0.22 & 0.23 & -0.01 & Yes & 0.02 & * \\
Gemini-3.1-Pro & 0.28 & 0.28 & -0.00 & Yes & 0.33 &   \\

GPT-4o & 0.28 & 0.29 & -0.01 & Yes & 0.08 &   \\
GPT-4o-mini & 0.26 & 0.27 & -0.01 & Yes & 0.02 & * \\
DeepSeek-R1 & 0.20 & 0.22 & -0.02 & Yes & 0.01 & ** \\
DeepSeek-V3 & 0.19 & 0.24 & -0.05 & Yes & 0.00 & *** \\
Qwen-2.5-72B & 0.20 & 0.24 & -0.04 & Yes & 0.00 & *** \\
Llama-3.3-70B & 0.21 & 0.24 & -0.03 & Yes & 0.00 & *** \\
Qwen-2.5-32B & 0.22 & 0.23 & -0.01 & Yes & 0.06 &   \\
QwQ-32B & 0.19 & 0.21 & -0.02 & Yes & 0.00 & ** \\
Qwen-2.5-14B & 0.22 & 0.23 & -0.01 & Yes & 0.03 & * \\
Llama-3.1-8B & 0.19 & 0.18 & 0.00 & No & 0.77 &   \\
Qwen-2.5-7B & 0.18 & 0.20 & -0.02 & Yes & 0.00 & *** \\
\bottomrule
\end{tabular}
\label{tab:fig4_wilcoxon_Sf}
\end{table*}

\vspace{-10pt}
\paragraph{Hallucination ($\mathbf{S_h}$): Zero-Inflation and Directional Consistency}
The analysis of $\mathbf{S_h}$ (Table~\ref{tab:fig4_wilcoxon_Sh}) requires a more nuanced interpretation. 11 out of 13 models exhibit changes that are aligned with the expected degradation direction (Multi-turn $>$ Single-turn) or remain unchanged, but only three models reach one-sided $p<0.05$. This pattern is primarily driven by the sparsity and zero-inflation of hallucination events: many instances satisfy $\mathbf{S_h} = 0$, so the majority of paired differences satisfy $S_h^{\text{multi}} - S_h^{\text{single}} = 0$. In non-parametric rank tests like Wilcoxon, such ties substantially reduce the effective sample size and thus statistical power. Importantly, among the non-tied pairs where a change does occur, the shift is consistently towards higher hallucination. Consequently, the large one-sided $p$-values mainly reflect a floor effect caused by rare hallucination events, rather than evidence against a degradation trend.

\begin{table*}[t]
\centering
\small
\caption{Wilcoxon signed-rank test results for $\mathbf{S_h}$ (one-sided). The expected direction is Multi-turn $>$ Single-turn. The column ``Aligned with expectation?'' indicates whether the mean difference between Multi-turn and Single-turn follows this expected direction. The ``Significance'' column uses *, ** and *** to denote one-sided $p<0.05$, $p<0.01$ and $p<0.001$, respectively. MT and ST denote multi-turn and single-turn.}
\begin{tabular}{lrrrrcl}
\toprule
Model & MT mean & ST mean & Diff (M-S) & Fit expectation & $p$-value (one-sided) & Significance \\
\midrule
Gemini-2.5-Pro & 0.04 & 0.03 & 0.01 & Yes & 0.00 & *** \\
Gemini-3.1-Pro & 0.02 & 0.02 & -0.00 & No & 0.54 &   \\

GPT-4o & 0.02 & 0.02 & 0.00 & Yes & 0.24 &   \\
GPT-4o-mini & 0.03 & 0.02 & 0.00 & Yes & 0.08 &   \\
DeepSeek-R1 & 0.04 & 0.04 & 0.00 & Yes & 0.17 &   \\
DeepSeek-V3 & 0.03 & 0.03 & 0.00 & Yes & 0.22 &   \\
Qwen-2.5-72B & 0.03 & 0.02 & 0.00 & Yes & 0.19 &   \\
Llama-3.3-70B & 0.04 & 0.03 & 0.01 & Yes & 0.01 & * \\
Qwen-2.5-32B & 0.03 & 0.03 & 0.00 & Yes & 0.10 &   \\
QwQ-32B & 0.04 & 0.03 & 0.01 & Yes & 0.00 & ** \\
Qwen-2.5-14B & 0.03 & 0.03 & -0.00 & No & 0.70 &   \\
Llama-3.1-8B & 0.04 & 0.04 & 0.00 & Yes & 0.35 &   \\
Qwen-2.5-7B & 0.04 & 0.04 & 0.00 & Yes & 0.35 &   \\
\bottomrule
\end{tabular}
\label{tab:fig4_wilcoxon_Sh}
\end{table*}

\paragraph{Information Delivery Efficiency ($\mathbf{D_f}$) and Hallucination Sparsity ($\mathbf{D_h}$)}
The clearest degradation appears in $\mathbf{D_f}$ (Table~\ref{tab:fig4_wilcoxon_Df}): 12 of the 13 models require more tokens per correct fact in the multi-turn setting, and this increase is statistically significant for 11 models ($p<0.05$). This confirms that, as conversations become multi-turn, models systematically become less efficient in how they allocate tokens to correct factual content.

In contrast, the pattern for $\mathbf{D_h}$ (Table~\ref{tab:fig4_wilcoxon_Dh}) reflects an improvement in hallucination sparsity. For 10 models, the token cost per contradicted fact is significantly larger in the multi-turn setting (one-sided $p<0.05$), meaning that hallucinated facts become more token-expensive and thus sparser per token. In other words, when hallucinations do occur, they tend to be more “diluted” within longer responses.

\begin{table*}[t]
\centering
\small
\caption{Wilcoxon signed-rank test results for $\mathbf{D_f}$ (one-sided). The expected direction is Multi-turn $>$ Single-turn, corresponding to a degradation in efficiency for correct facts (more tokens per correct fact). The column ``Aligned with expectation?'' indicates whether the mean difference between Multi-turn and Single-turn follows this expected direction. The ``Significance'' column uses *, ** and *** to denote one-sided $p<0.05$, $p<0.01$ and $p<0.001$, respectively.  MT and ST denote multi-turn and single-turn.}
\begin{tabular}{lrrrrcl}
\toprule
Model & MT mean & ST mean & Diff (M-S) & Fit expectation & $p$-value (one-sided) & Significance \\
\midrule
Gemini-2.5-Pro & 697.13 & 673.89 & 23.24 & Yes & 1.00 &   \\
Gemini-3.1-Pro & 302.54 & 280.41 & 22.12 & Yes & 0.00 & *** \\

GPT-4o & 255.22 & 226.20 & 29.02 & Yes & 0.00 & *** \\
GPT-4o-mini & 343.61 & 255.35 & 88.26 & Yes & 0.00 & *** \\
DeepSeek-R1 & 626.28 & 604.11 & 22.17 & Yes & 0.01 & ** \\
DeepSeek-V3 & 443.92 & 339.39 & 104.52 & Yes & 0.00 & *** \\
Qwen-2.5-72B & 486.51 & 356.38 & 130.14 & Yes & 0.00 & *** \\
Llama-3.3-70B & 438.35 & 324.08 & 114.27 & Yes & 0.00 & *** \\
Qwen-2.5-32B & 433.89 & 365.37 & 68.52 & Yes & 0.00 & *** \\
QwQ-32B & 468.46 & 525.85 & -57.38 & No & 1.00 &   \\
Qwen-2.5-14B & 428.10 & 295.89 & 132.21 & Yes & 0.00 & *** \\
Llama-3.1-8B & 474.29 & 403.93 & 70.36 & Yes & 0.00 & *** \\
Qwen-2.5-7B & 544.48 & 367.15 & 177.33 & Yes & 0.00 & *** \\
\bottomrule
\end{tabular}
\label{tab:fig4_wilcoxon_Df}
\end{table*}

\begin{table*}[t]
\centering
\small
\caption{Wilcoxon signed-rank test results for $\mathbf{D_h}$ (one-sided). The expected direction is Multi-turn $>$ Single-turn, which corresponds to a larger token cost per contradicted fact and hence sparser hallucinations per token (an improvement from a robustness perspective). The column ``Aligned with expectation?'' indicates whether the mean difference between Multi-turn and Single-turn follows this expected direction. The ``Significance'' column uses *, ** and *** to denote one-sided $p<0.05$, $p<0.01$ and $p<0.001$, respectively.  MT and ST denote multi-turn and single-turn.}
\begin{tabular}{lrrrrcl}
\toprule
Model & MT mean & ST mean & Diff (M-S) & Fit expectation & $p$-value (one-sided) & Significance \\
\midrule
Gemini-2.5-Pro & 1403.42 & 1613.53 & -210.10 & No & 1.00 &   \\
Gemini-3.1-Pro & 733.68 & 705.55 & 28.13 & Yes & 0.00 & *** \\

GPT-4o & 552.83 & 504.97 & 47.86 & Yes & 0.00 & *** \\
GPT-4o-mini & 782.07 & 545.17 & 236.90 & Yes & 0.00 & *** \\
DeepSeek-R1 & 1175.41 & 1363.99 & -188.58 & No & 1.00 &   \\
DeepSeek-V3 & 877.23 & 737.92 & 139.31 & Yes & 0.00 & *** \\
Qwen-2.5-72B & 902.07 & 821.54 & 80.54 & Yes & 0.00 & *** \\
Llama-3.3-70B & 757.18 & 655.85 & 101.32 & Yes & 0.00 & *** \\
Qwen-2.5-32B & 743.46 & 683.28 & 60.18 & Yes & 0.00 & *** \\
QwQ-32B & 810.45 & 917.71 & -107.26 & No & 1.00 &   \\
Qwen-2.5-14B & 824.20 & 511.96 & 312.24 & Yes & 0.00 & *** \\
Llama-3.1-8B & 748.37 & 588.27 & 160.10 & Yes & 0.00 & *** \\
Qwen-2.5-7B & 843.06 & 590.67 & 252.38 & Yes & 0.00 & *** \\
\bottomrule
\end{tabular}
\label{tab:fig4_wilcoxon_Dh}
\end{table*}

\subsection{Validation of the Maximum Generation Length}
\label{app:QwQ2048}

Our evaluation includes models specifically designed for long Chain-of-Thought (CoT) reasoning, such as QwQ-32B, Gemini-2.5-Pro and DeepSeek-R1. For these models, a methodological concern is that their inherently verbose reasoning might consume a disproportionate share of the 1024-token generation limit, leaving insufficient space for the final answer. To investigate this, we conduct a validation experiment in both single-turn and multi-turn settings.

\begin{table}[t]
\small
\captionsetup{width=\linewidth}
\centering
    \caption{Performance comparison of QwQ-32B and QwQ-32B-2048. White and gray rows indicate the single-turn and multi-turn setting, respectively.}
\label{tab:qwen_compare_full}
\begin{tabular}{l cccc}
\toprule
Setting & {$\mathbf{S_f}$ (\%) $\uparrow$} & {$\mathbf{S_h}$ (\%) $\downarrow$} & {$\mathbf{D_f}$ $\downarrow$} & {$\mathbf{D_h}$ $\uparrow$} \\
\midrule
\rowcolor{white} QwQ-32B & 21.34 & 2.76 & 526 & 918 \\
\rowcolor{white} QwQ-32B-2048 & 20.24 & 3.82 & 699 & 1545 \\
\midrule[0.3pt] %
\rowcolor{gray!15} QwQ-32B & 19.31 & 3.91 & 468 & 810 \\
\rowcolor{gray!15} QwQ-32B-2048 & 17.61 & 3.92 & 807 & 1532 \\
\bottomrule
\end{tabular}
\end{table}

We compare the standard 1024-token limit QwQ-32B against QwQ-32B-2048, a variant with a 2048-token limit. The results, presented in Table~\ref{tab:qwen_compare_full}, demonstrate a consistent trend across both settings: the extended generation capacity fails to provide a clear advantage. In both single-turn and multi-turn scenarios, increasing the token limit led to a degradation in the $\mathbf{S_f}$ and offered no substantive improvement in $\mathbf{S_h}$. Furthermore, the token efficiency per correct fact ($\mathbf{D_f}$) consistently worsened with the larger budget. This consistent pattern strongly suggests that the model's performance is not primarily constrained by its reasoning, crowding out the answer space. The analysis thus validates our use of 1024 tokens as a sufficient and robust setting for the main experiments, regardless of the conversational context.

\subsection{Comparison to Traditional N-gram-based Metrics}
\label{app:comparison_traditional_metrics}

\begin{figure*}[t]
    \centering
    \begin{subfigure}[t]{0.49\linewidth}
        \centering
        \includegraphics[width=\linewidth]{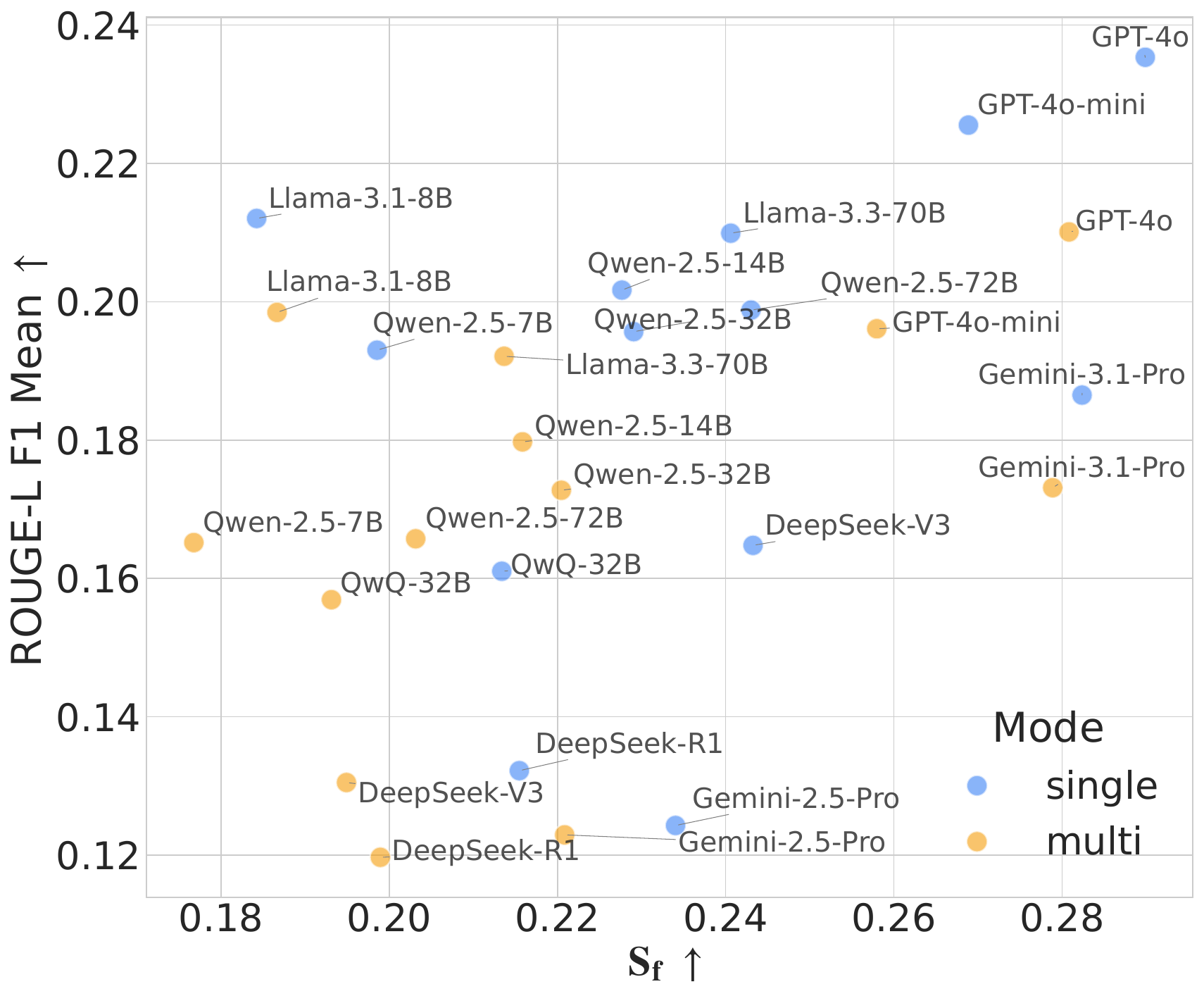}
        \caption{$\mathbf{S_f}$ vs. ROUGE-L}
        \label{fig:appendix_sub_corr_rouge}
    \end{subfigure}\hfill
    \begin{subfigure}[t]{0.49\linewidth}
        \centering
        \includegraphics[width=\linewidth]{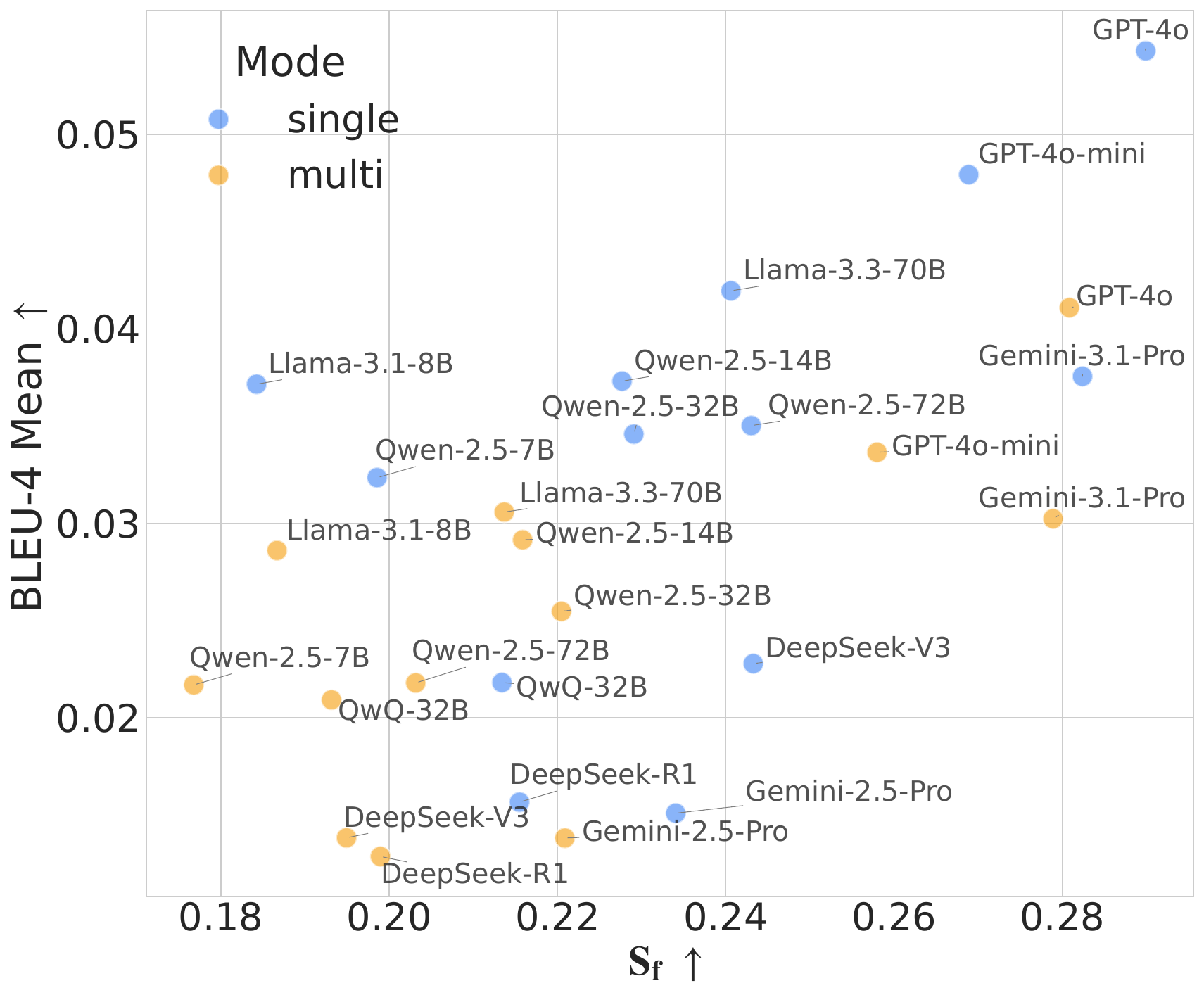}
        \caption{$\mathbf{S_f}$ vs. BLEU-4}
        \label{fig:appendix_sub_corr_bleu}
    \end{subfigure}

    \caption{
        Correlation between $\mathbf{S_f}$ and traditional lexical metrics. While a positive trend is observed, models optimized for CoT reasoning tend to be undervalued by n-gram metrics, potentially due to subtle stylistic differences in their final answers.
    }
    \label{fig:appendix_corr_traditional}
\end{figure*}

To provide further context for our proposed metrics, we analyze the relationship between $\mathbf{S_f}$ and two prevalent n-gram-based metrics: ROUGE-L and BLEU-4. The scatter plots in Figure~\ref{fig:appendix_corr_traditional} map the performance of evaluated models across these metrics. The numerical results are listed in Table~\ref{tab:comparison_traditional_metrics}.

The analysis reveals a positive correlation between $\mathbf{S_f}$ and both ROUGE-L and BLEU-4. This alignment is expected, as higher factual accuracy often coincides with greater lexical overlap with reference texts. This finding suggests that our metric is directionally consistent with established evaluation paradigms.

A closer inspection of the plots, however, reveals a systematic deviation. We find that a cluster of models, particularly those optimized for CoT reasoning, are consistently undervalued by ROUGE-L and BLEU-4 relative to their $\mathbf{S_f}$ scores. Notably, this occurs even after programmatically removing the CoT reasoning steps, with all metrics assessing only the final answer. We hypothesize this discrepancy stems not from the reasoning text itself, but from subtle stylistic artifacts in the final synthesized answer. It is plausible that the CoT generation process implicitly influences the model's final output style, leading to differences in sentence structure or lexical choice compared to the reference. While these stylistic variations may not compromise the underlying facts—which our metric is designed to capture by operating on decomposed statements—they can penalize scores for metrics sensitive to surface-level matching. This observation highlights the value of evaluation frameworks that can disentangle factual correctness from surface-level stylistic choices.

\subsection{Interpretability Analysis}
\label{app:interpretability_analysis}

To analyze how models use dialogue context in the final response, we attribute token-level importance by coupling attention weights with gradients computed only with respect to the last-turn answer. 
For a dialogue $d_k$, let the input token sequence for the final turn be
$X^{(k)}=(x_1,\dots,x_L)$, obtained by concatenating the history $H_k$, the final-turn question $q^{(k)}_{N_k}$, and the model's last-turn answer $a^{(k)}_{N_k}$. 
We denote the index sets of context, question, and answer tokens by $\mathcal{T}_{\mathrm{ctx}}$, $\mathcal{T}_{\mathrm{q}}$, and $\mathcal{T}_{\mathrm{ans}}$, respectively, forming a partition of $\{1,\dots,L\}$.

Let $\theta$ be model parameters. 
We define the target for attribution as the (token-averaged) log-likelihood of the last-turn answer:
\begin{equation}
s^{(k)} \;=\; \sum_{t \in \mathcal{T}_{\mathrm{ans}}} \log p_\theta\!\big(x_t \,\big|\, x_{<t},\, H_k,\, q^{(k)}_{N_k}\big).
\label{eq:target}
\end{equation}
All gradients below are taken with respect to\ this $s^{(k)}$, so they reflect how changing attention would affect the probability of the \emph{last-turn answer only}.

For each transformer layer $l\!\in\!\{1,\dots,L_\ell\}$ and head $h\!\in\!\{1,\dots,H\}$, let $A^{(l,h)}\!\in\!\mathbb{R}^{L\times L}$ be the row-stochastic attention matrix from the final forward pass (rows: query positions; columns: key positions). 
We compute its gradient
\begin{equation}
G^{(l,h)} \;=\; \frac{\partial s^{(k)}}{\partial A^{(l,h)}} \;\in\; \mathbb{R}^{L\times L}.
\end{equation}
We then attribute a \emph{source-side} importance to each token $j$ by aggregating, over layers, heads, and answer query positions $t\!\in\!\mathcal{T}_{\mathrm{ans}}$, the signed gradient-weighted attention received by $j$:
\begin{equation}
\begin{aligned}
I_j
&= \frac{1}{Z}
\sum_{l=1}^{L_\ell}\sum_{h=1}^{H}
\sum_{t\in\mathcal{T}_{\mathrm{ans}}}
A^{(l,h)}_{t j}\;\mathrm{sign}\!\big(G^{(l,h)}_{t j}\big), \\
Z
&= L_\ell \cdot H \cdot |\mathcal{T}_{\mathrm{ans}}|.
\end{aligned}
\label{eq:token_importance}
\end{equation}
A positive $I_j$ indicates that attending to token $j$ increases the likelihood of the last-turn answer, while a negative value indicates an inhibitory effect. 
(As a magnitude-sensitive variant, one may replace $\mathrm{sign}(\cdot)$ with $\mathrm{tanh}(\alpha |G^{(l,h)}_{t j}|)$ or simply $|G^{(l,h)}_{t j}|$; we use the signed version in Eq.~\eqref{eq:token_importance}.)

\begin{figure*}[t]
    \centering
    \includegraphics[width=\linewidth]{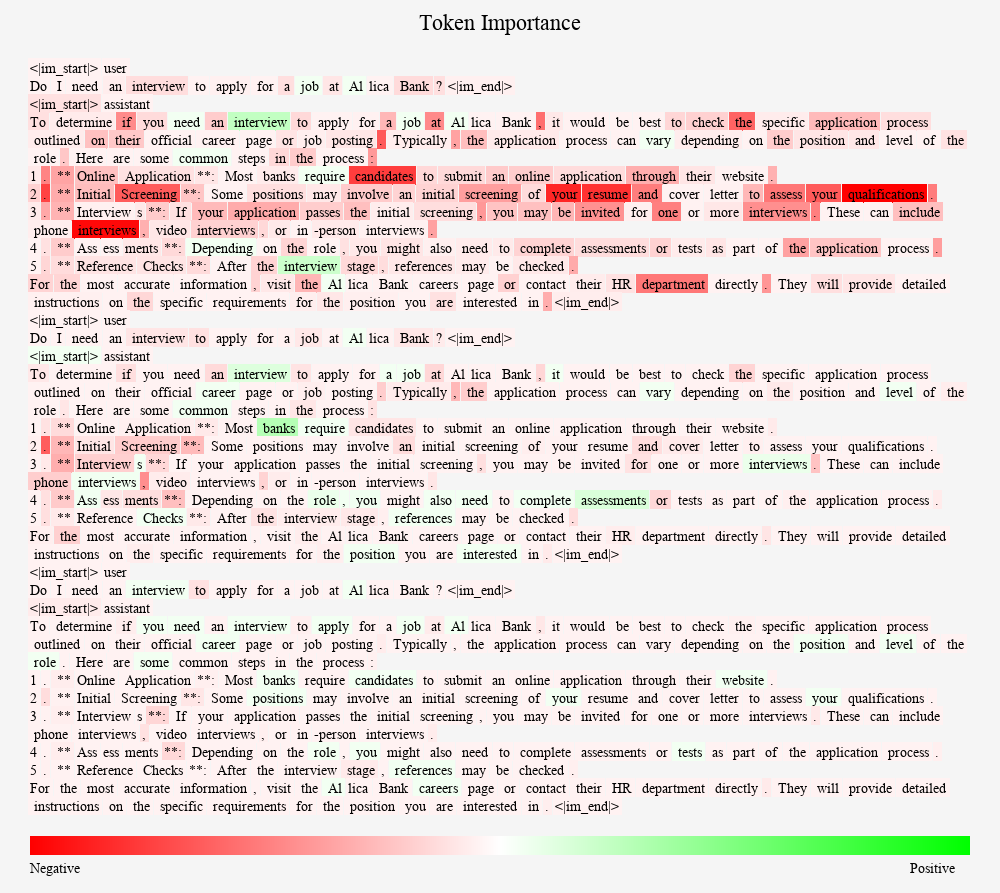}
    \caption{Visualization of contextual token importance. Green indicates tokens that positively contribute to the model's response, while red indicates tokens with a negative influence.}
    \label{fig:appendix_case}
\end{figure*}

In our final reporting, we aggregate only over the answer tokens that realize the useful atomic statement in the last-turn answer.
From $a^{(k)}_{N_k}$ we extract a set of useful atomic statements $\mathcal{S}_k^+$ and use \texttt{Qwen2.5-7B-Instruct} to align each $s\in\mathcal{S}_k^+$ back to its minimal supporting span(s) in $a^{(k)}_{N_k}$, yielding
\begin{equation}\label{eq:t_ans_plus}
\mathcal{T}_{\mathrm{ans}}^{+} \;\subseteq\; \mathcal{T}_{\mathrm{ans}}.
\end{equation}
All gradients are still taken with respect to $s^{(k)}$ (Eq.~\ref{eq:target}), but the aggregation over query positions $t$ is restricted to $t\in\mathcal{T}_{\mathrm{ans}}^{+}$. We define the useful-span source-side importance as
\begin{equation}\label{eq:token_importance_useful}
I_j^{(+)} \;=\; \frac{1}{Z^{(+)}} \sum_{l=1}^{L_\ell} \sum_{h=1}^{H} \sum_{t\in\mathcal{T}_{\mathrm{ans}}^{+}}
A^{(l,h)}_{t j}\;\mathrm{sign}\!\big(G^{(l,h)}_{t j}\big),
\end{equation}
with the normalization
\begin{equation}\label{eq:Z_plus}
Z^{(+)} \;=\; L_\ell \cdot H \cdot \big|\mathcal{T}_{\mathrm{ans}}^{+}\big|.
\end{equation}
We quantify the net contribution of the dialogue context to producing the useful statement(s) via
\begin{equation}\label{eq:ctx_avg_useful}
\overline{I}_{\mathrm{ctx}}^{(+)} \;=\; \frac{1}{\big|\mathcal{T}_{\mathrm{ctx}}\big|}\sum_{j\in \mathcal{T}_{\mathrm{ctx}}} I_j^{(+)} ,
\end{equation}
and analogously define $\overline{I}_{\mathrm{q}}^{(+)}$ and $\overline{I}_{\mathrm{ans}}^{(+)}$ if needed.
Empirically, under this useful-span restriction, the contextual averages for \textbf{Qwen2.5-7B-Instruct} and \textbf{LLaMA3.1-8B-Instruct} are $-0.0025$ and $-0.0029$, indicating that context tokens exert a net negative (noise-like) influence on the helpful parts of the final answer. See Figure~\ref{fig:appendix_case} for a visualization.

\begin{table}[t]
\captionsetup{width=\linewidth}
\setlength{\tabcolsep}{4pt}
\small
\centering
\caption{Impact of Topic Drift on Qwen-2.5-7B. We compare the standard setting against a "Topic Drift" setting where an unrelated question is prepended to the context.}
\label{tab:topic_drift}
\begin{tabular}{lccccc}
\toprule
\textbf{Setting} & \textbf{Mode} & \textbf{$S_f$ (\%)} & \textbf{$S_h$ (\%)} & \textbf{$D_f$} & \textbf{$D_h$} \\
\midrule
Original & Single-turn & 19.86 & 3.52 & 367 & 591 \\
Topic Drift & Single-turn & 19.50 & 3.43 & 367 & 590 \\
\midrule
Original & Multi-turn & 17.68 & 3.82 & 544 & 843 \\
Topic Drift & Multi-turn & 20.48 & 4.81 & 536 & 833 \\
\bottomrule
\end{tabular}%
\end{table}

\subsection{Robustness to Topic Drift}
\label{app:topic_drift}

While \textsc{KnowMT-Bench} primarily targets domain-focused dialogue, real-world interactions sometimes involve spontaneous topic switching or "drift." To probe the impact of such disjointed contexts and ensure our evaluation does not overestimate model robustness, we conducted a controlled experiment simulating topic drift.

We prepended a logically unrelated question (randomly sampled from a disjoint domain) before each question sequence in our standard evaluation set. This setup introduces irrelevant contextual noise, forcing the model to discern the current intent amidst a drifted history. We evaluate Qwen-2.5-7B under this setting compared to the original baseline.

The results are presented in Table~\ref{tab:topic_drift}. 
First, in the single-turn setting, performance remains essentially invariant, suggesting that unrelated prefixes have a negligible impact on the model's immediate instruction following. 
Second, in the multi-turn scenario, we observe a nuanced trade-off: topic drift leads to a simultaneous increase in both Factual Score ($S_f$: $17.68\% \to 20.48\%$) and Hallucination Score ($S_h$: $3.82\% \to 4.81\%$). This indicates that irrelevant context disrupts the model's generation pattern, potentially breaking the "conservative" constraint observed in standard multi-turn dialogues, thereby yielding more facts but at the cost of higher hallucination risk.

Crucially, these findings align with our core analysis in Section~\ref{sec:pd}, confirming that the detrimental impact of \textit{relevant misinformation} outweighs that of irrelevant noise or errors. Consequently, the exclusion of spontaneous topic drift does not substantially compromise the validity of our evaluation in knowledge-intensive scenarios, where robustness against relevant hallucinations is the primary concern.

\section{License Statement}

This paper, together with the associated code and dataset, is licensed under the Creative Commons Attribution 4.0 International License (CC BY 4.0).

\section{Topic Taxonomy}
\label{sec:taxonomy}
We manually classified these topics into three categories: \emph{institutions and products}, \emph{policies and events}, and \emph{concepts}. Below we formalize the category descriptions (kept from the original, translated to English) and provide filled examples in a unified \textbf{T/Q/A/R} format.

\subsection{Organizations \& Their Products/Tools}
\label{sec:organizations_products}
\textbf{Category Descriptions} 
This category covers various types of organizations and their products or services. 
For \emph{insurance institutions}, the focus is on product details such as claim standards, premium/payment flows, and required documents, as well as the scope of the institution from official ``About us'' pages. 
For \emph{banks}, examples include the materials required to handle services, the types of deposits, deposit safety in case of failure, and application conditions for specific credit cards. 
For \emph{government departments}, the emphasis lies in service-handling procedures and official responsibilities. 
\emph{Financial institutions (non-bank)} are described through their products and company details, while \emph{other companies} are captured by their main businesses and notable events.

\subsubsection*{Filled Examples — Organizations \& Their Products/Tools}
\textbf{Insurance Institutions}\\
\noindent\textbf{T:} AXA\\
\textbf{Q:} What are the main businesses of AXA?\\
\textbf{A:} AXA's main businesses include property-casualty insurance business, life \& savings business and asset management business.\\
\textbf{R:} \href{https://www.axa.com/en/about-us/what-we-do}{https://www.axa.com/en/about-us/what-we-do}

\medskip
\noindent\textbf{T:} Allstate\\
\textbf{Q:} What information do I need to provide to purchase auto insurance with Allstate?\\
\textbf{A:} When purchasing auto insurance with Allstate, you will need to provide the following information: 1.\ Vehicle Information: the make, model, and vehicle identification number for each car you want to insure. 2.\ Covered Drivers Information: the names of all drivers covered by the policy. 3.\ Current Insurance Information: your existing coverage and limits. 4.\ Personal Information: dates of birth, Social Security numbers, and driver's license numbers for each family member to be included in the policy. 5.\ Driving History: an overview of any past tickets or auto accidents incurred by family members you want to cover on the policy. 6.\ Background Information: whether you own or rent your home, and some details about your assets and income. 7.\ Coverage Types and Limits: the types of coverage you want.\\
\textbf{R:} \href{https://www.allstate.com/resources/information-needed-for-insurance-quote}{https://www.allstate.com/resources/information-needed-for-insurance-quote}

\textbf{Banks}\\
\noindent\textbf{T:} Bank of America\\
\textbf{Q:} What are the cash rewards like on this credit card, Bank of America Customized Cash Rewards?\\
\textbf{A:} The Bank of America Customized Cash Rewards credit card offers cashback as follows: 1.\ 1\% base cashback on all eligible net purchases, meaning you earn \$0.01 for every \$1 spent. 2.\ 2\% cashback on eligible purchases at grocery stores, wholesale clubs, supermarkets, meat and seafood markets, candy, nut, and confectionery stores, dairy stores, and bakeries, which includes the 1\% base cashback plus an additional 1\% bonus, but excludes supermarkets and small stores that sell groceries like drugstores. 3.\ 3\% cashback on purchases in one chosen ``Choice'' category, which includes gas and EV charging stations (default), online shopping, dining, travel, drug stores, or home improvement/furnishings, with the 3\% consisting of the 1\% base cashback plus an additional 2\% bonus. 4.\ Transactions such as balance transfers, cash advances, fees, interest charges, fraudulent transactions, and certain other fees do not earn cashback.\\
\textbf{R:} \href{https://www.bankofamerica.com/credit-cards/terms-and-conditions/?campaignid=4071205&productoffercode=UN&locale=en_US}{https://www.bankofamerica.com/credit-cards/terms-and-conditions/?campaignid=4071205\\\&productoffercode=UN\&locale=en\_US}

\medskip
\noindent\textbf{T:} NatWest Bank\\
\textbf{Q:} What types of personal savings accounts are available at NatWest Bank?\\
\textbf{A:} NatWest Bank's Individual Savings Accounts include the Digital Regular Saver, Fixed Rate ISA, Fixed Term Savings, Flexible Saver, Stocks \& Shares ISA and First Saver.\\
\textbf{R:} \href{https://www.natwest.com/savings.html?intcam=HP-TTB-DEF-Default#productFilter}{https://www.natwest.com/savings.html?intcam=\\HP-TTB-DEF-Default\#productFilter}

\medskip
\noindent\textbf{T:} Silicon Valley Bank\\
\textbf{Q:} Is my deposit safe if Silicon Valley Bank fails?\\
\textbf{A:} If your deposits are with Silicon Valley Bank and meet the requirements of the Federal Deposit Insurance Corporation (FDIC), your deposits are safe. The FDIC provides insurance coverage of up to \$250{,}000 per depositor, including principal and interest. If your deposits exceed this amount, additional coverage may apply based on different account ownership categories. You can use the FDIC's Electronic Deposit Insurance Estimator (EDIE) to verify if your deposits are fully covered. Please note that FDIC insurance does not cover investments such as stocks, bonds, or mutual funds.\\
\textbf{R:} \href{https://www.svb.com/fdic/}{https://www.svb.com/fdic/}

\medskip
\noindent\textbf{T:} Wells Fargo\\
\textbf{Q:} When I can't earn Rewards Points Bonus on my Wells Fargo travel rewards credit card\\
\textbf{A:} You will not earn Rewards Points on the following types of transactions with your travel rewards credit card: 1.\ Cash Advances and Equivalents: This includes ATM transactions, cash advances, money orders, prepaid gift cards, traveler's checks, wire transfers, and balance transfers. 2.\ Disputed or Illegal Transactions: Any purchases that are disputed, illegal, or violate the terms of the Credit Card Account agreement. 3.\ Fees and Interest: Any fees or interest charges that post to your Credit Card Account, such as annual fees, monthly fees, late fees, and returned payment fees. 4.\ Gambling Transactions: This includes any gambling-related transactions, such as online bets or wagers, casino gaming chips, lottery tickets, and off-track wagers.\\
\textbf{R:} \href{https://www.wellsfargo.com/credit-cards/autograph-journey-visa/terms/}{https://www.wellsfargo.com/credit-cards/autograph-journey-visa/terms/}

\textbf{Government Departments}\\
\noindent\textbf{T:} IRS\\
\textbf{Q:} What is the mission of the IRS?\\
\textbf{A:} The IRS mission is to provide America's taxpayers top quality service by helping them understand and meet their tax responsibilities and to enforce the law with integrity and fairness to all.\\
\textbf{R:} \href{https://www.irs.gov/about-irs}{https://www.irs.gov/about-irs}

\medskip
\noindent\textbf{T:} DWP\\
\textbf{Q:} I am dissatisfied with my service at DWP, how do I make a complaint?\\
\textbf{A:} 1.\ If you’d like to complain about any aspect of the service you’ve received, let the office you have been dealing with know as soon as possible. You can contact them by phone, in person or in writing. Universal Credit claimants can also use their journal. 2.\ You need to provide the necessary details, including your National Insurance number (unless you are an employer), your full name, address and contact details, the benefit you are complaining about, what happened, when it happened, and how it affected you, and what you want to happen to resolve the issue. 3.\ You can use the contact details on any recent letters we’ve sent you or use the contact information below. If you live in Northern Ireland, visit the Department for Communities website for more information.\\
\textbf{R:} \href{https://www.gov.uk/government/organisations/department-for-work-pensions/about/complaints-procedure\#contact-the-office-you've-been-dealing\%20with}{https://www.gov.uk/government/organisations/\\department-for-work-pensions/about/complaints-procedure\#contact-the-office-you've-been-dealing\%20with}

\textbf{Financial Institutions (non-bank)}\\
\noindent\textbf{T:} S\&P 500\\
\textbf{Q:} What are the key components of Fidelity's iShares Core S\&P 500 ETF?\\
\textbf{A:} The fund typically invests at least 80\% of its assets in the component securities of the S\&P 500 index or in investments that have economic characteristics substantially identical to those component securities. The remaining 20\% of its assets may be invested in certain futures, options, swap contracts, cash, and cash equivalents.\\
\textbf{R:} \href{https://digital.fidelity.com/prgw/digital/research/quote/dashboard/summary?symbol=IVV}{https://digital.fidelity.com/prgw/digital/research\\/quote/dashboard/summary?symbol=IVV}

\textbf{Other Companies}\\
\noindent\textbf{T:} TMTG\\
\textbf{Q:} What is the main business of Trump Media \& Technology Group Corp.\\
\textbf{A:} TMTG's main businesses include Truth Social, a social media platform established as a safe harbor for free expression amid increasingly harsh censorship by Big Tech corporations, as well as Truth+, a TV streaming platform focusing on family-friendly live TV channels and on-demand content. TMTG is also launching Truth.Fi, a financial services and FinTech brand incorporating America First investment vehicles.\\
\textbf{R:} \href{https://s3.amazonaws.com/sec.irpass.cc/2660/0001140361-25-004822.html}{https://s3.amazonaws.com/sec.irpass.cc/2660/00\\01140361-25-004822.html}

\subsection{Policies, Laws, or Events}
\paragraph{Category Description}
What is it? What does it include? Key details and clauses to note.

\subsubsection*{Filled Examples — Policies, Laws, or Events}
\noindent\textbf{T:} Bitcoin legal\\
\textbf{Q:} What was the first country to make Bitcoin legal tender?\\
\textbf{A:} El Salvador is the first country in the world to make Bitcoin legal tender. On June 9, 2021, El Salvador's Congress passed the Bitcoin Law, making Bitcoin legal tender alongside the U.S. dollar, which went into effect on September 7 of the same year.\\
\textbf{R:} \href{https://legaljournal.princeton.edu/el-salvadors-bitcoin-law-contemporary-implications-of-forced-tender-legislation/}{https://legaljournal.princeton.edu/el-salvadors-bitcoin-law-contemporary-implications-of-forced-tender-legislation/}

\medskip
\noindent\textbf{T:} PPP Loan Forgiveness\\
\textbf{Q:} How can I qualify for the Full Forgiveness Terms of the First Draw PPP Loans based on the information provided?\\
\textbf{A:} To qualify for the Full Forgiveness Terms of the First Draw PPP Loans, the following conditions must be met: 1.\ Employee and compensation levels are maintained: During the 8-- to 24--week period following loan disbursement, the business must maintain the same number of employees and compensation levels. 2.\ Loan proceeds are spent on eligible expenses: The loan funds must be spent on payroll costs and other eligible expenses, such as rent, interest, and utilities. 3.\ At least 60\% of the proceeds are spent on payroll costs: At least 60\% of the loan amount must be used for payroll costs. By meeting these conditions, the loan may be fully forgiven.\\
\textbf{R:} \href{https://home.treasury.gov/system/files/136/Top-line-Overview-of-First-Draw-PPP.pdf}{https://home.treasury.gov/system/files/136/Top-line-Overview-of-First-Draw-PPP.pdf}

\subsection{Concepts}
\textbf{Category Description}\\
\textbf{Currency}
\begin{itemize}
  \item Exchange-rate history.
  \item Anchor/peg history and regimes.
\end{itemize}
\textbf{Other Concepts}
\begin{itemize}
  \item What (definition)
  \item Why (motivation, use cases)
  \item Features (key characteristics)
  \item Concept explanation (intuition + precise definition)
\end{itemize}

\subsubsection*{Filled Examples — Concepts}
\noindent\textbf{T:} British pound\\
\textbf{Q:} What is the anchor of the British pound?\\
\textbf{A:} The pound's anchor has changed many times throughout history and can be divided into three main stages. The first stage was the gold standard, where the value of the pound was directly linked to gold reserves, ensuring its stability due to gold's intrinsic value. The second stage was the Bretton Woods system, established after World War II, where the pound was pegged to the US dollar, and the dollar itself was pegged to gold. The third stage began with the collapse of the Bretton Woods system, and since then, the value of the pound has been based on the economic strength and creditworthiness of the UK rather than any physical commodity like gold or silver.\\
\textbf{R:} \href{https://www.britannica.com/money/money/The-decline-of-gold\#ref1089594}{https://www.britannica.com/money/money/The-decline-of-gold\#ref1089594}, \href{https://www.britannica.com/story/how-are-currency-exchange-rates-determined}{https://www.britannica.com/story/how-are-currency-exchange-rates-determined}

\medskip
\noindent\textbf{T:} interest rate\\
\textbf{Q:} What do the nominal interest rate and real interest rate refer to in the Fisher Effect?\\
\textbf{A:} In the Fisher Effect, the nominal interest rate refers to the interest rate that does not account for inflation, which is the stated rate provided by financial institutions, reflecting how the amount of deposits or loans grows over time. The real interest rate, on the other hand, is the interest rate that takes inflation into account, indicating how the purchasing power of deposits or loans changes over time. The relationship between the two can be expressed by the Fisher equation, which states that the nominal interest rate is approximately equal to the real interest rate plus the expected inflation rate.\\
\textbf{R:} \href{https://www.investopedia.com/terms/f/fishereffect.asp}{https://www.investopedia.com/terms/f/fisher\\effect.asp}

\section{Prompts}\label{app:prompts}

\subsection{Build Questions List Prompts}\label{app:build_question_list_prompt}
\label{app:questions template}
\begin{tcblisting}{
  enhanced,breakable,
  colback=white,colframe=Gray!50!black,
  colbacktitle=Gray!60!black,coltitle=white,
  title=Seed Question Expansion Prompt (2-Turn),
  listing engine=listings,
  listing only,
  listing options={
    basicstyle=\ttfamily\scriptsize,
    breaklines=true, columns=fullflexible, keepspaces=true,
    showstringspaces=false
  }
}
Task:
You will be given an original question (QO).
Please choose an random appropriate template from the 8 predefined types below, and extend the original question to two-turn questions (Q1 and Q2), where:
- Q1 is a new question that introduces a different but related aspect of the topic
- Q2 is completely identical to QO
- Questions can be relatively natural and colloquial
Q1 and Q2 must seek completely non-overlapping information - they should cover distinct aspects of the topic without any duplication.

Available Template Types:

1. Existence + Details  
* Q1: Does X have Y?  
* Q2: Specific details about Y within X.

2. Definition/Concept + Details  
* Q1: What is X or Y?  
* Q2: A follow-up question about X or Y.

3. Introductory Question + Details  
* Q1: A background question that introduces X.(not include X)  
* Q2: A detailed question about X.

4. Comparison + Focus on One  
* Q1: A comparison involving X.  
* Q2: A deeper question about X specifically.

5. Different Angles on the Same Topic  
* Q1: One common question about X.  
* Q2: Another question about X from a different perspective.

6. Cause + Details  
* Q1: Why did X happen?  
* Q2: A detailed question about X.

7. Evolution + Current State  
* Q1: How did X develop or evolve?  
* Q2: What is the current status of X?

8. Conditional Trigger + Consequence  
* Q1: What happens if X occurs?  
* Q2: In that case, how would X affect Y?

Examples:

Input:  
QO: What are the main businesses of AXA?

Action:  
Choose template type: 2. Definition/Concept + Details

Output:  
Q1: What kind of company is AXA?  
Q2: What are the main businesses of AXA?

---

Input:  
QO: How did the 2008 financial crisis affect AIG?

Action:  
Choose template type: 6. Cause + Details

Output:  
Q1: Why did the 2008 financial crisis happen?  
Q2: How did the 2008 financial crisis affect AIG?

---

Input:  
QO: How would a recession impact small businesses?

Action:  
Choose template type: 8. Conditional Trigger + Consequence

Output:  
Q1: What happens to the economy during a recession?
Q2: How would a recession impact small businesses?

---

Input:  
QO:  What is the mission of the IRS?

Action:  
Choose template type: 3. Introductory Question + Details  

Output:  
Q1: What is the U.S. federal government's tax agency?
Q2: What is the mission of the IRS?

---

Input:  
QO: What information do I need to provide to purchase auto insurance with Allstate?

Action:  
Choose template type: 1. Existence + Details

Output:  
Q1: Does Allstate have auto insurance?
Q2: What information do I need to provide to purchase auto insurance with Allstate?

---

Input:  
QO: What are the key components of Fidelity's iShares Core S&P 500 ETF?

Action:  
Choose template type: 4. Comparison + Focus on One

Output:  
Q1: Of Fidelity's ETF products, is the iShares Core S&P 500 ETF less risky or the ETP Fidelity Wise Origin Bitcoin Fund less risky.
Q2: What are the key components of Fidelity's iShares Core S&P 500 ETF?

---

Input:  
QO: What is the current status of ChatGPT technology?

Action:  
Choose template type: 7. Evolution + Current State

Output:  
Q1: How has ChatGPT evolved since its initial release? 
Q2: What is the current status of ChatGPT technology?

---

Input:  
QO: How did the 2008 financial crisis affect AIG?

Action:
Choose template type: 6. Cause + Specific Impact

Incorrect Output:
Q1: Why did the 2008 financial crisis happen?
Q2: How did the 2008 financial crisis affect AIG?

---

Now, based on the input QO, choose the most suitable template and generate corresponding Q1 and Q2 that seek completely non-overlapping information, where Q2 is completely identical to QO.

---

Input:  
QO: {{QO_PLACEHOLDER}}
\end{tcblisting}

\begin{tcblisting}{
  enhanced,breakable,
  colback=white,colframe=Gray!50!black,
  colbacktitle=Gray!60!black,coltitle=white,
  title=Seed Question Expansion Prompt (3-Turn),
  listing engine=listings,
  listing only,
  listing options={
    basicstyle=\ttfamily\scriptsize,
    breaklines=true, columns=fullflexible, keepspaces=true,
    showstringspaces=false
  }
}
Task:
You will be given an original question (QO).  
Please choose a random appropriate template from the 10 predefined types below, and extend the original question into a three-turn sequence (Q1, Q2, Q3), where:

- Q1 is a new question that introduces a different but related aspect of the topic  
- Q2 is a follow-up that brings the conversation closer to the original question  
- Q3 is completely identical to QO  
- Questions can be relatively natural and colloquial  
- Q1, Q2, and Q3 must seek completely non-overlapping information --- they should cover distinct aspects of the topic without duplication

Available Template Types:

1. Lead-in Question + Lead-in Question + Specific Detail  
* Q1: A question that leads to Y  
* Q2: A question that leads from Y to X  
* Q3: A detailed question about X

2. Evolution + Current State + Specific Detail  
* Q1: How did X develop or evolve?  
* Q2: What is the current status or key traits of X?  
* Q3: A specific question about X

3. Different Angles on the Same Topic  
* Q1: One question about X  
* Q2: Another question about X  
* Q3: A third question about X

4. Definition + Existence + Specific Detail  
* Q1: What is X?  
* Q2: Does X have Y?  
* Q3: Specific details about Y in X

5. Lead-in + Existence + Specific Detail  
* Q1: A question that introduces X  
* Q2: Does X have Y?  
* Q3: Specific details about Y in X

6. Definition + Lead-in + Specific Detail  
* Q1: A question about Y  
* Q2: A related question that introduces X  
* Q3: Specific details about X

7. Definition + Cause + Specific Detail  
* Q1: What is X?  
* Q2: Why did X happen?  
* Q3: A specific consequence or detail about X

8. Different Angles + Existence + Specific Detail  
* Q1: A general question about X  
* Q2: Does X have a specific attribute Y?  
* Q3: Detailed information about Y in X

9. Lead-in + Comparison + Specific Detail  
* Q1: A question that introduces X  
* Q2: Comparison between X and Y  
* Q3: Specific detail about X

10. Definition + Definition + Specific Detail  
* Q1: What is X?  
* Q2: What is Y?  
* Q3: A question about how Y functions in X

Examples:

Input:  
QO: What is the mission of the IRS?

Action:  
Choose template type: 1. Lead-in + Lead-in + Specific Detail

Output:  
Q1: How does the U.S. government fund its public services and programs?  
Q2: What is the U.S. federal government's tax agency?  
Q3: What is the mission of the IRS?

---

Input:  
QO: What does ChatGPT charge?

Action:  
Choose template type: 2. Evolution + Current State + Specific Detail

Output:  
Q1: How has ChatGPT evolved since its initial release?  
Q2: What is the current status of ChatGPT technology?
Q3: What does ChatGPT charge?

---

Input:  
QO: Will taking allopurinol affect my fertility?

Action:  
Choose template type: 3. Different Angles on the Same Topic

Output:  
Q1: What is allopurinol used for?  
Q2: Will taking allopurinol help me prevent gout?  
Q3: Will taking allopurinol affect my fertility?

---

Input:  
QO: What information do I need to provide to purchase auto insurance with Allstate?

Action:  
Choose template type: 4. Definition + Existence + Specific Detail

Output:  
Q1: Can you introduce me to Allstate?  
Q2: Does Allstate have auto insurance?  
Q3: What information do I need to provide to purchase auto insurance with Allstate?

---

Input:  
QO: What information do I need to provide to open a bank account at Chase?

Action:  
Choose template type: 5. Lead-in + Existence + Specific Detail

Output:  
Q1: What is the largest commercial bank in the United States?  
Q2: Does Chase offer personal bank accounts?  
Q3: What information do I need to provide to open a bank account at Chase?

---

Input:  
QO: What are the main responsibilities of the World Health Organization (WHO)?

Action:  
Choose template type: 6. Definition + Lead-in + Specific Detail

Output:  
Q1: What is the United Nations?  
Q2: What are the branches of the United Nations?  
Q3: What are the main responsibilities of the World Health Organization (WHO)?

---

Input:  
QO: How does climate change affect the insurance industry?

Action:  
Choose template type: 7. Definition + Cause + Specific Detail

Output:  
Q1: What is climate change?  
Q2: Why is climate change becoming a global concern?  
Q3: How does climate change affect the insurance industry?

---

Input:  
QO: What privacy features does WhatsApp offer?

Action:  
Choose template type: 8. Different Angles + Existence + Specific Detail

Output:  
Q1: Can I use WhatsApp for international calls?  
Q2: Does WhatsApp have built-in privacy protection?  
Q3: What privacy features does WhatsApp offer?

---

Input:  
QO: What is AWS's pricing model like?

Action:  
Choose template type: 9. Lead-in + Comparison + Specific Detail

Output:  
Q1: How do companies choose a cloud provider?  
Q2: Which is bigger, Google Cloud or AWS?  
Q3: What is AWS's pricing model like?

---

Input:  
QO: What is the role of smart contracts in blockchain platforms?

Action:  
Choose template type: 10. Definition + Definition + Specific Detail

Output:  
Q1: What is a blockchain platform?  
Q2: What is a smart contract?  
Q3: What is the role of smart contracts in blockchain platforms?

---

Input:  
QO: {{QO_PLACEHOLDER}}
\end{tcblisting}

\begin{tcblisting}{
  enhanced,breakable,
  colback=white,colframe=Gray!50!black,
  colbacktitle=Gray!60!black,coltitle=white,
  title=Detect Information Overlap Prompt,
  listing engine=listings,
  listing only,
  listing options={
    basicstyle=\ttfamily\scriptsize,
    breaklines=true, columns=fullflexible, keepspaces=true,
    showstringspaces=false
  }
}
Q1: {question_1}
A1: {answer_1}

Q2: {question_2}  
A2: {answer_2}

Do A1 and A2 contain more than 10% of the overlapping information? If yes, answer 'Yes'. If not, answer 'No'.

Purpose:
\end{tcblisting}

\subsection{Decompose Prompt}

\begin{tcblisting}{
  enhanced,breakable,
  colback=white,colframe=Gray!50!black,
  colbacktitle=Gray!60!black,coltitle=white,
  title=Decompose Prompt,
  listing engine=listings,
  listing only,
  listing options={
    basicstyle=\ttfamily\scriptsize,
    breaklines=true, columns=fullflexible, keepspaces=true,
    showstringspaces=false
  }
}
# OVERALL INSTRUCTIONS  
You are an expert in understanding logical relationships. This is a Semantic Content Unit (SCU) extraction task. Given a pair of Question and Answer, your goal is to create a list of self-contained and concise claims. Each claim should be able to stand alone and be independent of other claims. Your claims should encompass all the information present in the answer.

# TASK INSTRUCTIONS  
- List of Possible Causes: For scenarios involving multiple entities like red flags, vaccines, symptoms, etc., generate separate claims for each entity. This increases the number of claims.  
- OR Claims: When entities are presented in an "OR" context, treat them as distinct claims.  
- IF Claims: When an "if statement" is present, preserve the "if statement" context while creating the claim.  
- XOR Claims: When entities have an XOR logical relationship (e.g., treatment options), create separate claims for each option.
- Try your best to list all the information. Do not miss any information.
- Instead of summarizing the original answer, break it down.

# EXAMPLE CLAIM FORMAT  
- List Format: "Possible cause for [CONDITION] in [DEMOGRAPHIC] can be [ENTITY]."  
- OR Format: "Possible causes include: [ENTITY X], [ENTITY Y], and [ENTITY Z]."  
- OR Format: "The [CONTEXT] of treatments such as [TREATMENT X], [TREATMENT Y], and [TREATMENT Z], is not well established."  
- IF Format: "[CONTEXT], please seek medical attention if [CONDITIONS]."  
- XOR Format: "Either take [TREATMENT X] or [TREATMENT Y], but not both."

------

# TASK EXAMPLE  
Question: I am a 33-year-old female with right lower abdominal pain, what could it be?  
Answer: Possible causes for right lower abdominal pain in a young female are Appendicitis, Inflammatory bowel disease, Diverticulitis, Kidney stone, urinary tract infection, Ovarian cyst or torsion, Ectopic pregnancy, Pelvic inflammatory disease, endometriosis. Please seek medical attention if the pain is sudden and severe, does not go away, or gets worse, is accompanied by fever, nausea and vomiting, or if you have noticed blood in urine or in stool.  
Claims:  
[  
Possible cause for right lower abdominal pain in a young female: Appendicitis,  
Possible cause for right lower abdominal pain in a young female: Ovarian cyst or torsion,  
Possible cause for right lower abdominal pain in a young female: Ectopic pregnancy,  
Possible cause for right lower abdominal pain in a young female: Pelvic inflammatory disease,  
Possible cause for right lower abdominal pain in a young female: Kidney stone,  
Possible cause for right lower abdominal pain in a young female: Urinary tract infection,  
Possible cause for right lower abdominal pain in a young female: Diverticulitis,  
Possible cause for right lower abdominal pain in a young female: Inflammatory bowel disease,  
Possible cause for right lower abdominal pain in a young female: Endometriosis,  
Please seek medical attention if the pain is sudden and severe,  
Please seek medical attention if the pain is accompanied by fever,  
Please seek medical attention if the pain is accompanied by nausea and vomiting,  
Please seek medical attention if the pain is accompanied by blood in urine,  
Please seek medical attention if the pain is accompanied by blood in stool,  
Possible cause for right lower abdominal pain in a young female: Emotional stress  
]

# TASK EXAMPLE  
Question: So what does the non reactive mean for the hep a igm  
Answer: Hep A IgM refers to a specific type of antibody called Immunoglobulin M (IgM) against the virus hepatitis A. When infected with hepatitis A, these antibodies are detectable at symptom onset and remain detectable for approximately three to six months. These antibodies might also be detectable in the first month after hepatitis A vaccination. A negative or non-reactive result means no IgM antibodies against hepatitis A found in your serum, meaning the absence of an acute or recent hepatitis A virus infection.  
Claims:  
[  
A negative or non-reactive result means that there were no IgM antibodies against hepatitis A found in your serum,  
The absence of IgM antibodies against hepatitis A in your serum indicates the absence of an acute or recent hepatitis A virus infection,  
Hep A IgM refers to a specific type of antibodies called Immunoglobulin M (IgM) against the virus hepatitis A,  
These antibodies might also be detectable in the first month after hepatitis A vaccination,  
These antibodies remain detectable for approximately three to six months after infection,  
When infected with hepatitis A, these antibodies are detectable at the time of symptom onset  
]

# TASK EXAMPLE  
Question: What medications are contraindicated for a pregnant woman with ulcerative colitis?  
Answer: methotrexate (Otrexup, Rasuvo, RediTrex) and thalidomide (Contergan, Thalomid) are both considered contraindicated for treatment of UC in pregnancy. Possible treatment for UC during pregnancy include low-risk drugs such as aminosalicylates (sulfasalazine and mesalamine), immunomodulators (azathioprine, cyclosporine A, 6-mercaptopurine) and corticosteroids. Biological agents such as Infliximab, Adalimumab, Vedolizumab and Ustekinumab are generally avoided during pregnancy as their safety in pregnancy is not well established yet.  
Claims:  
[  
Methotrexate (Otrexup, Rasuvo, RediTrex) is contraindicated for treatment of ulcerative colitis in pregnancy,  
Thalidomide (Contergan, Thalomid) is contraindicated for treatment of ulcerative colitis in pregnancy,  
Aminosalicylates (sulfasalazine and mesalamine) are considered low-risk drugs for treatment of ulcerative colitis during pregnancy,  
Immunomodulators (azathioprine, cyclosporine A, 6-mercaptopurine) are considered low-risk drugs for treatment of ulcerative colitis during pregnancy,  
Corticosteroids are considered low-risk drugs for treatment of ulcerative colitis during pregnancy,  
Treatment for ulcerative colitis during pregnancy with biological agents such as Adalimumab is generally avoided during pregnancy as their safety in pregnancy is not well established yet,  
Treatment for ulcerative colitis during pregnancy with biological agents such as Vedolizumab is generally avoided during pregnancy as their safety in pregnancy is not well established yet,  
Treatment for ulcerative colitis during pregnancy with biological agents such as Infliximab is generally avoided during pregnancy as their safety in pregnancy is not well established yet,  
Treatment for ulcerative colitis during pregnancy with biological agents such as Ustekinumab is generally avoided during pregnancy as their safety in pregnancy is not well established yet  
]

Question: What is the most safe amount of Advil to take at one time 
Answer: The safe amount of Advil (ibuprofen) to take at one time depends on various factors, including your age, weight, and overall health condition. Here are general guidelines for adults:\n\n- **Standard Dosage**: The typical adult dose for pain relief is 200 mg to 400 mg every 4 to 6 hours as needed.\n- **Maximum Daily Dose**: The maximum daily dose for adults is usually 1200 mg to 3200 mg, depending on the formulation and the manufacturer's recommendations. However, it's important to stay within the lower end of this range unless specifically advised otherwise by a healthcare provider.\n\nFor example:\n- For mild to moderate pain, you might start with 200 mg every 4 to 6 hours.\n- For more severe pain, you might take 400 mg every 4 to 6 hours.\n\nIt's crucial to follow the specific instructions on the package or those provided by your healthcare provider. Overuse can lead to serious side effects, including stomach ulcers, bleeding, and kidney problems.\n\nIf you have any underlying health conditions or are taking other medications, it's best to consult with a healthcare provider to determine the safest dosage for you. Always read the label and follow the recommended dosages carefully.
Claims:  
[
The safe amount of Advil (ibuprofen) to take at one time depends on various factors, including your age, weight, and overall health condition.
The safe amount of Advil (ibuprofen) to take at one time for an adult typically ranges from 200 mg to 400 mg,
Adults should take the standard dosage of Advil every 4 to 6 hours as needed for pain relief,
The maximum daily dose of Advil for adults usually ranges from 1200 mg to 3200 mg, depending on the formulation and manufacturer's recommendations,
It's important to stay within the lower end of ranges from 1200 mg to 3200 mg unless specifically advised otherwise by a healthcare provider,
For mild to moderate pain, the initial dosage of Advil for adults is 200 mg every 4 to 6 hours,
For more severe pain, the dosage of Advil for adults can be increased to 400 mg every 4 to 6 hours,
Overuse of Advil can lead to serious side effects such as stomach ulcers, bleeding, and kidney problems,
It is crucial to follow the specific instructions on the package or those provided by a healthcare provider when taking Advil,
Consulting a healthcare provider is advisable if you have underlying health conditions or are taking other medications before determining the safest dosage of Advil,
Always read the label and follow the recommended dosages carefully when taking Advil.
]

# YOUR TASK
Question: {question}  
Answer: {answer}  
Claims:
\end{tcblisting}

\subsection{Detect Contradict Prompt}
\begin{tcblisting}{
  enhanced,breakable,
  colback=white,colframe=Gray!50!black,
  colbacktitle=Gray!60!black,coltitle=white,
  title=Detect Contradict Prompt,
  listing engine=listings,
  listing only,
  listing options={
    basicstyle=\ttfamily\scriptsize,
    breaklines=true, columns=fullflexible, keepspaces=true,
    showstringspaces=false
  }
}
# OVERALL INSTRUCTIONS
- You have a deep understanding of logical relationships, such as entailment and contradiction, to evaluate given triplets of (question, premise, hypothesis).

# TASK INSTRUCTIONS
Your goal is to determine whether the Premise effectively contradicts the corresponding Hypothesis. Carefully analyze each triplet, focusing on details, not introducing knowledge.
- If the premise and the hypothesis are unrelated or lack sufficient evidence to ascertain their truthfulness, label your answer as False.
- be vigilant in identifying cases where the premise doesn't rule out the possibility of an entity (e.g., vaccine, symptom) appearing in the hypothesis. In such cases, classify the answer as False. 
- If the answer is true, the answer should include "The answer is True". If the answer is false, the answer should include "The answer is False".
- Approach each question methodically, considering the step-by-step process outlined below.

# INPUT DATA
Question: What does trich test for? Let's think step by step.
Premise: The term "trich test" can refer to two different medical tests, depending on the context. Here are the two possibilities:
Trichomoniasis Test: Trichomoniasis is a sexually transmitted infection (STI) caused by the parasite Trichomonas vaginalis.
The trichomoniasis test, also known as a trich test or trichomonas test, is used to detect the presence of this parasite in the body. The test is typically performed on a sample of vaginal discharge in women or urine in men.
Trichogram: A trichogram is a diagnostic test used to evaluate hair loss and assess the health and condition of hair follicles. It involves plucking a small number of hairs from the scalp and examining them under a microscope.
It's important to note that without additional context, it's difficult to determine which specific test you are referring to.
Hypothesis: Trichamoniasis- a parasitic infection that can cause your symptoms.
Answer: According to the premise "trich test" refer to two different medical tests. A Trichamoniasis test is one of them, which is used to detect this parasite's presence.
As stated in the hypothesis, the trich test is used to diagnose parasitic infections. Ths premise entails the hypothesis. The answer is False.
###
Question: Can diabetics eat sweets? Let's think step by step.
Premise: Individuals with diabetes are recommended to limit their consumption of sweets to one or two times per week.
It is also suggested being selective with desserts and to focus on foods with a low glycemic index, such as high fiber foods like whole grains and legumes, as well as certain lower sugar fruits like berries, melons, and apples.
Hypothesis: It is recommended that diabetics avoid sweets.
Answer: The premise suggests that diabetics can eat sweets but limit their consumption. According to the hypothesis diabetics should avoid sweets.
Diabetics are allowed to consume sweets according to the premise, but they are prohibited according to the hypothesis. There is a contradiction between the premise and the hypothesis. The answer is True.
###
Question: 25 yo female with right lower abdominal pain, what might be causing it? Let's think step by step.
Premise: Right lower abdominal pain in a 25-year-old female could be caused by a variety of medical conditions.
Some potential causes include: Ovarian cyst: a fluid-filled sac on the ovary - Ectopic pregnancy: a pregnancy that occurs outside the uterus.
Hypothesis: possible cause for right lower abdominal pain in a young female can be Appendicitis.
Answer: The premise lists several potential causes of right lower abdominal pain in a 25-year-old female, not including appendicitis. The hypothesis states that Appendicitis could be a cause of right lower abdominal pain in a young female.
There is no direct contradiction between the premise and the hypothesis, as the premise does not exclude the possibility of appendicitis as the cause of the pain. The answer is False.
###
Question: Can a headache last longer than a few days? Let's think step by step.
Premise: Yes, it is possible. If you are experiencing a headache that lasts longer than a few days, it is important to see a doctor to get the appropriate treatment.
This will help to relieve the pain and prevent any further complications.
Hypothesis: It is not a cause for concern if a headache lasts longer than a few days.
Answer: This premise acknowledges that a headache can last for several days, but emphasizes that seeing a doctor to prevent further complications is important. According to this hypothesis, headaches lasting longer than a few days are not cause of concern.
There is a contradiction between the premise and hypothesis due to the premise emphasizing the importance of seeking medical consultation, while the hypothesis posits that there is no cause for concern. The answer is True.
###
Question: {question} Let's think step by step.
Premise: {llm_answer}
Hypothesis: {answer}
Answer:
\end{tcblisting}

\subsection{Detect Entail Prompt}
\begin{tcblisting}{
  enhanced,breakable,
  colback=white,colframe=Gray!50!black,
  colbacktitle=Gray!60!black,coltitle=white,
  title=Detect Entail Prompt,
  listing engine=listings,
  listing only,
  listing options={
    basicstyle=\ttfamily\scriptsize,
    breaklines=true, columns=fullflexible, keepspaces=true,
    showstringspaces=false
  }
}
# OVERALL INSTRUCTIONS
- You have a deep understanding of logical relationships, such as entailment and contradiction, to evaluate given triplets of (question, premise, hypothesis).

# TASK INSTRUCTIONS
Your goal is to determine whether the Premise effectively entails the corresponding Hypothesis. Carefully analyze each triplet, focusing on details, not introducing knowledge.
- If the premise disagrees with, is unrelated to, or does not support the hypothesis, there is not enough evidence to determine whether it is true, and so you answer should be False.
- If the answer is true, the answer should include "The answer is True". If the answer is false, the answer should include "The answer is False".
- Approach each question methodically, considering the step-by-step process outlined below.

# INPUT DATA
Question: What does trich test for? Let's think step by step.
Premise: The term "trich test" can refer to two different medical tests, depending on the context. Here are the two possibilities:
Trichomoniasis Test: Trichomoniasis is a sexually transmitted infection (STI) caused by the parasite Trichomonas vaginalis.
The trichomoniasis test, also known as a trich test or trichomonas test, is used to detect the presence of this parasite in the body. The test is typically performed on a sample of vaginal discharge in women or urine in men.
Trichogram: A trichogram is a diagnostic test used to evaluate hair loss and assess the health and condition of hair follicles. It involves plucking a small number of hairs from the scalp and examining them under a microscope.
It's important to note that without additional context, it's difficult to determine which specific test you are referring to.
Hypothesis: Trichamoniasis- a parasitic infection that can cause your symptoms.
Answer: According to the premise "trich test" refer to two different medical tests. A Trichamoniasis test is one of them, which is used to detect this parasite's presence.
As the hypothesis suggested, the trich test is used to diagnose parasitic infections. The premise entails the hypothesis. The answer is True.
###
Question: Can diabetics eat sweets? Let's think step by step.
Premise: Individuals with diabetes are recommended to limit their consumption of sweets to one or two times per week.
It is also suggested to be selective with desserts and to focus on foods with a low glycemic index, such as high fiber foods like whole grains and legumes, as well as certain lower sugar fruits like berries, melons, and apples.
Hypothesis: After eating sweets, must monitor blood and sugar level
Answer: The premise suggests that diabetics can eat sweets but limit their consumption. The hypothesis highlights the necessity of monitor blood and sugar after eating sweets.
There is no relationship between the premise and hypothesis, therefore they do not entail one another. The answer is False.
###
Question: Can diabetics eat sweets? Let's think step by step.
Premise: Individuals with diabetes are recommended to limit their consumption of sweets to one or two times per week.
It is also suggested being selective with desserts and to focus on foods with a low glycemic index, such as high fiber foods like whole grains and legumes, as well as certain lower sugar fruits like berries, melons, and apples.
Hypothesis: It is recommended that diabetics avoid sweets.
Answer: The premise suggests that diabetics can eat sweets but limit their consumption. According to the hypothesis diabetics should avoid sweets.
The premise allows diabetics to consume sweets in moderate consumption, while the hypothesis prohibits them. There premise don't entail the hypothesis. The answer is False.
###
Question: What is the best hypertension treatment for patients who are also have Crohn's disease? Let's think step by step.
Premise: For patients with Crohn's disease and hypertension, the recommended treatment is a combination of lifestyle changes and medication. The ACC/AHA recommends initiation of antihypertensive drug therapy at a BP \u2265130/80 mm Hg for adults with hypertension.
It is also important to monitor your blood pressure regularly to make sure that it is under control.
Hypothesis: reducing sodium intake, are the first-line treatment for hypertension in individuals with  Crohn's disease
Answer: The premise suggests that the recommended treatment for patients with diabetes and hypertension is a combination of lifestyle changes and medication, including antihypertensive drug therapy. The hypothesis focuses on reducing sodium intake as the first-line treatment.
A reduction in sodium intake could be a part of the lifestyle changes, but since it is not mentioned in the premise, the premise do not entail the hypothesis. The answer is False.
###
Question: 25 yo female with right lower abdominal pain, what might be causing it? Let's think step by step.
Premise: Right lower abdominal pain in a 25-year-old female could be caused by a variety of medical conditions.
Some potential causes include: - Appendicitis: inflammation of the appendix - Appendiceal abscess: a collection of pus in the appendix - Ovarian cyst: a fluid-filled sac on the ovary - Ectopic pregnancy: a pregnancy that occurs outside the uterus.
Hypothesis: possible cause for right lower abdominal pain in a young female can be Appendicitis.
Answer: The premise lists several potential causes of right lower abdominal pain in a 25-year-old female, including appendicitis. The hypothesis states that Appendicitis could be a cause of right lower abdominal pain in a 25-year-old female.
Both the premise and hypothesis mention appendicitis as a possible cause of pain, so the premise entails the hypothesis. The answer is True.
###
Question: {question} Let's think step by step.
Premise: {llm_answer}
Hypothesis: {answer}
Answer:
\end{tcblisting}

\section{Case Study}
\subsection{Case Study 1: Enhancement of Fact Accuracy Through Retrieval}
\begin{tcblisting}{
  enhanced,breakable,
  colback=white,colframe=Gray!50!black,
  colbacktitle=Gray!60!black,coltitle=white,
  title=Case Study 1: Enhancement of Fact Accuracy Through Retrieval,
  listing engine=listings,
  listing only,
  listing options={
    basicstyle=\ttfamily\scriptsize,
    breaklines=true, columns=fullflexible, keepspaces=true,
    showstringspaces=false
  }
}
Q: How long would I be on Saxenda before I could move to the next dose?

GT: ''It is recommended that adult patients start with a dose of 0.6 mg per day for one week. The dose is then increased by 0.6 mg each week over the following 4 weeks until the daily maintenance dose of 3 mg is reached. If a patient does not tolerate an increased dose during the dosing schedule, dose escalation can be delayed by one extra week.''

base_answer: ''The dosing schedule for Saxenda (liraglutide) is designed to gradually increase over time to achieve the desired therapeutic effect. Here's a general outline of how the dosing progression works:

1. Initial Dose: Saxenda is typically started at a low dose, usually 0.6 mg once daily via subcutaneous injection. This initial dose is often given for 2 weeks to allow the body to adjust to the medication.

2. First Increase: After the first 2 weeks, the dose is increased to 1.2 mg once daily. This step is also typically maintained for 2 weeks to ensure the body continues to adapt.

3. Second Increase: After another 2 weeks at 1.2 mg, the dose is increased to 1.8 mg once daily. This final dose is then maintained for the duration of treatment unless otherwise directed by a healthcare provider.

The exact timing and duration can vary based on individual response and medical supervision. Some patients might stay on a particular dose longer if they are experiencing side effects or if their healthcare provider decides it is necessary.

It's important to follow the specific instructions provided by your healthcare provider, as they will monitor your progress and adjust the dosing schedule as needed. Regular follow-up appointments are crucial to assess your response to the medication and make any necessary adjustments.

If you have any questions about your dosing schedule or if you experience any side effects, it's important to contact your healthcare provider for guidance.''

RAG_answer: ''For Saxenda, you would be on each dose for one week before moving to the next higher dose. Specifically, the dose escalation schedule is as follows:

- Week 1: 0.6 mg daily
- Week 2: 1.2 mg daily
- Week 3: 1.8 mg daily
- Week 4: 2.4 mg daily
- Week 5 and onward: 3 mg daily

After completing one week at each dose, you would then increase to the next higher dose. If you or a pediatric patient cannot tolerate the increased dose during the dose escalation process, you may delay the escalation for approximately one additional week.''

RAG_retrieved_chunks: "The following information has been retrieved and may be helpful:

1.
\"\"\"
*
Use SAXENDA with a reduced-calorie diet and increased physical activity.
*
SAXENDA is injected under the skin (subcutaneously) in your stomach area (abdomen), upper leg (thigh), or upper arm.
Do not
inject into a muscle (intramuscularly) or vein (intravenously).
*
SAXENDA is injected 1 time each day, at any time during the day.
*
Start SAXENDA with 0.6 mg per day in your first week. In your second week, increase your daily dose to 1.2 mg. In the third week, increase your daily dose to 1.8 mg. In the fourth week, increase your daily dose to 2.4 mg and in the fifth week onwards, increase your daily dose to the full dose of 3 mg. After that, do not change your dose unless your healthcare provider tells you to. Children may reduce their dose to 2.4 mg daily if the maximum dose is not tolerated.
*
If you miss your daily dose of SAXENDA, just take your next daily dose as usual on the following day. Do not take an extra dose of SAXENDA or increase your dose on the following day to make up for your missed dose. If you miss your dose of SAXENDA for
3 days or more
, call your healthcare provider to talk about how to restart your treatment.
*
SAXENDA may be taken with or without food.
*
Change (rotate) your injection site with each injection.
Do not
use the same site for each injection.
*
Do not share your SAXENDA pen with other people, even if the needle has been changed.
You may give other people a serious infection or get a serious infection from them.
*
If you take too much SAXENDA, call your healthcare provider or Poison Help line at 1-800-222-1222 or go to the nearest hospital emergency room right away.
*
Throw away the used SAXENDA pen after 30 days.
What are the possible side effects of SAXENDA?
SAXENDA may cause serious side effects, including:
*
See "What is the most important information I should know about SAXENDA?"
*
inflammation of the pancreas
(
pancreatitis).
\"\"\"
2.
\"\"\"
Table 1. Dose Escalation Schedule
Week
Daily Dose
1
0.6 mg
2
1.2 mg
3
1.8 mg
4
2.4 mg
5 and onward
3 mg
Adult Patients
*
For adults, the recommended dosage of SAXENDA is 3 mg daily, lower dosages are for titration only.
*
Discontinue SAXENDA if the patient cannot tolerate the 3 mg dosage.
*
If patients do not tolerate an increased dose during dosage escalation, consider delaying dosage escalation for approximately one additional week.
*
Evaluate the change in body weight 16 weeks after initiating SAXENDA and discontinue SAXENDA if the patient has not lost at least 4% of baseline body weight, since it is unlikely that the patient will achieve and sustain clinically meaningful weight loss with continued treatment.
*
In adult patients with type 2 diabetes, monitor blood glucose prior to starting SAXENDA and during SAXENDA treatment.
Pediatric Patients
*
For pediatric patients, the recommended maintenance dosage of SAXENDA is 3 mg daily. Pediatric patients who do not tolerate 3 mg daily may have their maintenance dose reduced to 2.4 mg daily. Discontinue SAXENDA if the patient cannot tolerate the 2.4 mg dose.
*
If pediatric patients do not tolerate an increased dose during dosage escalation, the dose may also be lowered to the previous level. Dosage escalation for pediatric patients may take up to 8 weeks.
*
Evaluate the change in BMI after 12 weeks on the maintenance dose and discontinue SAXENDA if the patient has not had a reduction in BMI of at least 1% from baseline, since it is unlikely that the patient will achieve and sustain clinically meaningful weight loss with continued treatment.
3 DOSAGE FORMS AND STRENGTHS
Injection: 6 mg/mL clear, colorless solution in a 3 mL prefilled, single-patient-use pen that delivers doses of 0.6 mg, 1.2 mg, 1.8 mg, 2.4 mg, or 3 mg.
\"\"\"
3.
\"\"\"
Home
/
About Saxenda
(R)
/ Dosing Schedule
Saxenda
(R)
Dosing Schedule
Get help staying on track with Saxenda
(R)
by connecting with your prescribing health care provider, reviewing the daily dosing schedule below, or calling the Saxenda
(R)
Hotline at 1-844-845-6913 Monday through Friday 8:30
AM
to 6:00
PM
ET.
Need reminders for when to take Saxenda
(R)
? Try setting an alarm on a smartphone or AI assistant (like Alexa or Siri).
Review the Saxenda
(R)
dosing schedule
The Saxenda
(R)
daily dosing schedule was designed to help patients get used to their new medicine while also minimizing gastrointestinal side effects. Review the schedule below, and be sure to schedule a 4-month follow-up with your prescribing health care provider. This visit can be sooner if there are questions or concerns about dosing or side effects. As a reminder, a Saxenda
(R)
prescription contains enough medicine for 30 days.
Below is a dosing schedule to guide you through the starting dosage of 0.6 mg to the dosage of 3 mg.
Saxenda
(R)
should be taken exactly as prescribed. After starting Saxenda
(R)
, provided there are no issues with tolerating it, the dose should be increased weekly until the 3 mg dose is reached. If you or your teen have trouble tolerating an increased dose, or if you have any other questions, be sure to call your (or your teen's) health care provider. The Saxenda
(R)
dose should only be changed if your (or your teen's) health care provider advises to do so.
Nausea is the most common side effect when first starting Saxenda
(R)
. Learn more about ways to manage it
here
. For additional side effects, please refer to the
Medication Guide
.
If you (or your teen) miss your daily dose of Saxenda
(R)
, just take the next daily dose as usual on the following day. Do not take an extra dose of Saxenda
(R)
or increase the dose on the following day to make up for the missed dose. If Saxenda
(R)
is not taken for 3 days or more, call your respective health care provider to talk about how to restart treatment.
\"\"\"
4.
\"\"\"
How long each pen lasts depends on the dose you take. There is a scale on the pen that shows you about how much Saxenda is left inside.
What is the average weight loss with Saxenda?
In clinical studies, a greater number of people who took Saxenda for 56 weeks lost 5 to 10% of their body weight, compared with people who took a placebo. (A placebo is a substance that has no effect and is used as a control when testing medications). This means participants lost 12 to 23 pounds. People in the studies were overweight (BMI 27-29.9 kg/m2) or obese (BMI greater than or equal to 30 kg/m2) before taking Saxenda.
Can you drink alcohol with Saxenda?
There is no specific warning about drinking alcohol with Saxenda. Keep in mind that alcohol can lower blood sugar. If you are taking other drugs for diabetes, drinking alcohol increases your risk of hypoglycemia. In addition, many alcoholic drinks contain high levels of carbohydrates and sugar. If you're trying to lose weight, you may want to avoid alcohol.
How long does it take for Saxenda to work?
Saxenda reaches its maximum concentration in the body 11 hours after injection. It's recommended to follow up with your doctor 2 to 8 weeks after starting Saxenda to see if it's working.
If you have not lost 4% of your body weight after 16 weeks, your doctor may tell you to stop taking it.
In children ages 12 and up, Saxenda may be stopped after 12 weeks on the maintenance dose if BMI has not decreased by 1%.
Why am I not losing weight on Saxenda?
It takes time to lose weight with Saxenda. It may take about 8 weeks before you start to see significant weight loss (about 5%) with Saxenda, but in the first 2 to 4 weeks you may lose about 2% to 4% of your weight. You should also adhere to a long-term reduced-calorie diet and exercise program as prescribed by your doctor for maximum weight loss.
Continue reading
Liraglutide vs Semaglutide: How do they compare?
\"\"\"
5.
\"\"\"
Refer to the accompanying Instructions for Use for complete administration instructions with illustrations.
*
Inspect SAXENDA visually prior to each injection. Only use if solution is clear, colorless, and contains no particles.
*
Inject SAXENDA subcutaneously once daily at any time of day, without regard to the timing of meals.
*
Inject SAXENDA subcutaneously in the abdomen, thigh, or upper arm. No dose adjustment is needed if changing the injection site and/or timing.
*

Rotate injection sites within the same region in order to reduce the risk of cutaneous amyloidosis.
*
If a dose is missed, resume the once-daily regimen as prescribed with the next scheduled dose. Do not administer an extra dose or increase the dose to make up for the missed dose.
*
If more than 3 days have elapsed since the last SAXENDA dose, reinitiate SAXENDA at 0.6 mg daily and follow the dose escalation schedule in Table 3, to reduce the occurrence of gastrointestinal adverse reactions associated with reinitiation of treatment.
Dosage in Adults and Pediatric Patients Aged 12 Years and Older
*
Initiate SAXENDA with a dose of 0.6 mg daily for one week. Then follow the dose escalation schedule in Table 3 to minimize gastrointestinal adverse reactions.
Table 3: Dose Escalation Schedule
Week
Daily Dose
1
0.6 mg
2
1.2 mg
3
1.8 mg
4
2.4 mg
5 and onward
3 mg
Adult Patients
*
For adults, the recommended dosage of SAXENDA is 3 mg daily, lower doses are for titration only.
*
Discontinue SAXENDA if the patient cannot tolerate the 3 mg dose.
*
If patients do not tolerate an increased dose during dose escalation, consider delaying dose escalation for approximately one additional week.
*
Evaluate the change in body weight 16 weeks after initiating SAXENDA and discontinue SAXENDA if the patient has not lost at least 4% of baseline body weight, since it is unlikely that the patient will achieve and sustain clinically meaningful weight loss with continued treatment.
*
In adult patients with type 2 diabetes, monitor blood glucose prior to starting SAXENDA and during SAXENDA treatment.
Pediatric Patients
*
For pediatric patients, the recommended maintenance dosage of SAXENDA is 3 mg daily.
\"\"\"

Please answer the following question based only on the information above:
How long would I be on Saxenda before I could move to the next dose?"

\end{tcblisting}

The question concerns how long a patient typically stays on each dose of Saxenda (liraglutide) before increasing to the next one. The gold-standard text clearly provides the recommended escalation schedule: adults start at 0.6 mg per day for one week, and the dose is then increased by 0.6 mg each week over the next four weeks until reaching the 3 mg daily maintenance dose. If a patient has difficulty tolerating a specific dose, the escalation can be delayed by an extra week.

The base model's answer performs poorly in terms of factual accuracy because it does not reproduce this stepwise weekly schedule. Instead, it offers a generic description of how GLP-1 agonists are usually titrated, using vague statements such as "starting at a low dose" and "often given for 2 weeks," without providing the precise weekly increments or the fixed five-week escalation plan. Its statements focus on a general idea of gradual dose adjustment under medical supervision rather than the specific information contained in the official product guidelines, which explains why its factual score is essentially zero.

In contrast, the RAG-enhanced answer closely matches the gold-standard dosing schedule. It explicitly lays out the weekly plan---0.6 mg in week one, followed by 1.2 mg, 1.8 mg, and 2.4 mg, and finally reaching 3 mg by week five. It also includes the key detail that dose escalation may be delayed by approximately one week if the patient does not tolerate the increase. This alignment with the official instructions results in a high factual score. The retrieval system appears to have pulled information directly from authoritative sources such as Drugs.com, enabling the model to provide a precise, accurate dose-escalation timeline rather than relying on general clinical intuition.

This case illustrates how retrieval greatly improves performance for topics where the correct answer depends on exact, non-negotiable details---such as fixed titration schedules for prescription medications. Without retrieval, the model tends to generalize from patterns it has learned, but with retrieval, it can reproduce the precise regimen specified in official guidelines.

\subsection{Case Study 2: Performance Degradation Resulting from Misaligned Retrieval Focus}
\begin{tcblisting}{
  enhanced,breakable,
  colback=white,colframe=Gray!50!black,
  colbacktitle=Gray!60!black,coltitle=white,
  title=Case Study 2: Performance Degradation Resulting from Misaligned Retrieval Focus,
  listing engine=listings,
  listing only,
  listing options={
    basicstyle=\ttfamily\scriptsize,
    breaklines=true, columns=fullflexible, keepspaces=true,
    showstringspaces=false
  }
}

Q: What are the two most prevalent short-term debt instruments in the money market?

GT: "Most Common Short-Term Bonds in Money Markets: T-Bills and Commercial Paper."

base_answer: "The two most prevalent short-term debt instruments in the money market are typically:

1. Treasury Bills (T-Bills): These are short-term debt obligations issued by the government with maturities ranging from a few days to one year. They are considered very safe due to the backing of the government and are widely used for short-term investments and liquidity management.

2. Commercial Paper: This is an unsecured promissory note issued by corporations with maturities ranging from a few days to 270 days. Commercial paper is used for short-term financing and is often issued by large, creditworthy companies to raise funds quickly.

These two instruments are highly liquid and are commonly traded in the money market, making them essential components of short-term investment strategies."

RAG_answer: "Based on the information provided, the two most prevalent short-term debt instruments in the money market are:

1. Treasury Bills (T-bills) - Short-term debt obligations backed by the U.S. government, sold in denominations of $1,000 up to a maximum of $5 million, with maturities ranging from four to 52 weeks.
2. Commercial Paper - Unsecured, short-term debt issued by large corporations to fund day-to-day operations, typically maturing within 270 days and issued at a discount to face value.

These instruments are widely recognized for their liquidity, low risk, and role in stabilizing interest rates and managing short-term funding needs."

RAG_retrieved_chunks: "The following information has been retrieved and may be helpful:

1.
\"\"\"
Treasury notes and bonds are not included here. Bonds and other fixed-income instruments with longer terms are considered part of the capital markets.
Repurchase Agreements (Repos)
:
These assets involve the sale of securities with an agreement to repurchase them at a slightly higher price on a specific future date---often the next day. They're essentially short-term loans, typically used by dealers in government securities. The securities serve as collateral, making repos relatively low-risk.
Commercial Paper:
Commercial paper consists of unsecured, short-term debt issued by large corporations to fund day-to-day operations. These instruments typically mature within 270 days and are issued at a discount to face value.
While riskier than T-bills, commercial paper from top-rated companies often offers slightly higher yields.
CDs:
CDs have fixed terms ranging from a few weeks to several years and pay higher interest rates than standard savings accounts, though the depositor has to wait a period to obtain the funds back. Keep in mind that CDs that mature after one year aren't part of the money market.
Prime funds typically invest in short-term obligations issued by banks and corporations. Two of the most common are listed below.
Bankers' Acceptances:
These are short-term debt instruments guaranteed by a bank, often used for international trade. When a bank accepts a
bankers' acceptance
, it assumes responsibility for paying the holder when the instrument matures. This bank guarantee makes them relatively safe investments.
Short-Term Corporate Bonds:
For inclusion in the money market, these corporate bonds have maturities of one year or less. This is debt issued by companies that offer a way for corporations to borrow money from investors for relatively brief periods, often to fund operations, finance projects, or refinance existing debt.
Fast Fact
The highest quality (and safest, lower yielding) bonds are commonly referred to as Triple-A bonds, while the least creditworthy are termed junk.
Tax-exempt funds are composed of short-term government-issued debt obligations. The interest income earned on these investments is exempt from federal taxes. Some may also be exempt from local and state taxes.
Municipal Bonds (Munis):
Tax-exempt money market funds are primarily municipal bonds or notes, which are issued by state and local governments. These are often tax-exempt at the federal level, making them attractive to investors in high tax brackets. The chart below shows the $6.57 trillion U.S.
\"\"\"
2.
\"\"\"
They work through instruments like
commercial paper
, Treasury bills (T-bills), and
certificates of deposit (CDs)
. These instruments facilitate quick fund transfers and help to stabilize interest rates. They are often regarded as a haven for investors to park their surplus cash and keep the system liquid and stable.
In the money markets, banks, corporations, and government entities buy and sell financial instruments to manage liquidity. These transactions involve instruments like
T-bills
and commercial paper where terms are shorter and settlement is quick. This fast-paced activity helps participants to manage their short-term funding needs.
Money market operations are crucial for the level of liquidity and interest rates in the economy. They provide quick access to cash and stabilize interest rates so they are more predictable. The quick turnaround of funds allows investors to park their money temporarily and supports monetary stability.
Types and Examples of Money Markets
Money markets play a crucial role in the financial system, providing a place for institutions and individuals to park cash safely for short periods. These markets deal in highly liquid, short-term debt instruments, typically with maturities of one year or less. Let's explore the main types of money market instruments and how they function.
Government money market funds primarily invest in short-term securities issued by the U.S. government, such as T-bills and other government-backed instruments. They are considered very safe and liquid, offering a lower yield but greater security compared with prime funds.
T-Bills:
These are short-term debt obligations backed by the U.S. government and sold in denominations of $1,000 up to a maximum of $5 million. T-bills mature in four, eight, 13, 26, or 52 weeks.
Investors buy them at a discount and receive the full face value when they mature with the difference representing the interest earned. Treasury notes and bonds are not included here. Bonds and other fixed-income instruments with longer terms are considered part of the capital markets.
Repurchase Agreements (Repos)
:
These assets involve the sale of securities with an agreement to repurchase them at a slightly higher price on a specific future date---often the next day. They're essentially short-term loans, typically used by dealers in government securities. The securities serve as collateral, making repos relatively low-risk.
Commercial Paper:
Commercial paper consists of unsecured, short-term debt issued by large corporations to fund day-to-day operations.
\"\"\"
3.
\"\"\"
Money Markets vs. Capital Markets: What's the Difference?
Table of Contents
Expand
Table of Contents
An Overview
Money Markets
Capital Markets
Key Differences
Alternatives
Regulation & Oversight
FAQs
The Bottom Line
Money Markets vs. Capital Markets: An Overview
Money and capital markets are fundamental to the economy, serving investors and businesses alike.
Money markets
deal in short-term debt instruments, usually for one year or less. It's where governments, banks, and large corporations go to manage their immediate cash needs.
Capital markets
involve long-term securities, such as stocks and bonds, that mature in more than one year. This is where companies and governments raise funds for major projects and long-term growth.
Money markets are the lifeblood of day-to-day financial operations, while capital markets sustain long-term
economic growth
. They differ in three ways: the types of financial instruments traded, the duration of investments, and the level of risk. While the money market prioritizes liquidity and safety, the capital market offers the potential for higher returns with increased risk. Below, we'll explore each market's characteristics and how they work.
Key Takeaways
Money markets involve short-term lending that borrowers can tap into for cash for day-to-day operations
Capital markets are geared toward long-term investing.
Money markets are less risky than capital markets, which can be more rewarding.
Both markets are subject to comprehensive regulation to ensure transparency, fairness, and stability.
ANGELA WEISS
/ Contributor / Getty Images
Money Markets
Money markets are meant for short-term lending and borrowing, usually for a year or less. It's like a fast lane where businesses, governments, and
financial institutions
can meet their quick funding needs. Thus, it is important for
liquidity
management. These markets are known for their high liquidity, generally low risk, and ease of access to capital.
They work through instruments like
commercial paper
, Treasury bills (T-bills), and
certificates of deposit (CDs)
. These instruments facilitate quick fund transfers and help to stabilize interest rates. They are often regarded as a haven for investors to park their surplus cash and keep the system liquid and stable.
In the money markets, banks, corporations, and government entities buy and sell financial instruments to manage liquidity. These transactions involve instruments like
T-bills
and commercial paper where terms are shorter and settlement is quick.
\"\"\"
4.
\"\"\"
Tax-exempt funds are composed of short-term government-issued debt obligations. The interest income earned on these investments is exempt from federal taxes. Some may also be exempt from local and state taxes.
Municipal Bonds (Munis):
Tax-exempt money market funds are primarily municipal bonds or notes, which are issued by state and local governments. These are often tax-exempt at the federal level, making them attractive to investors in high tax brackets. The chart below shows the $6.57 trillion U.S. money market broken down under the main headings used here:
Tip
The money market is far broader than money market funds or accounts available at banks and other financial institutions. While related, the latter is a mutual fund that invests in high-quality, short-term debt instruments and cash equivalents. Many are also insured by the
Federal Deposit Insurance Corporation (FDIC)
.
Capital Markets
Capital markets play a vital role in economic growth by channeling savings into productive investments. They are where longer-term securities are bought and sold. Companies and governments raise funds by issuing stocks (equities) and bonds (fixed-income securities). Investors can earn returns via value
appreciation
or distributions.
Transactions enable individuals and institutions to tap into future opportunities. Investors buy long-term instruments like
stocks and bonds
from issuers in primary markets or trade them in secondary markets. This helps companies and governments get the funds they need for various projects and objectives. Investors hope to get returns through
dividends
or interest and potential appreciation.
Buyers and sellers are matched through exchanges or
over-the-counter (OTC)
platforms. Brokers and dealers play a key role in facilitating transactions and keeping things smooth. Pricing in the capital markets is driven by supply and demand, investor sentiment, and economic indicators.
Facilitating the trade of financial assets helps set asset prices and ensures a certain degree of liquidity, allowing funds to move smoothly. Consequently, this market underpins business
expansion
and bolsters the overall economy.
Types and Examples of Capital Markets
Capital markets can be broken down into the primary and secondary markets. The primary market is where new securities are sold for the first time, such as when a company goes public with an
initial public offering (IPO)
. This allows companies to raise capital directly from investors who buy these new shares.
\"\"\"
5.
\"\"\"
If you want short-term, low-risk investments with quick returns, the money market is probably the way to go. Instruments like Treasury bills help you preserve capital and provide liquidity over shorter periods.
Most investors have a long-term time horizon and turn to capital markets. Investing in stocks and/or bonds can build wealth and align with long-term financial goals while riding out market fluctuations.
How Do Geopolitical Events Affect Money Markets?
Geopolitical events increase volatility and risk and cause a flight to safety in money markets as investors seek safe havens.
What Role Do Central Banks Have in the Money Markets?
Central banks
influence money markets by setting interest rates and conducting open market operations to manage liquidity. The U.S. Federal Reserve serves in this role in the U.S.
Why Are the Capital Markets Important for Startups?
Capital markets provide startups with access to funding through IPOs and venture capital, fueling their growth.
The Bottom Line
Capital and money markets are the fundamental pillars of the modern financial system, each serving distinct yet complementary roles. Capital markets, comprising stocks, bonds, and other long-term securities, enable businesses and governments to raise funds for long-term investments and expansion. These markets offer investors the potential for higher returns, but often with increased risk and
volatility
.
Money markets, meanwhile, focus on short-term, highly liquid instruments such as Treasury bills and commercial paper. They serve as the economy's lubricant, facilitating short-term borrowing and lending and providing a relatively safe haven for cash management. While money market instruments typically offer lower returns, they provide essential liquidity and stability to the financial system.
Article Sources
Investopedia requires writers to use primary sources to support their work. These include white papers, government data, original reporting, and interviews with industry experts. We also reference original research from other reputable publishers where appropriate. You can learn more about the standards we follow in producing accurate, unbiased content in our
editorial policy.
Statista. \"
Largest stock exchange operators worldwide as of March 2024, by market capitalization of listed companies
.\"
Board of Governors of the Federal Reserve System. \"
About Commercial Paper
.\"
Compare Accounts
Advertiser Disclosure
x
The offers that appear in this table are from partnerships from which Investopedia receives compensation. This compensation may impact how and where listings appear. Investopedia does not include all offers available in the marketplace.
\"\"\"

Please answer the following question based only on the information above:
What are the two most prevalent short-term debt instruments in the money market?"
\end{tcblisting}

This case looks at a situation where retrieval makes the answer worse. The question asks which two short-term debt instruments are the most common in the money market. The gold-standard answer is very short and simply states that the most prevalent instruments are Treasury Bills and Commercial Paper.

The base model, without retrieval, performs reasonably well and reaches a factual F1 score of about 0.62. It correctly identifies the two instruments and provides accurate explanations of their issuers, maturities, and uses. Its statements clearly match the key facts in the gold-standard answer. Although the base answer contains more description and is longer than necessary, the evaluation system still recognizes that it accurately covers the essential information.
With retrieval, the factual score drops dramatically to around 0.125. The answer produced with retrieval still names the correct instruments. It also includes additional details such as typical denominations and precise maturity ranges. The retrieved evidence itself even supports the idea that T-Bills and commercial paper are commonly used in the money market.

The problem comes from the answer's shift in emphasis. After incorporating retrieved content, the model focuses more on technical characteristics, such as the structure and numerical parameters of T-Bills. It focuses less on the central point that these two instruments are the most prevalent ones. Since the evaluation places strong weight on that specific fact about prevalence, the retrieved answer ends up covering the key point less directly. As a result, the factual match score becomes much lower, even though the content is technically correct and more detailed.

This case shows that retrieval does not always help. When the gold-standard answer highlights a specific categorical fact about importance or ranking, too much detail from retrieval can distract the model from stating that key fact clearly. The answer becomes more elaborate, but the essential information becomes blurred, which leads to a decline in factual accuracy during evaluation.

\subsection{Case Study 3: Neutral Impact of Retrieval on Regulatory Interpretation}
\begin{tcblisting}{
  enhanced,breakable,
  colback=white,colframe=Gray!50!black,
  colbacktitle=Gray!60!black,coltitle=white,
  title=Case Study 3: Neutral Impact of Retrieval on Regulatory Interpretation,
  listing engine=listings,
  listing only,
  listing options={
    basicstyle=\ttfamily\scriptsize,
    breaklines=true, columns=fullflexible, keepspaces=true,
    showstringspaces=false
  }
}
Q: "In producing the comprehensive cash flow projections for its U.S. operations, will an FBO be required to consolidate cash flows of entities subject to Regulation W, such as banks, with cash flows of entities not subject to Regulation W?"

GT: "Yes. As set forth in section 252.156(d) of Regulation YY, an FBO must produce comprehensive cash-flow projections for its combined U.S. operations. The comprehensive cash flow projections must include the cash flows of all entities included in the combined U.S. operations, regardless of whether the entities are subject to Regulation W."

base_answer: "In preparing comprehensive cash flow projections for its U.S. operations, an FBO (Financial Holding Company) is generally required to consolidate the cash flows of all its subsidiaries, including those subject to Regulation W, with the cash flows of entities not subject to Regulation W. This consolidation is typically required under Generally Accepted Accounting Principles (GAAP) and the rules governing financial reporting for holding companies.

Here are some key points to consider:

1. Consolidation Requirements: Under GAAP, an FBO must consolidate the financial statements of all its subsidiaries, including banks and other financial institutions, into its own financial statements. This means that the cash flows of all consolidated entities should be included in the FBO's cash flow projections.

2. Regulation W: Entities subject to Regulation W, such as banks, are part of the FBO's consolidated financial statements. Therefore, their cash flows must be included in the FBO's consolidated cash flow projections. This includes any cash flows related to loans extended to insiders, which are subject to the provisions of Regulation W.

3. Comprehensive Cash Flow Projections: When preparing comprehensive cash flow projections, the FBO should consider all sources of cash inflows and outflows, including those from its banking subsidiaries. This ensures a complete picture of the company's liquidity and financial health.

4. Regulatory Compliance: While Regulation W itself does not specifically mandate the inclusion of cash flows in projections, the broader regulatory framework and the need for accurate financial reporting under GAAP require this consolidation.

5. Disclosure Requirements: The FBO must ensure that its financial reports and disclosures comply with all applicable regulations, including those related to cash flow projections. This includes providing transparent and accurate information about the cash flows of all consolidated entities.

In summary, an FBO must consolidate the cash flows of all its subsidiaries, including those subject to Regulation W, with the cash flows of entities not subject to Regulation W, in order to produce comprehensive and accurate cash flow projections for its U.S. operations."

RAG_answer: "Based on the information provided, an FBO (Foreign Banking Organization) will be required to consolidate cash flows of entities subject to Regulation W, such as banks, with cash flows of entities not subject to Regulation W when producing comprehensive cash flow projections for its U.S. operations. Here's the reasoning:

1. Regulation W Requirements: The regulations specify that an FBO must produce comprehensive cash-flow projections for its combined U.S. operations. This includes projecting cash flows arising from transactions with entities that are not affiliates.

2. Cash Flow Projections Scope: The cash flow projections must account for cash-flow needs and sources resulting from transactions between the U.S. intermediate holding company and entities that are not its affiliates. This implies that all relevant cash flows, regardless of whether the entities are subject to Regulation W, must be included in these projections.

3. Net Internal and External Cash Flow Needs: The calculations for net internal and external stressed cash-flow needs involve detailed projections of cash flows from various transactions, including those with non-affiliates. These calculations are integral parts of the overall cash flow projection process.

Given these points, it is clear that the FBO must consider and include the cash flows of all relevant entities, whether or not they are subject to Regulation W, in its comprehensive cash flow projections for its U.S. operations. Therefore, the FBO will be required to consolidate cash flows of entities subject to Regulation W with those of entities not subject to Regulation W."

RAG_retrieved_chunks: "The following information has been retrieved and may be helpful:

1.
\"\"\"
(d)Cash-flow pr ojections.
(1)A for eign banking or ganization subject t o this subpar t must pr oduce compr ehensiv e cash-flow
projections for its combined U.S.The net external str essed cash-flow need for a
U.S. intermediate holding company equals the diff erence between:
(A)The pr ojected amount of cash-flow needs that r esults fr om tr ansactions between the U.S.
intermediate holding company and entities that ar e not its affiliates; and
(B)The pr ojected amount of cash-flow sour ces that r esults fr om tr ansactions between the
U.S. intermediate holding company and entities that ar e not its affiliates.(C)Net intr agroup cash flow .For any giv en da y of the planning horiz on, the net intr agroup cash
flow must equal the diff erence between:
(1)The amount of pr ojected cash-flow needs r esulting fr om tr ansactions between a U.S.
branch or U.S. agency and the for eign bank' s non-U.S. offices and its affiliates; and
(2)The amount of pr ojected cash-flow sour ces r esulting fr om tr ansactions between a
U.S.12 CF R P art 25 2 (up t o d ate as of 12/14/2023)
Enhanc ed P rudenti al S tand ards (R egul ation Y Y)12 CF R 25 2.156(b)(4)
12 CF R 25 2.156( d)(1) ( enhanc ed displ ay) pag e 115 of 192

(2)The for eign banking or ganization must establish a methodology for making cash-flow pr ojections for
its combined U.S.t o the extent permitted b y applicable law .
(e)Cash-flow pr ojections.
(1)A bank holding company subject t o this subpar t must pr oduce compr ehensiv e cash-flow pr ojections
that pr oject cash flows arising fr om assets, liabilities, and off-balance sheet exposur es o ver, at a
minimum, shor t- and long-term time horiz ons.
\"\"\"
2.
\"\"\"
(e)Cash-flow pr ojections.
(1)A bank holding company subject t o this subpar t must pr oduce compr ehensiv e cash-flow pr ojections
that pr oject cash flows arising fr om assets, liabilities, and off-balance sheet exposur es o ver, at a
minimum, shor t- and long-term time horiz ons. The bank holding company must update shor t-term
cash-flow pr ojections daily and must update longer-term cash-flow pr ojections at least monthly .br anch or U.S. agency and the for eign bank' s non-U.S. offices and its affiliates.
(D)Amounts secur ed b y highly liquid assets. For the purposes of calculating net intr agroup
cash flow of the U.S. br anches and agencies under this par agraph, the amounts of
intragroup cash-flow needs and intr agroup cash-flow sour ces that ar e secur ed b y highly
liquid assets (as defined in paragraph (c)(7) of this section) must be ex cluded fr om the
calculation.
\"\"\"
3.
\"\"\"
For purposes of this subpart, any company that would be an affiliate of a U.S. branch, agency, or commercial lending company of a foreign bank if such branch, agency, or commercial lending company were a member bank is an affiliate of the branch, agency, or commercial lending company if the company also is: (1) Directly engaged in the United States in any of the following activi-ties: 
VerDate Sep<11>2014 17:04 Jun 02,Any company that the Board determines by regula-tion or order, or that the appropriate Federal banking agency for the mem-ber bank determines by order, to have a relationship with the member bank, or any affiliate of the member bank, such that covered transactions by the member bank with that company may be affected by the relationship to the detriment of the member bank. (b) ''Affiliate'' with respect to a mem-ber bank does not include: (1) Subsidiaries. Any company that is a subsidiary of the member bank, un-less the company is: (i) A depository institution;[Reg. V, 72 FR 63758, Nov. 9, 2007, as amended at 74 FR 22642, May 14, 2009; 79 FR 30711, May 29, 2014] PART 223---TRANSACTIONS BE-TWEEN MEMBER BANKS AND THEIR AFFILIATES (REGULATION W) Subpart
A---Introduction and Definitions Sec. 223.1 Authority, purpose, and scope. 223.2 What is an ''affiliate'' for purposes of sections 23A and 23B and this part?Provision of Regulation W Application (1) 12 CFR 223.2(a)(8)---''Affiliate'' includes a fi-nancial subsidiary.Does not apply. Savings association subsidiaries do not meet the statutory definition of financial subsidiary.(a) In general. In some situations in which a member bank purchases an asset from an affiliate, the asset pur-chase qualifies for an exemption under this regulation, but the member bank's resulting ownership of the purchased asset also represents a covered trans-action (which may or may not qualify for an exemption under this part).
\"\"\"
4.
\"\"\"
offices and its U.S. and non-U.S. affiliates; and
(2)The pr ojected amount of cash-flow sour ces that r esults fr om tr ansactions between
the U.S. br anches and agencies and entities other than the for eign bank' s non-U.S.
offices and its U.S. and non-U.S. affiliates.
(iv)Net internal str essed cash-flow need calculation ---
(A)Gener al.The net internal str essed cash-flow need of the U.S. br anches and agencies of the
foreign banking or ganization equals the gr eater of:
(1)The gr eatest daily cumulativ e net intr agroup cash-flow need o ver the first 14 da ys of
the 30-da y planning horiz on, as calculated under paragraph (c)(3)(iv)(B) of this
section; and
(2)Zero.The net external str essed cash-flow need for a
U.S. intermediate holding company equals the diff erence between:
(A)The pr ojected amount of cash-flow needs that r esults fr om tr ansactions between the U.S.
intermediate holding company and entities that ar e not its affiliates; and
(B)The pr ojected amount of cash-flow sour ces that r esults fr om tr ansactions between the
U.S. intermediate holding company and entities that ar e not its affiliates.br anch or U.S. agency and the for eign bank' s non-U.S. offices and its affiliates.
(D)Amounts secur ed b y highly liquid assets. For the purposes of calculating net intr agroup
cash flow of the U.S. br anches and agencies under this par agraph, the amounts of
intragroup cash-flow needs and intr agroup cash-flow sour ces that ar e secur ed b y highly
liquid assets (as defined in paragraph (c)(7) of this section) must be ex cluded fr om the
calculation.(C)Net intr agroup cash flow .For any giv en da y of the planning horiz on, the net intr agroup cash
flow must equal the diff erence between:
(1)The amount of pr ojected cash-flow needs r esulting fr om tr ansactions between a U.S.
branch or U.S.
\"\"\"
5.
\"\"\"
12, 2002, unless otherwise noted. Subpart A---Introduction and Definitions Sec. 223.1 Authority, purpose, and scope. (a) Authority. The Board of Governors of the Federal Reserve System (Board) has issued this part (Regulation W) under the authority of sections 23A(f) and 23B(e) of the Federal Reserve Act (FRA) (12 U.S.C. 371c(f), 371c--1(e)) sec-tion 11 of the Home Owners' Loan Act (12 U.S.C.[Reg. V, 72 FR 63758, Nov. 9, 2007, as amended at 74 FR 22642, May 14, 2009; 79 FR 30711, May 29, 2014] PART 223---TRANSACTIONS BE-TWEEN MEMBER BANKS AND THEIR AFFILIATES (REGULATION W) Subpart A---Introduction and Definitions Sec. 223.1 Authority, purpose, and scope. 223.2 What is an ''affiliate'' for purposes of sections 23A and 23B and this part?(ii) If the Board determines that a particular transaction is,
in substance, a loan or extension of credit to an affil-iate that is engaged in activities other than those described at 12 U.S.C. 1467a(c)(2)(F)(i), as defined in Sec.238.54 of Regulation LL (12 CFR 238.54), or the Board has other supervisory concerns concerning the transaction, the Board may inform the savings association that the transaction is prohibited under this paragraph (c)(1), and require the savings association to divest the loan, unwind the transaction, or take other appropriate action.12 and the collateral requirements of Sec.223.14, and is otherwise permitted under this regulation; or (2) Making reference to such a guar-antee, acceptance, letter of credit, or cross-affiliate netting arrangement if otherwise required by law. Sec. 223.55 What are the standards under which the Board may
grant exemp-tions from the requirements of sec-tion 23B?12, 2002, as amended at 73 FR 54308, Sept. 19, 2008; 73 FR 55709, Sept. 26, 2008; 74 FR 6226, 6227, Feb.
\"\"\"

Please answer the following question based only on the information above:
In producing the comprehensive cash flow projections for its U.S. operations, will an FBO be required to consolidate cash flows of entities subject to Regulation W, such as banks, with cash flows of entities not subject to Regulation W?"
\end{tcblisting}

This question concerns whether a foreign banking organization must consolidate the cash flows of entities that are subject to Regulation W with those that are not, when producing comprehensive cash-flow projections for its U.S. operations. The gold-standard answer makes the requirement clear. Under Regulation YY section 252.156(d), an FBO must generate comprehensive cash-flow projections for its combined U.S. operations, and these projections must include all entities within that scope, regardless of whether they fall under Regulation W.

The base model, without retrieval, achieves a factual F1 score of about 0.44. Its answer relies mostly on accounting logic and the general principle of consolidated reporting. It notes that an FBO, under GAAP, consolidates the financial statements of its subsidiaries and therefore should include the cash flows of banks subject to Regulation W. Although the answer does not reference Regulation YY directly, it still captures part of the key idea in the gold-standard answer, namely that the cash flows of these entities must be included.

With retrieval, the factual score remains roughly the same at about 0.44. Retrieval brings in direct references to Regulation YY, and the retrieved statements explicitly note that an FBO must produce comprehensive projections for its combined U.S. operations and that these projections must include both entities subject to Regulation W and entities that are not. However, from the perspective of the evaluation metric, the base answer already covers the essential fact that these entities must be included. The retrieval-supported answer restates this using more formal regulatory language but does not introduce many new factual points that would increase coverage, so the score does not improve.

The retrieved source text confirms this pattern. In this case, retrieval mainly improves the clarity, structure, and alignment of the answer with the regulatory text, rather than changing the factual content. As a result, while retrieval makes the answer more authoritative and closer to official wording, it does not provide a noticeable improvement under a fact-coverage evaluation framework.

\end{document}